\documentclass[10pt,twocolumn,letterpaper]{article}

\usepackage[table,dvipsnames]{xcolor}
\usepackage{multirow}
\usepackage[pagenumbers]{cvpr} 
\usepackage{adjustbox}

\definecolor{cvprblue}{rgb}{0.21,0.49,0.74}

\newcommand{\redtext}[1]{\noindent \textcolor{red}{{#1}}}

\newcommand{\designbenchmark}{\textsc{Design-Multi-Layer-Bench}\xspace}
\newcommand{\overflowdesignbenchmark}{\textsc{overflowerflow-Design-Bench}\xspace}

\newlength\savewidth\newcommand\shline{\noalign{\global\savewidth\arrayrulewidth
  \global\arrayrulewidth 1pt}\hline\noalign{\global\arrayrulewidth\savewidth}}
\newcommand{\tablestyle}[2]{\setlength{\tabcolsep}{#1}\renewcommand{\arraystretch}{#2}\centering\footnotesize}

\definecolor{cvprblue}{rgb}{0.21,0.49,0.74}
\usepackage[pagebackref,breaklinks,colorlinks,allcolors=cvprblue]{hyperref}

\def\paperID{} 
\def\confName{CVPR}
\def\confYear{2026}

\title{MRT: Masked Region Transformer \\for Layered Image Generation and Editing at Scale}

\makeatletter
\let\cvpr@oldmaketitle\@maketitle
\renewcommand{\@maketitle}{%
  \cvpr@oldmaketitle
  \begin{center}
    \vspace{-2pt}
    \includegraphics[width=\textwidth]{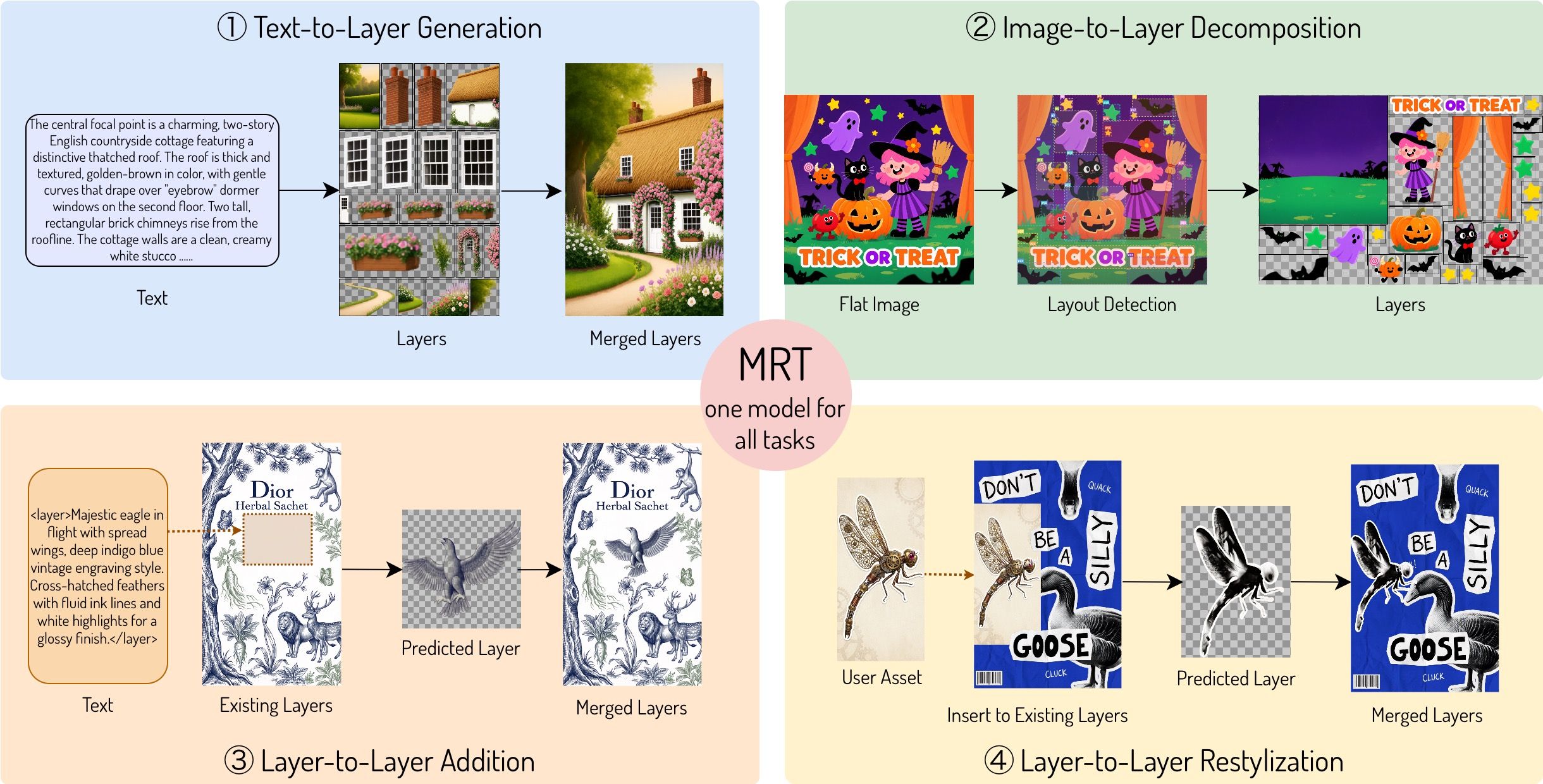}
    \vspace{-5mm}
    \captionsetup{type=figure}
    \captionof{figure}{\footnotesize{\textbf{Overview of Masked Region Transformer capabilities.} Our framework supports four tasks: (1) multilingual text-to-layers generation, (2) image-to-layers decomposition (including natural images), (3) layer addition, and (4) layer restylization for user-provided layers.}}
    \label{fig:teaser}
    \vspace{6pt}
  \end{center}
}
\makeatother

\author{
\small Zhicong Tang{$^\dagger$}\thanks{The superscript $\dagger$ indicates equal first contribution. Corresponding author: \texttt{ryanyuan@canva.com}}\;\;    Zhao Zhang{$^\dagger$}\;\;  Jingye Chen\;\;  Mohan Zhou\;\;  
 Yifan Pu\;\; Yuchi Liu\;\;  Yalong Bai\;\;  Ethan Smith\;\;  Yuhui Yuan \\ 
\small Canva Research
}

\begin{document}
\maketitle
\begin{abstract}
Layered image generation and editing is a fundamental capability that enables layer-wise reuse, editing, and composition of generated visual content, analogous to word-level editing in natural language. Despite its importance, this remains an underexplored area at scale. To address this gap, we present MRT, a 20B-parameter masked region diffusion model tailored for multi-layer transparent image generation and editing, trained on over 10M multilingual design samples spanning diverse aspect ratios and textual prompts. To fully leverage this scale, we make two key technical contributions. First, we unify three complementary tasks—text-to-layers, image-to-layers, and layers-to-layers—within a shared masked region diffusion framework, where selective token masking enables flexible layer-wise generation and editing. Second, to enable overflow layer generation, we introduce an overflow-aware canvas layer that handles boundary inconsistencies and supports semi-transparent background synthesis, enabling complete editable layers extending beyond visible canvas boundaries. Additionally, we apply diffusion distillation to achieve 8-step, real-time multi-layer generation with minimal quality degradation. Extensive experiments demonstrate that our framework substantially outperforms prior state-of-the-art approaches, including various commercial systems, across all three tasks, establishing a new benchmark for multi-layer transparent image generation.
Notably, our model significantly outperforms the concurrent Qwen-Image-Layered model in image-to-layers quality according to user-study results, while achieving {$10$$\sim$$100\times$} faster inference and saving a $50\%$$\sim$$ 90\%$ activation GPU memory consumption during image-to-layer inference.
\end{abstract}    
\section{Introduction}
\label{sec:intro}
Text-to-image generation has achieved remarkable quality improvements in recent years through various technological advances, including large-scale diffusion transformers~\cite{peebles2023scalable,esser2024scaling,ma2024sit}, distributed training on billions of high-quality text-image pairs~\cite{wu2025qwen,gong2025seedreamv2,gao2025seedreamv3,seedream2025seedreamv4}, rectified flow matching~\cite{esser2024scaling,lipman2022flow} that transforms simple prior distributions into complex data distributions via straight paths, distribution matching distillation~\cite{yin2024one,yin2024improved,sauer2024adversarial,zhou2024simple,zhu2025di,frans2024one,luo2024one,lu2024simplifying,zheng2025large} for accelerated inference, and advanced text encoder architectures~\cite{gong2025seedreamv2,liu2024playground,liu2024glyph}.
In contrast, generative models for layered image generation~\cite{zhang2024transparent,tudosiu2024mulan,li2023layerdiffusion,huang2024layerdiff,zhang2023text2layer,inoue2024opencole,jia2023cole,pu2025art,chen2025prismlayers} remain significantly underdeveloped. This gap primarily stems from two factors: the absence of large-scale, high-quality datasets comparable to LAION-5B~\cite{schuhmann2022laion}, and limited exploitation of prior knowledge from state-of-the-art open-source text-to-image models. These constraints have hindered systematic exploration of critical research directions in layered image synthesis.

We address this fundamental research gap through a comprehensive study on a high-quality, large-scale multi-layer dataset comprising over $\ge10$ million samples---an order of magnitude larger than recent work~\cite{pu2025art}. Our dataset spans diverse resolutions and aspect ratios, encompassing over $43$ million unique layers and over $7$ million unique oversized visual elements to support overflow layer generation. We employ GPT-5 mini to generate global captions for all graphic designs. For visual text layers, we utilize ground-truth typography attributes, ensuring comprehensive high-quality annotations. To fully leverage this dataset at scale, we build our multi-layer generative model by implementing the masked region transformer on Qwen-Image~\cite{wu2025qwen}, the largest open-source text-to-image diffusion model with approximately $\sim20$B parameters.

To advance the efficiency of layered image generation and editing during both training and inference, we introduce the following key technical contributions:
\emph{First}, we propose a unified masked region transformer framework that handles three complementary tasks: text-to-layers, image-to-layers, and layers-to-layers generation and editing. The key innovation lies in our adaptive masking mechanism, which determines whether to initialize each layer from clean latents or noise based on the specific task requirements. \emph{Second}, our masked region transformer operates directly on the full-size canvas by treating the background as a special transparent foreground layer and encapsulating overflow layers that extend partially beyond the background region. This architecture ensures that all foreground layers maintain full reusability and can be arbitrarily repositioned on the canvas, which is illustrated in Figure~\ref{fig:overflow_layers} and experimental section. \emph{Third}, we further propose leveraging distribution matching distillation schema to develop a few-step multi-layer generator with minimal quality degradation.

We conduct thorough ablation experiments to study the effects of different components. We empirically demonstrate that scaling both the model and dataset elevates performance to a new level, and that joint multi-task training further enhances performance while improving the user experience. We show that our image-to-layers task generalizes exceptionally well to various out-of-domain design images and natural images. Our layers-to-layers task readily supports multi-image fusion, seamlessly integrating any given user image into an existing design. We hope our masked region transformer advances the understanding of this fundamentally challenging task at an unprecedented scale.

\section{Related Work}
\label{sec:related_work}
Layered image generation and editing task follows two paradigms: simultaneous generation (Text2Layer~\cite{zhang2023text2layer}, LayerDiff~\cite{huang2024layerdiff}, ART~\cite{pu2025art}, PrismLayer~\cite{chen2025prismlayers}, Qwen-Image-Layered~\cite{yin2025qwenimagelayered}) and sequential generation (LayerDiffuse~\cite{zhang2024transparent}, COLE~\cite{jia2023cole}, OpenCOLE~\cite{inoue2024opencole}, LayerD~\cite{suzuki2025layerd}). Related layout generation and control methods fall into two categories: (1) generating layouts from visual elements~\cite{cheng2024graphic,shabani2024visual,kikuchi2024multimodal,feng2024layoutgpt,yamaguchi2021canvasvae,inoue2023layoutdm,chai2023layoutdm,hui2023unifying,kong2022blt,cheng2023play,tang2023layoutnuwa,jiang2022coarse,jiang2023layoutformer++,weng2024desigen,wang2023dolfin,guerreiro2025layoutflow,yang2024posterllava,inoue2023towards,chen2024textlap,fontanella2024generating,braunstein2024slayr}, and (2) controlling generation via spatial conditioning~\cite{li2023gligen,wang2024instancediffusion,wang2024msdiffusion,bar2023multidiffusion,yang2024mastering,kim2023dense,omost,sarukkai2024collage,zhang2024itercomp,feng2024layoutgpt,chen2024textlap}.
Compared to the most closely related work, ART~\cite{pu2025art} and Qwen-Image-Layered~\cite{yin2025qwenimagelayered}, our masked region transformer unifies three tasks: text-to-layers, image-to-layers, and layers-to-layers generation. We further introduce native support for overflow layers and enable few-step multi-layer generation through distillation.
\section{Approach}
\label{sec:approach}

\subsection{Scaling-up Layered Data and Diffusion Model}

\begin{figure*}[!t]
\includegraphics[width=\linewidth]{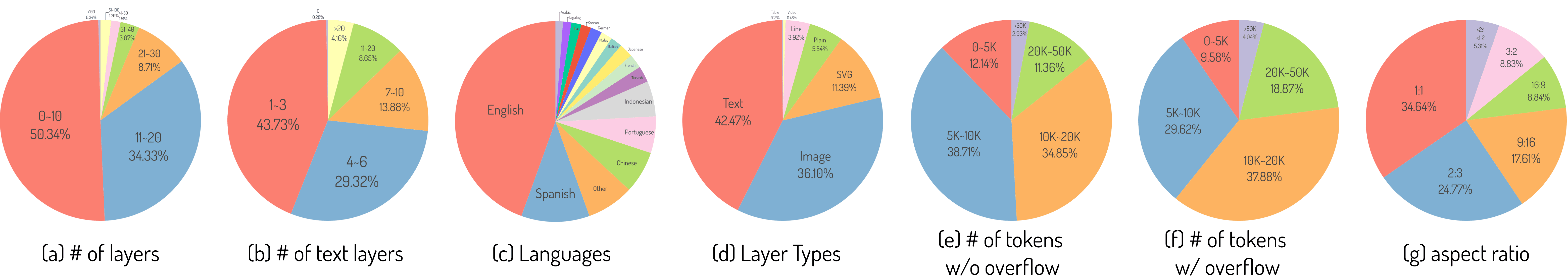}
\caption{\footnotesize{
\textbf{Illustrating the dataset statistics}. Figures (a) and (b) show the distribution of the number of unique layers per design. Figures (c) and (d) show the distribution of different languages in visual text and the distribution of different layer types, respectively. Figures (e) and (f) show the distribution of total visual token counts for all transparent layers before and after supporting overflow layers. Figure (g) shows the distribution of width-to-height aspect ratios.}}
\label{fig:statistics}
\vspace{-3mm}
\end{figure*}

\noindent\textbf{Scaled Layered Dataset.}
The scarcity of large-scale, high-quality multi-layer transparent images presents a fundamental challenge for advancing multi-layer generative modeling.
Rather than relying on noisy, uncurated internet sources, we construct a curated in-house dataset comprising over 10M multi-layer graphic designs from one of the world's largest graphic design platforms. All designs are created by professional designers and fully licensed for generative model training. Figure~\ref{fig:statistics} illustrates key dataset statistics, showing that our dataset spans diverse aspect ratios and resolutions while supporting multilingual visual text rendering and bilingual text prompts.

\begin{figure}[!t]
\begin{minipage}[!t]{1\linewidth}
    \begin{subfigure}[b]{1\textwidth}
    \centering
    \includegraphics[width=0.99\textwidth]{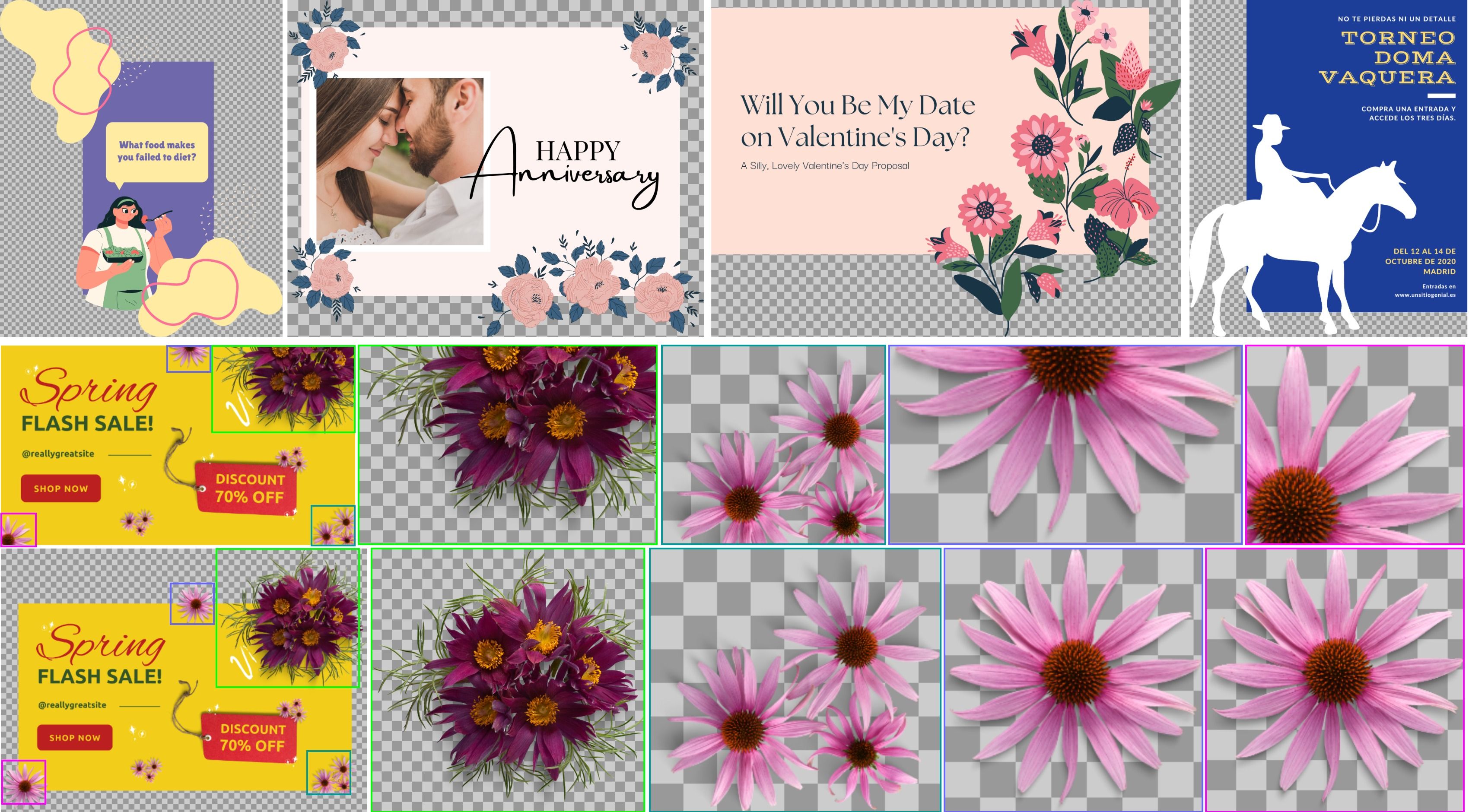}
    \end{subfigure}
\end{minipage}
\vspace{-2mm}
\caption{\footnotesize{
\textbf{Illustrating the overflowing layers}. The first row visualizes the canvas layer with a fully transparent background, exposing pixels beyond the main background region. Rows 2-3 compare multi-layer generation without overflow support (baseline) and with overflow support (ours). Full-size overflow layer generation is essential for maintaining complete editability and reusability, preventing layer content from being truncated at background boundaries.}}
\label{fig:overflow_layers}
\vspace{-5mm}
\end{figure}

\vspace{1mm}
\noindent\textbf{Scaled Region Transformer.}
To incorporate the generation of overflow layers, we follow ART~\cite{pu2025art} to perform the denoising diffusion process in a regional manner as follows:
First, we represent a multi-layer transparent image as \{$\mathbf{I}_\text{canvas}$, $\mathbf{I}_\text{bg}$, $\{\mathbf{I}_\text{fg}^{i}\}_{i=1}^K$\}, where $\mathbf{I}_\text{canvas}$ is the composed image on the full-size canvas, $\mathbf{I}_\text{bg}$ is a semi-transparent RGBA background layer, and $\{\mathbf{I}_\text{fg}^{i}\}_{i=1}^K$ are $K$ RGBA foreground layers.
Second, we perform the diffusion process on a merged image that integrates the fully transparent canvas as the base layer and overlays $\mathbf{I}_\text{bg}$ and all $\mathbf{I}_\text{fg}^{i}$ layers according to a predefined layout.
Third, we use the WAN-2.1-VAE~\cite{wan2025wan} encoder to extract the regional cropped representations for all foreground layers, the representation of the background layer, and the representation of the composed full design.
Last, we implement an anonymous regional diffusion transformer~\cite{pu2025art} with $20$B parameters following Qwen-Image~\cite{wu2025qwen} to perform full attention jointly on these regional foreground layer tokens, background layer tokens, and composed full design image tokens.

\vspace{1mm}
\noindent\textbf{Overflow Layer Support.}
Previous work~\cite{pu2025art,chen2025prismlayers} generates foreground layers only within the visible canvas region, producing incomplete elements that extend beyond background boundaries. This limits layer reusability, as shown in the second row of Figure~\ref{fig:overflow_layers}.
However, we find that over $60\%$ of samples in our training set contain overflow layers, making this a critical practical concern.
To address this, we introduce an additional full-size canvas layer that supports generation of complete semi-transparent backgrounds and overflowing elements. This is feasible since we have access to ground-truth complete layers for all samples in our dataset.
This design is essential for practical editing workflows: without it, layers extending beyond the canvas would be cropped and rendered non-editable, severely limiting their usability in downstream compositional tasks. Figure~\ref{fig:overflow_layers} shows representative overflow layer examples from our dataset (first row) and compares layered samples with and without overflow layer support (second and third rows).

\subsection{Masked Region Transformer}

We illustrate how our masked region diffusion transformer framework addresses three challenging multi-layer generation tasks—Text-to-Layers, Image-to-Layers, and Layers-to-Layers—in a unified manner in Figure~\ref{fig:framework}. The key insight is to conditionally mask either the global image tokens or the combination of reference tokens and existing layer tokens within the regional diffusion transformer. \emph{Masked latents} denote clean tokens encoding pre-existing conditions, with noise injection and diffusion supervision applied exclusively to non-masked tokens. We apply full attention between masked clean tokens and noise tokens, enabling the model to adaptively learn their relationships across different tasks. The detailed masking mechanism for each task is described as follows:

\begin{figure*}[!t]
\includegraphics[width=\linewidth]{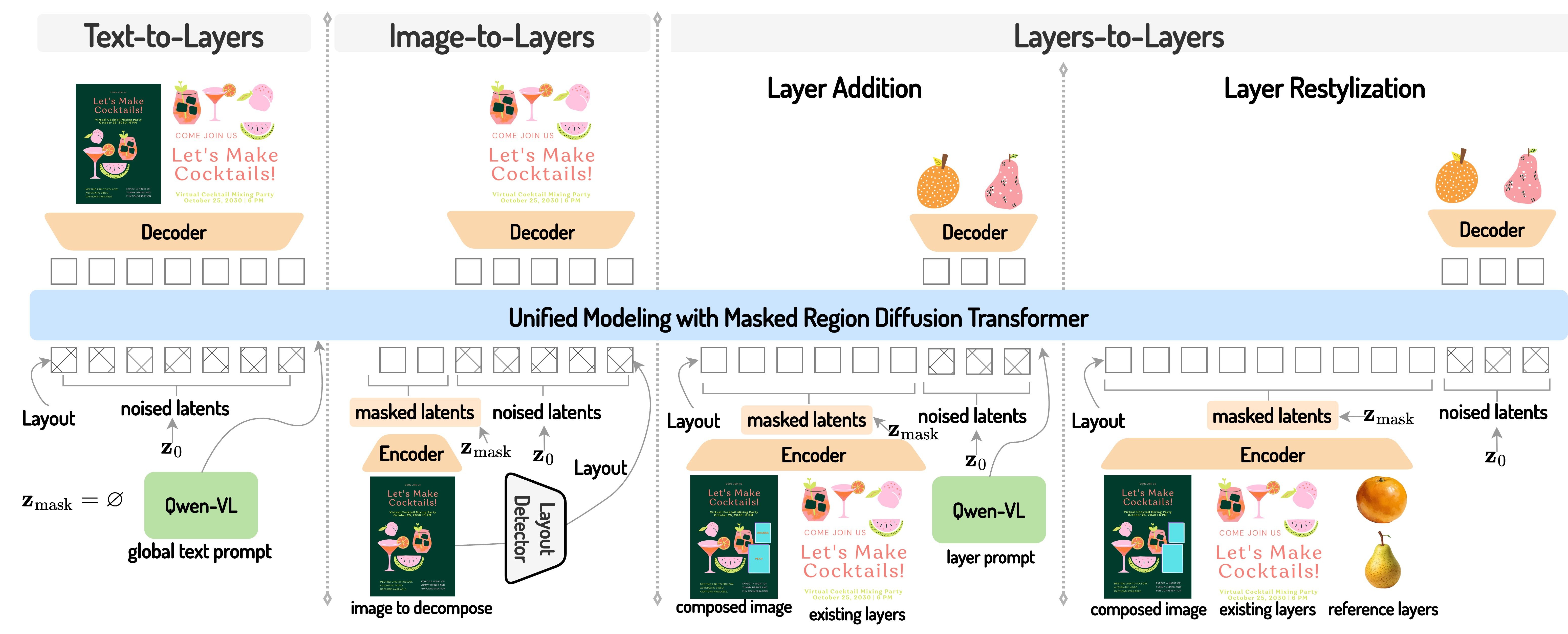}
\vspace{-5mm}
\caption{\footnotesize{
\textbf{Illustrating the Masked Region Transformer framework}.
We unify three different tasks including text-to-layers, image-to-layers, and layers-to-layers with a shared masked regional diffusion transformer.
\emph{Left}:~\textbf{Text-to-Layers} directly transforms a stack of noise latents into a set of transparent layers and a composed canvas image (panel \#1). We add noise to the latents of all transparent layers during training.
\emph{Middle}:~\textbf{Image-to-Layers} aims to decompose a raster image into a set of high-quality transparent layers. We set masked latents to the noise-free global image tokens and apply the diffusion process to layer tokens corresponding to spatial regions defined by either automatic layout detector or manual annotation (panel \#2).
\emph{Right}:~\textbf{Layers-to-Layers} enables two editing capabilities: (i) generating new layers from layer prompt conditioned on existing ones (panel \#3), and (ii) transforming reference images into layers with visual style aligned to the existing composition (panel \#4). In both layer addition and layer restylization scenarios, we define masked tokens as the noise-free latent representations of the reference content, existing layers, and global composition. Some text layers are omitted for clarity.
}
}
\label{fig:framework}
\vspace{-2mm}
\end{figure*}

\vspace{1mm}
\noindent\textbf{Text-to-Layers.}
The text-to-layers generation task aims to synthesize a multi-layer transparent design from a text prompt $\mathbf{c}$, comprising a canvas layer $\mathbf{I}_\text{canvas}$, a semi-transparent background layer $\mathbf{I}_\text{bg}$, and $K$ foreground layers $\{\mathbf{I}_\text{fg}^{i}\}_{i=1}^K$ that compose into $\mathbf{I}_\text{composed}$ with overflow support.
The canvas layer defines the full design dimensions to accommodate overflowing elements and is fully transparent by construction.
Thus we apply diffusion to the concatenation of latents $[\mathbf{z}_\text{composed}; \mathbf{z}_\text{bg}; \{\mathbf{z}_\text{fg}^{i}\}_{i=1}^K]$, excluding the canvas layer, conditioned on shared text embeddings $\mathbf{c}$. Following~\cite{pu2025art}, we include $\mathbf{z}_\text{composed}$ to ensure layer coherence. Since no pre-existing layers exist, we set masked token $\mathbf{z}_\text{mask}$ as $\varnothing$.
See Figure~\ref{fig:framework} (panel 1) for details.

Let $\mathbf{z}_0 = [\mathbf{z}_\text{composed}; \mathbf{z}_\text{bg}; \mathbf{z}_\text{fg}^{1}; \ldots; \mathbf{z}_\text{fg}^{K}]$ denote the concatenation of all non-masked clean latents, and $\epsilon \sim \mathcal{N}(\mathbf{0}, \mathbf{I})$ denote the noise prior. The flow matching framework learns a vector field that transports samples from the noise distribution to the data distribution through a continuous-time interpolation path.
At time-step $t \in [0, 1]$, the interpolated latent is given by:
\begin{equation}
\mathbf{z}_t = (1-t) \mathbf{z}_0 + t \epsilon,
\end{equation}

We train the diffusion model $f_\theta$ predicts the flow velocity conditioned on the interpolated latent $\mathbf{z}_t$, time-step $t$, and text prompt $\mathbf{t}$: $\hat{\mathbf{v}} = f_\theta(\mathbf{z}_t, t, \mathbf{c})$.
The training objective minimizes the mean squared error between the predicted and ground-truth velocity:
\begin{equation}
\mathcal{L}_\text{flow} = \mathbb{E}_{\mathbf{z}_0, \epsilon \sim \mathcal{N}(\mathbf{0}, \mathbf{I}), t} \left[ \| \mathbf{v}_t - f_\theta(\mathbf{z}_t, t, \mathbf{c}) \|^2 \right],
\end{equation}
where the ground-truth velocity $\mathbf{v}_t$ along the interpolation path is ($\mathbf{z}_0-\epsilon$), the expectation is taken over the clean latents $\mathbf{z}_0$, random noise $\epsilon$, and uniformly sampled time-steps $t$.

\vspace{1mm}
\noindent\textbf{Image-to-Layers.}
The image-to-layers task has emerged as a critical capability in commercial generative systems, with products such as Adobe Firefly's \emph{Layered Image Editing} and Lovart's \emph{Edit Elements} recently introducing support for this functionality.
The image-to-layers task aims to decompose a raster image $\mathbf{I}_\text{input}$ (or $\mathbf{I}_\text{composed}$) into a multi-layer transparent design comprising a canvas layer $\mathbf{I}_\text{canvas}$, a background layer $\mathbf{I}_\text{bg}$ and $K$ foreground layers $\{\mathbf{I}_\text{fg}^{i}\}_{i=1}^K$, conditioned on a target layout specifying each layer's spatial location and an optional text prompt for semantic guidance. This task inherently involves two subtasks: segmentation to identify layer regions with accurate alpha masks and inpainting to complete occluded areas.
We either use human annotations or a layout detector to extract the target layout from the input raster image.

The masked clean tokens $\mathbf{z}_\text{mask}$ are set to the global composed image representation $\mathbf{z}_\text{composed}$, encoding the conditional image targeted for decomposition.
We add noise to the concatenation of the non-masked tokens $\mathbf{z}_0 = [\mathbf{z}_\text{bg}; \mathbf{z}_\text{fg}^{1}; \ldots; \mathbf{z}_\text{fg}^{K}]$.
Through the regional diffusion process, the diffusion model $f_\theta$ is trained to extract all transparent layers conditioned on the given global image and layout. Since requiring users to provide designs with overflow layers is impractical, we instead use the latent encoding of pixels located within the visible canvas.
See Figure~\ref{fig:framework} (panel 2) for details.

We observe that individual layers often exhibit structural ambiguity and can be further decomposed. To address this, we propose \emph{layer grouping augmentation}, which randomly groups overlapping or adjacent layers during training. This strategy increases structural diversity, improves robustness to ambiguous boundaries, and enhances generalization to out-of-domain images with noisy layouts.

\vspace{1mm}
\noindent\textbf{Layers-to-Layers.}
To enable a flexible, layer-wise interaction experience, we frame the layered image editing task as a layer-to-layer task that covers two key scenarios: (i) \emph{layer addition}, which generates new coherent layers from text prompts conditioned on existing layers while maintaining spatial and stylistic consistency across the composition; and (ii) \emph{layer restylization}, which focuses on transforming any user-provided images or transparent layers into stylistically aligned layers that match the appearance and visual identity of the existing composition.

To model the layers-to-layers task, we retain existing layer latents as masked clean tokens $\mathbf{z}_\text{mask}$ and apply diffusion only to: (i) newly added layers conditioned on text prompts, or (ii) designated layers conditioned on visual references for restylization.
Given the challenge of constructing training data for these scenarios, we randomly select a subset of layers from each design to serve as conditional existing layers, treating the remaining layers as generation targets. For layer restylization training, we use Image editing model to transfer the style of non-selected layers, creating style-transformed variants as training pairs. See the appendix for details on the dataset construction pipeline.

Formally, in the layer addition task, we aim to synthesize a subset of foreground layers conditioned on the remaining layers and layer-level textual descriptions.
We apply diffusion to the latent token sequence \([\mathbf{z}_\text{composed}; \mathbf{z}_\text{bg}; \{\mathbf{z}_\text{fg}^{i}\}_{i=1}^{K}]\), where \(\mathbf{z}_\text{composed}\) encodes the alpha-composited context formed by the background and all non-target layers.
Let \(A \subseteq \{1,\ldots,K\}\) denote the indices of layers to be generated (an arbitrary subset, not necessarily contiguous).
We set the masked clean tokens as \(\mathbf{z}_\text{mask} = [\mathbf{z}_\text{composed}; \mathbf{z}_\text{bg}; \{\mathbf{z}_\text{fg}^{i}\}_{i\notin A}]\), and treat the target slots \(\mathbf{z}_0 = [\{\mathbf{z}_\text{fg}^{i}\}_{i\in A}]\) as the non-masked tokens to be noised and denoised.
The text condition \(\mathbf{c}_A\) is derived from a layer-caption prompt constructed by concatenating \texttt{<layer>} \(c_i\) \texttt{</layer>} for all \(i\in A\) in layer order, where \(c_i\) is the caption of layer \(i\).
During training, we add noise to \(\mathbf{z}_0\) and optimize the flow-matching objective conditioned on \((\mathbf{z}_\text{mask}, \mathbf{c}_A)\); during inference, we initialize \(\mathbf{z}_0\) from noise and denoise it under the same conditions, yielding the added layers in their original indices.

In the layer restylization task, we update a user-uploaded layered design by restylizing selected layers under additional appearance conditions while preserving the remaining layers.
Given target indices \(\mathcal{I} \subseteq \{1,\ldots,K\}\), we construct \(\mathbf{z}_\text{composed}\) by compositing the background with the non-target original layers \(\{\mathbf{z}_\text{fg}^{i}\}_{i\notin \mathcal{I}}\), and keep \(\mathbf{z}_\text{mask} = [\mathbf{z}_\text{composed}; \mathbf{z}_\text{bg}; \{\mathbf{z}_\text{fg}^{i}\}_{i\notin \mathcal{I}}]\) as masked clean conditions.
For each \(i \in \mathcal{I}\), we are additionally given a conditional latent \(\mathbf{z}_{\text{cond}}^{i}\) that specifies the desired appearance of layer \(i\).
We append \(\{\mathbf{z}_{\text{cond}}^{i}\}_{i\in \mathcal{I}}\) as extra conditioning tokens and treat them as masked, so they are not prediction targets.
To make this role explicit, we add a learnable condition-token embedding to the appended conditional tokens.
We further copy the RoPE positional encoding from the corresponding original layer token to its conditional token, ensuring that the two tokens share identical spatial positional cues.
Accordingly, we apply diffusion only to the non-masked original target slots \(\mathbf{z}_0 = [\{\mathbf{z}_\text{fg}^{i}\}_{i\in \mathcal{I}}]\), conditioned on \([\mathbf{z}_\text{mask}; \{\mathbf{z}_{\text{cond}}^{i}\}_{i\in \mathcal{I}}]\) and a fixed instruction prompt such as \texttt{Harmonize these layers}.
During training, noise is added only to \(\mathbf{z}_0\) and the model is trained to denoise the original target slots under the conditional latents.
During inference, we initialize \(\mathbf{z}_0\) from noise and denoise it under the same conditions, reading the final restylized layers from the original target slots while excluding the appended conditional tokens from the output layer set.

\subsection{Accelerated Multi-Layer Generator}
We adopt the improved \textit{distribution matching distillation} (DMD) technique~\cite{yin2024one,yin2024improved,luo2025tdm,dong2025glanceacceleratingdiffusionmodels}
to compress our multi-step diffusion model (teacher) into a few-step generator (student) while maintaining
distributional consistency between the teacher and student models.
Let the teacher model $f_{\theta_T}(\mathbf{z}_{t-1}|\mathbf{z}_t)$ denote the reverse process of a standard
multi-step diffusion model, and let the student model $f_{\theta_S}(\mathbf{z}_{t-1}|\mathbf{z}_t)$ approximate it
using fewer denoising steps. The objective of DMD is to minimize the Kullback--Leibler (KL) divergence
between the teacher and student transition distributions:
\begin{equation}
\footnotesize
\mathcal{L}_{\mathrm{DMD}}
=
\mathbb{E}_{\mathbf{z}_0 \sim p_{\textit{data}},\, t \sim \mathcal{U}(1, T)}
\!\left[
D_{\mathrm{KL}}
\!\big(
f_{\theta_T}(\mathbf{z}_{t-1}|\mathbf{z}_t)
\;\|\;
f_{\theta_S}(\mathbf{z}_{t-1}|\mathbf{z}_t)
\big)
\right].
\label{eq:dmd_kl}
\end{equation}

During inference, the distilled student model performs generation in a reduced number of steps
$T_S \ll T_T$, effectively approximating the teacher’s multi-step trajectory:
$\mathbf{z}_{t-1} = f_{\theta_S}(\mathbf{z}_t, t)$, where we set $t = T_S, \dots, 1$. We show that the distilled model preserves the sample quality of the teacher
while substantially reducing the number of sampling steps, resulting in faster and more efficient generation.
We also support various techniques, such as CacheDiT and sequence parallelization across multiple GPUs, to further accelerate inference speed.
\section{Experiment}
\label{sec:exp}

\subsection{Implementation Details}

We conduct all experiments using Qwen-Image as our base architecture, consisting of 60 layers with a hidden dimension of 3584 and 24 attention heads per layer. We initialize model weights from the open-source pretrained checkpoint available on HuggingFace. Unlike previous approaches~\cite{pu2025art,chen2025prismlayers} that fine-tune only LoRA~\cite{hu2022lora} weights due to resource constraints, we perform full-parameter fine-tuning with FSDP2 to explore the model's performance upper bound. This approach is necessary given the significant distribution shift from standard flat image generation and the inherent complexity of multi-layer synthesis.

For ablation experiments, we train on a curated subset of 0.5M layered designs for 4,000 iterations at $512\times512$ resolution using $8\times$H200 GPUs with the batch size 16 per GPU and 128 globally. We use the AdamW optimizer with a constant learning rate of $1\times10^{-4}$.
For system-level experiments, we employ two-stage training: $\sim$70,000 iterations at $512\times512$ on the full 10M dataset, followed by $\sim$20,000 iterations at $1024\times1024$. This progressive strategy allows the model to first establish multi-layer decomposition capabilities before scaling to high resolution. Training uses $64\times$H200 GPUs with batch size 16 per GPU and 1,024 globally.

\subsection{Evaluation Protocol}
\noindent\textbf{Benchmark.}
We compare our approach with previous state-of-the-art methods on \designbenchmark, introduced by ART~\cite{pu2025art}, which is curated from the VistaCreate graphic design platform~\cite{vistacreate}. However, this evaluation dataset does not include overflow layers. To address this gap, we construct \overflowdesignbenchmark to evaluate the model's ability to generate complete layers from full layouts, which is essential for ensuring overflow layer reusability.

\vspace{1mm}
\noindent\textbf{Metrics.}
We evaluate model performance from multiple perspectives. For merged image quality, we report PSNR${_{\text{layer}}}$, SSIM${_{\text{layer}}}$, PSNR${_{\text{merged}}}$, SSIM${_{\text{merged}}}$, FID${_{\text{merged}}}$ (measuring overall coherence), and FID following~\cite{pu2025art}. Since our layer is RGBA images with transparency, we only compute on non-transparent pixels as PSNR${_{\text{layer}}}$ and SSIM${_{\text{layer}}}$. For human evaluation, we collect multi-dimensional user preferences on a subset of \designbenchmark for the text-to-layers (T2L) task and image-to-layers (I2L) task, reflecting real user experience. The evaluation protocol and interface are described in the supplementary material.

\begin{figure*}[!t]
\centering
\includegraphics[width=\linewidth]{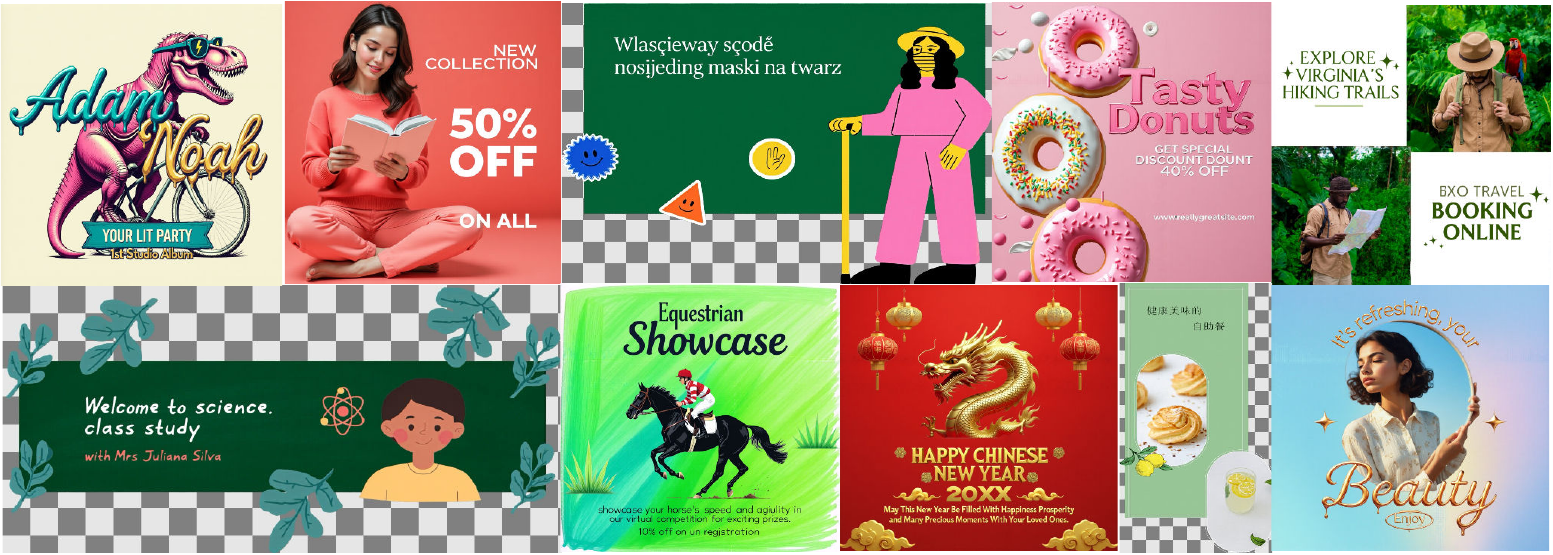}
\vspace{-5mm}
\caption{\footnotesize{\textbf{Qualitative results on text-to-layers.} See supplementary material for individual layer visualizations.}}
\label{fig:t2l_qual}
\end{figure*}

\begin{figure}[!t]
\centering
\includegraphics[width=\linewidth]{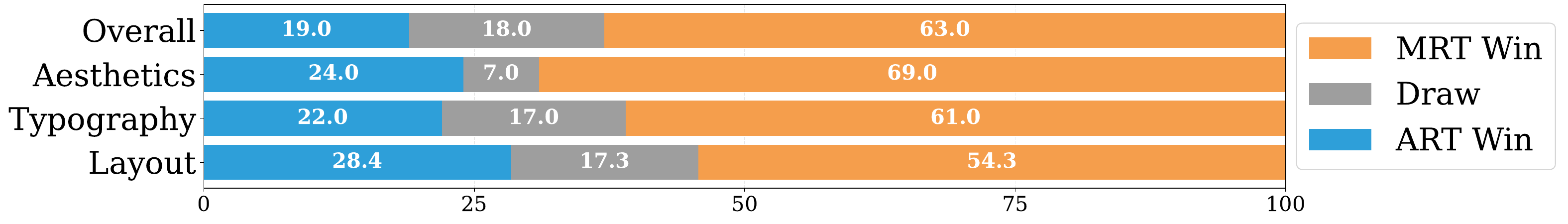}
\vspace{-5mm}
\caption{\footnotesize{\textbf{User study comparison with previous SOTA approach on text-to-layers task.} Our method significantly outperforms ART across multiple aspects.}}
\vspace{-3mm}
\label{fig:compete_t2l_user_study}
\end{figure}

\subsection{Main Results}

\subsubsection{Text-to-Layers: Comparison with SoTAs}
We compare our method with ART~\cite{pu2025art} on a subset of \designbenchmark. In our user study illustrated in Fig.~\ref{fig:compete_t2l_user_study}, participants consistently preferred our results over ART in instruction following, overall aesthetics, and layer quality. These findings indicate stronger alignment between prompts and layered compositions, further illustrated in Fig.~\ref{fig:t2l_qual} by layouts that better preserve spatial intent and stylistic consistency.

\begin{figure}[!t]
\centering
\includegraphics[width=\linewidth]{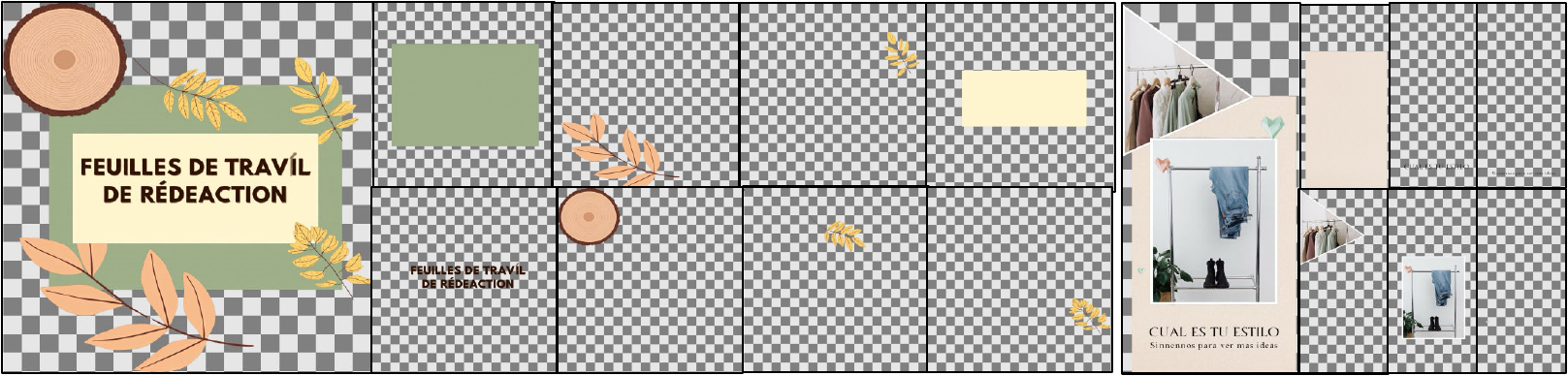}
\vspace{-5mm}
\caption{\footnotesize{\textbf{Qualitative results of layer overflow.} Our approach supports generating overflow layers with partially visible pixels extending beyond the background region.}}
\label{fig:t2l_overflow}
\vspace{-3mm}
\end{figure}

\begin{figure*}[!t]
\centering
\includegraphics[width=0.95\linewidth]{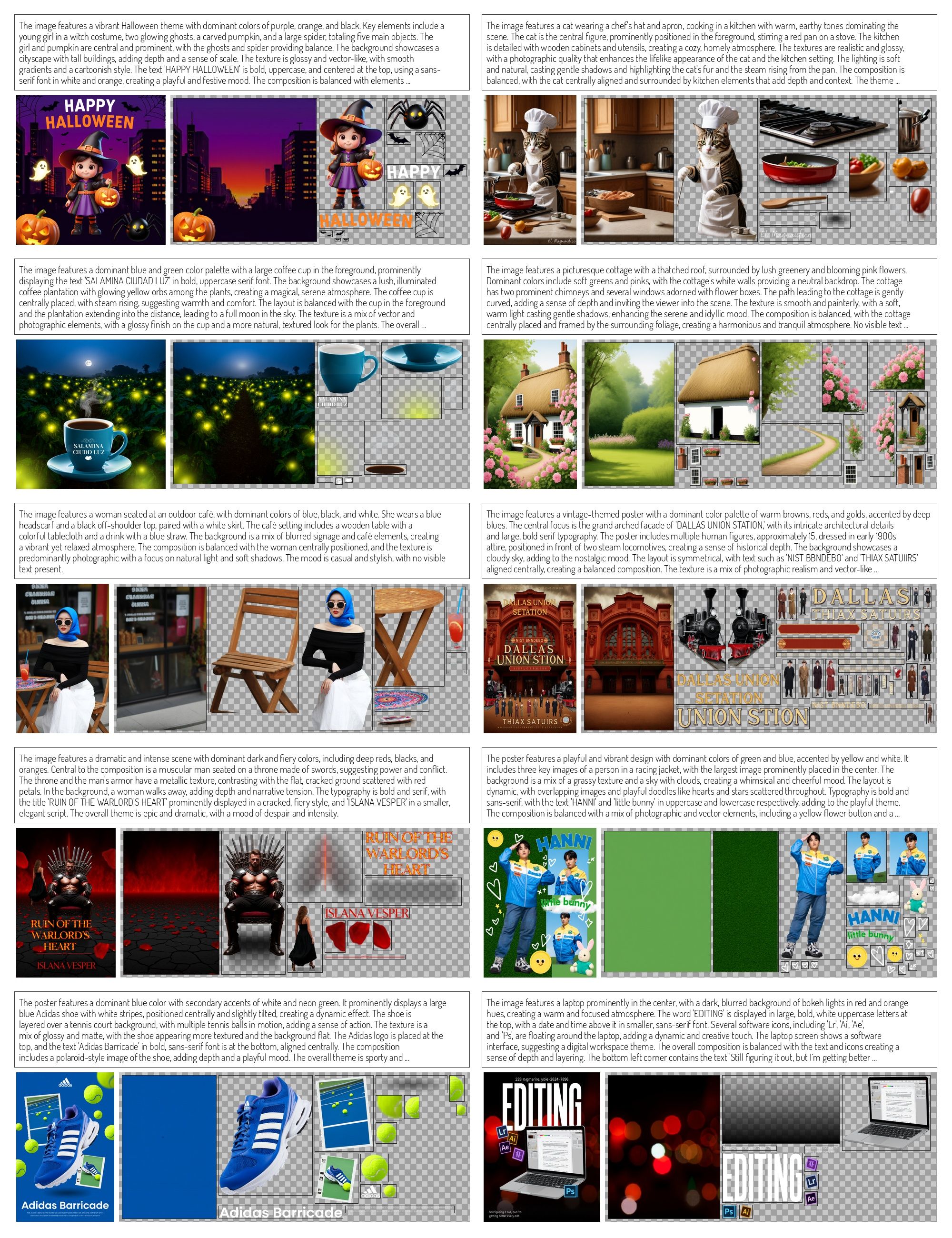}
\vspace{-2mm}
\caption{\footnotesize{{More Text-to-Layers Results.}}}
\vspace{-2mm}
\label{fig:t2l_ood}
\end{figure*}

Only our method natively supports generating overflow RGBA layers that extend beyond the background boundary on a full-size canvas, preserving editability and reuse; prior systems (\eg, ART) restrict pixels to the background region, leading to cropped or missing content. See Fig.~\ref{fig:t2l_overflow} and Fig.~\ref{fig:t2l_ood} for a visual results.

\subsubsection{Image-to-Layers: Comparison with SoTAs}
\begin{figure}[!t]
\centering
\includegraphics[width=1\linewidth]{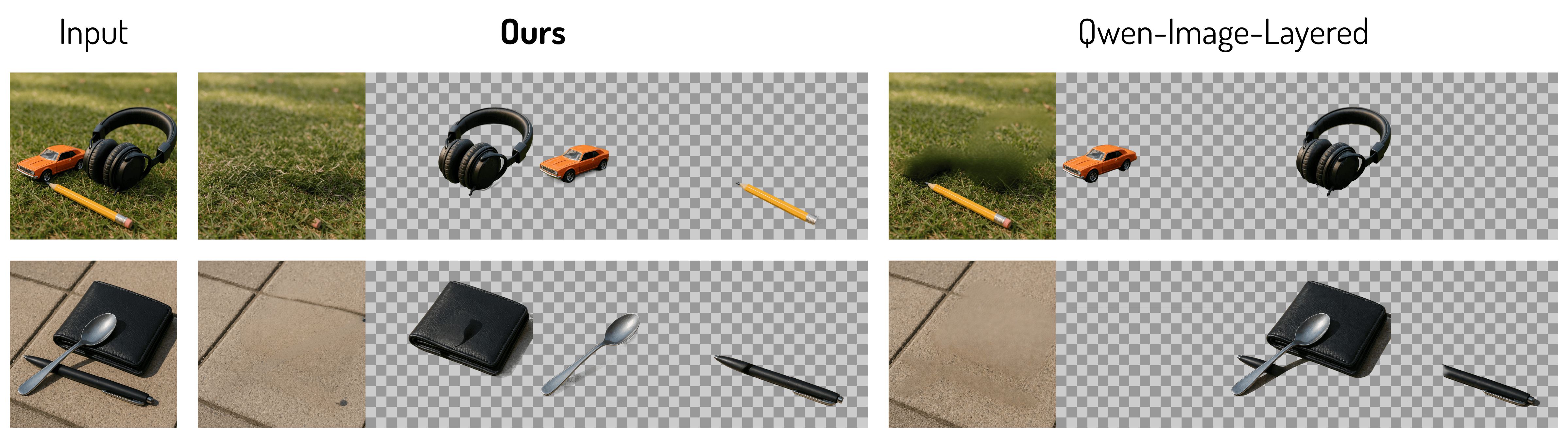}
\vspace{-5mm}
\caption{\footnotesize{\textbf{Qualitative results of image-to-layers on out-of-domain natural images.} Despite only trained on poster-style design datasets, our model generalizes to natural scenes.}}
\label{fig:i2l_natural}
\end{figure}

In a user study comparing the layer decomposition capabilities of the latest work LayerD~\cite{suzuki2025layerd} and commercial systems like RoboNeo and Lovart, participants consistently preferred our method for layer quality, content integrity, and decompose granularity.
Since I2L evaluation assumes a layer layout (bounding boxes with Z-order), we evaluate our method with the layout extracted by a {z-order-aware detector} (details in the supplementary). Qualitative comparisons in Fig.~\ref{fig:i2l_qual_layerd} also show that our method produces more complete, reusable RGBA layers with sharper boundaries. We further demonstrate the generalization of our model to natural scenes in Fig.~\ref{fig:i2l_natural}.

\subsubsection{Image-to-Layers: Comparison with con-current Qwen-Image-Layered}
Recently, Qwen-Image-Layered~\cite{yin2025qwenimagelayered} has attracted significant interest from the community since its release on Huggingface, due to its strong generalization capability on various design images. We demonstrate the advantages of our approach by  conducting rigorous comparisons from three aspects: \emph{quality}, \emph{latency}, and \emph{memory}.

\vspace{1mm}
\noindent\textbf{Better Quality.}
We first construct an out-of-domain test set consisting of 100 creative designs obtained from three sources: images generated by the latest Nano-Banana-Pro (and Ideogram 3.0) image generation model and test images from the official Qwen-Image-Layered repository~\cite{qwen-image-layered-repo}. We report the quantitative comparison results in Table~\ref{tab:i2l_qwen_compare}, which shows that our approach achieves significantly higher SNR${_{\text{merged}}}$ and SSIM${_{\text{merged}}}$. We calculate the metrics across three groups based on the number of layers, and our MRT consistently performs better across all groups.

\begin{figure}[!t]
\centering
\includegraphics[width=\linewidth]{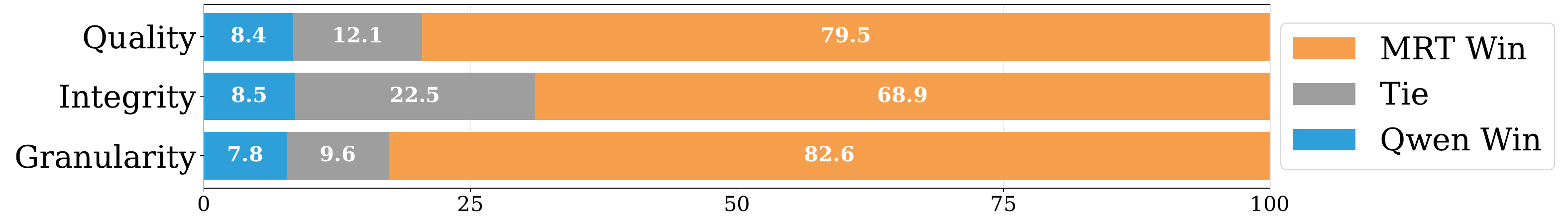}
\vspace{-5mm}
\caption{\footnotesize{\textbf{User study comparison with Qwen-Image-Layered on image-to-layers task.}}}
\vspace{-3mm}
\label{fig:compete_qwen_user_study}
\end{figure}

\begin{figure*}[!t]
\centering
\includegraphics[width=0.95\linewidth]{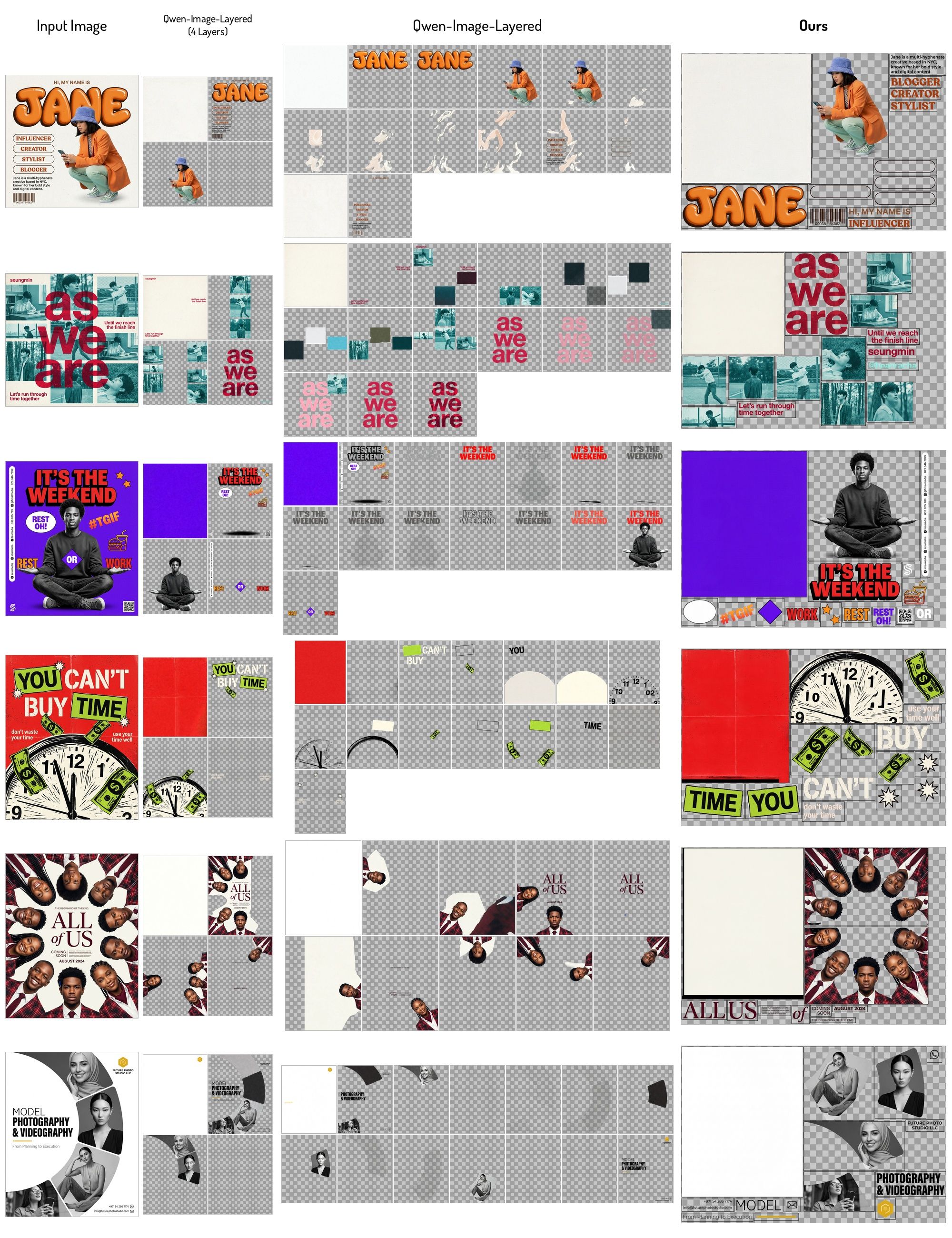}
\vspace{-2mm}
\caption{\footnotesize{{Image-to-Layers Results on Designs Generated with Nano-Banana-Pro (1/3): Comparison with Qwen-Image-Layered}}}
\vspace{-2mm}
\label{fig:i2l_qwen_compare_1}
\end{figure*}

\begin{figure*}[!t]
\centering
\includegraphics[width=0.95\linewidth]{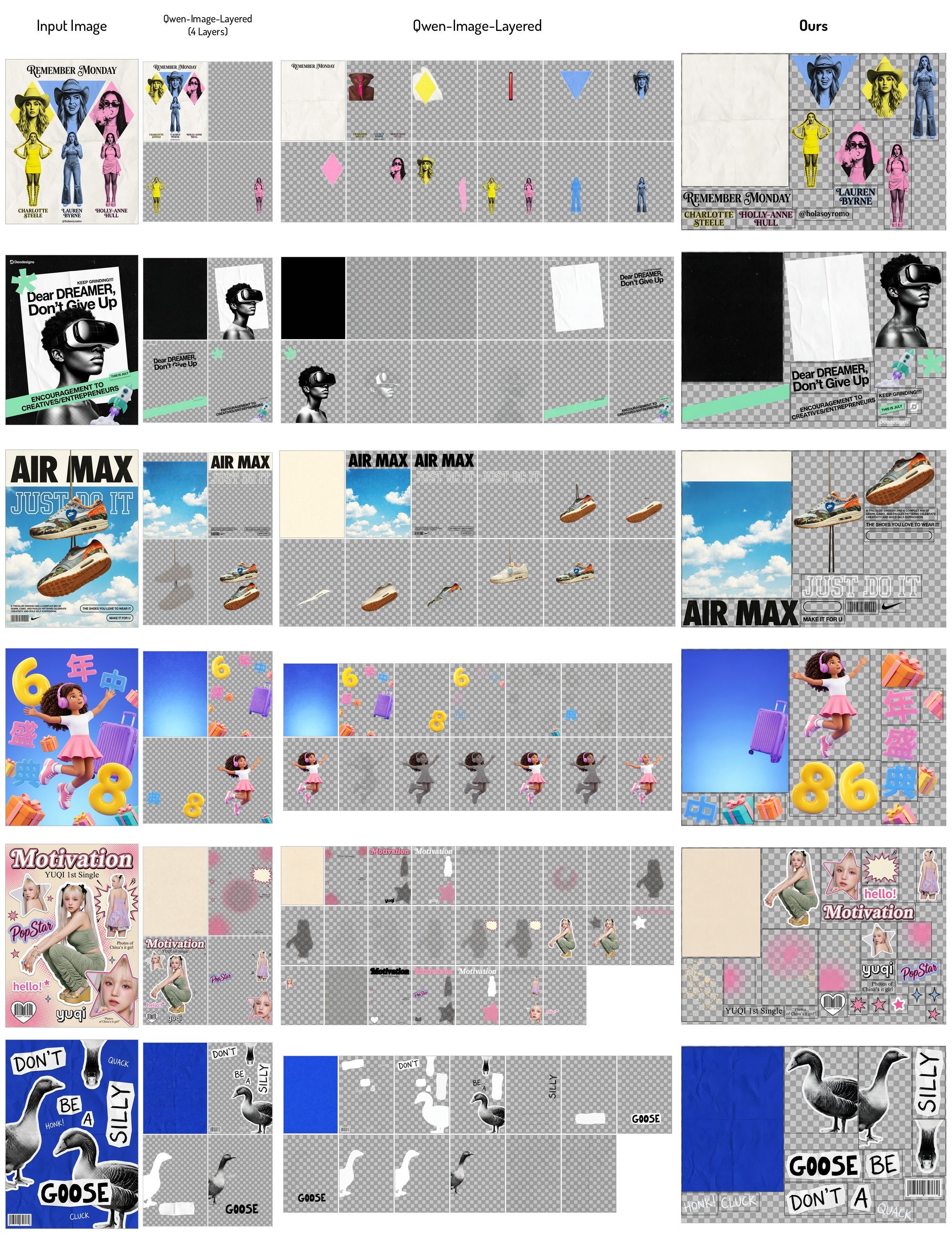}
\vspace{-2mm}
\caption{\footnotesize{{Image-to-Layers Results on Designs Generated with Nano-Banana-Pro (2/2): Comparison with Qwen-Image-Layered}}}
\vspace{-2mm}
\label{fig:i2l_qwen_compare_2}
\end{figure*}

\begin{figure*}[!t]
\centering
\includegraphics[width=0.95\linewidth]{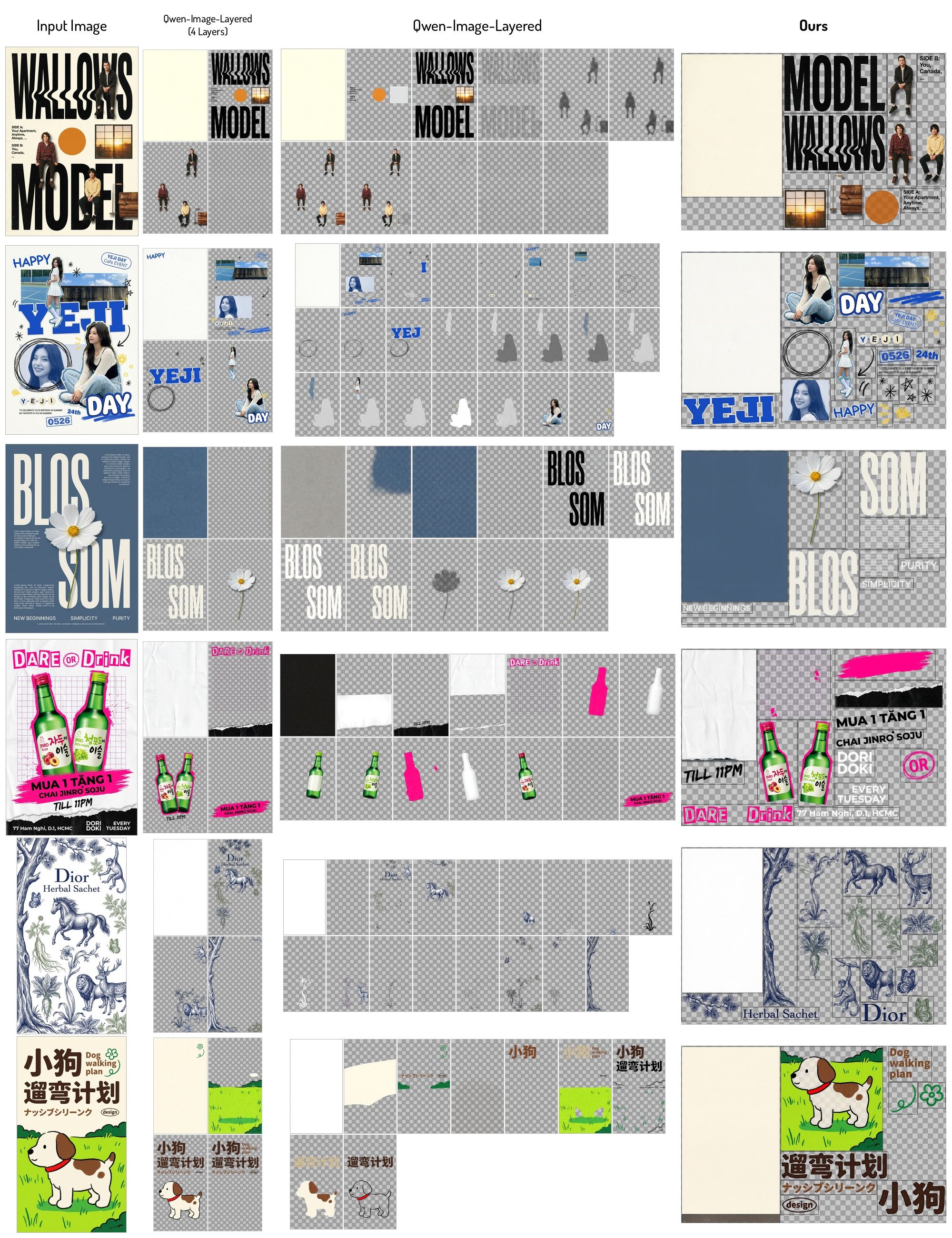}
\vspace{-2mm}
\caption{\footnotesize{{Image-to-Layers Results on Designs Generated with Nano-Banana-Pro (3/3): Comparison with Qwen-Image-Layered}}}
\vspace{-2mm}
\label{fig:i2l_qwen_compare_3}
\end{figure*}

\begin{figure*}[!t]
\centering
\includegraphics[width=0.95\linewidth]{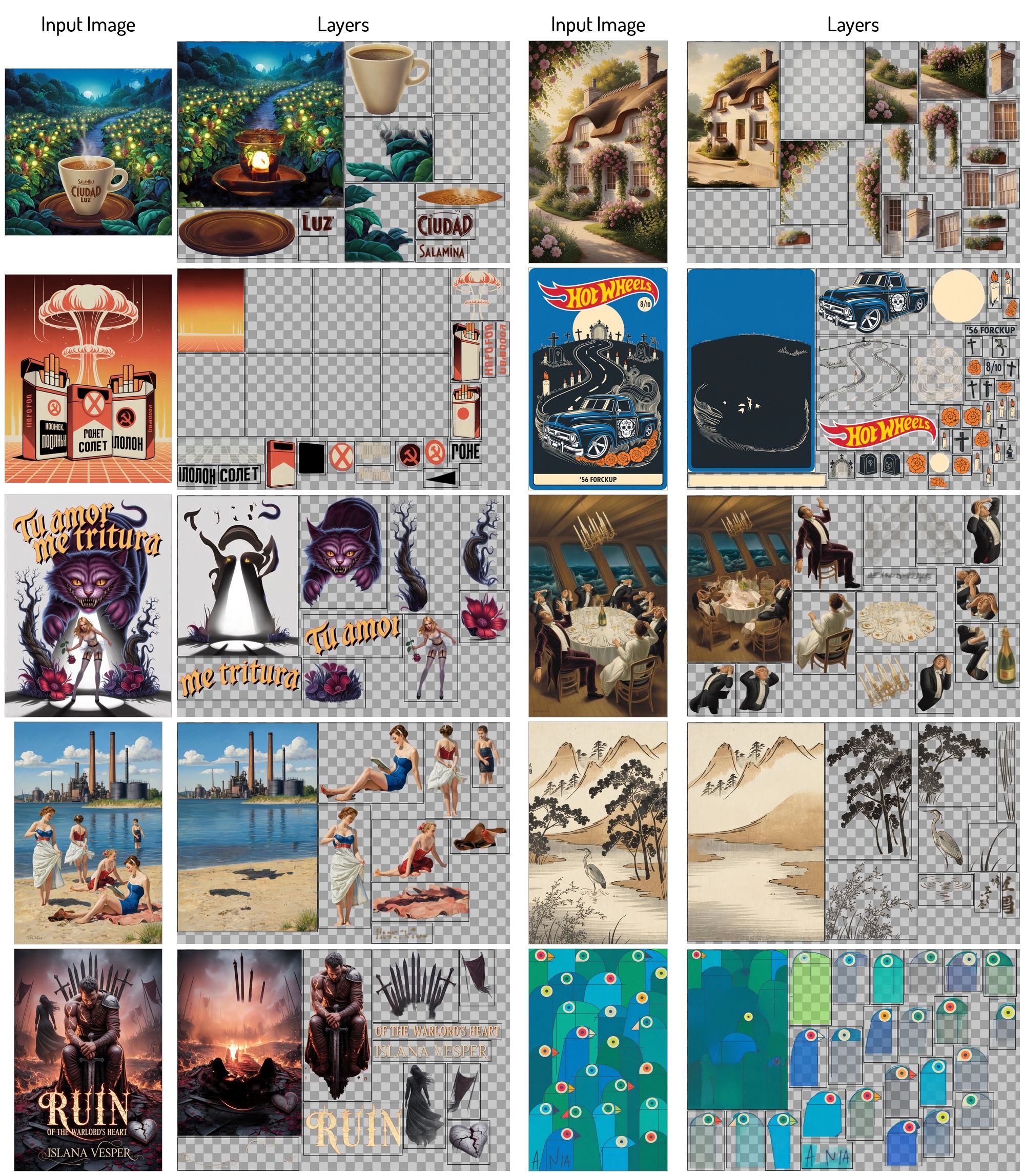}
\vspace{-2mm}
\caption{\footnotesize{{More Image-to-Layers Results on Designs Generated with Ideogram (1/2).}}}
\vspace{-2mm}
\label{fig:i2l_ideogram_1}
\end{figure*}

\begin{figure*}[!t]
\centering
\includegraphics[width=0.95\linewidth]{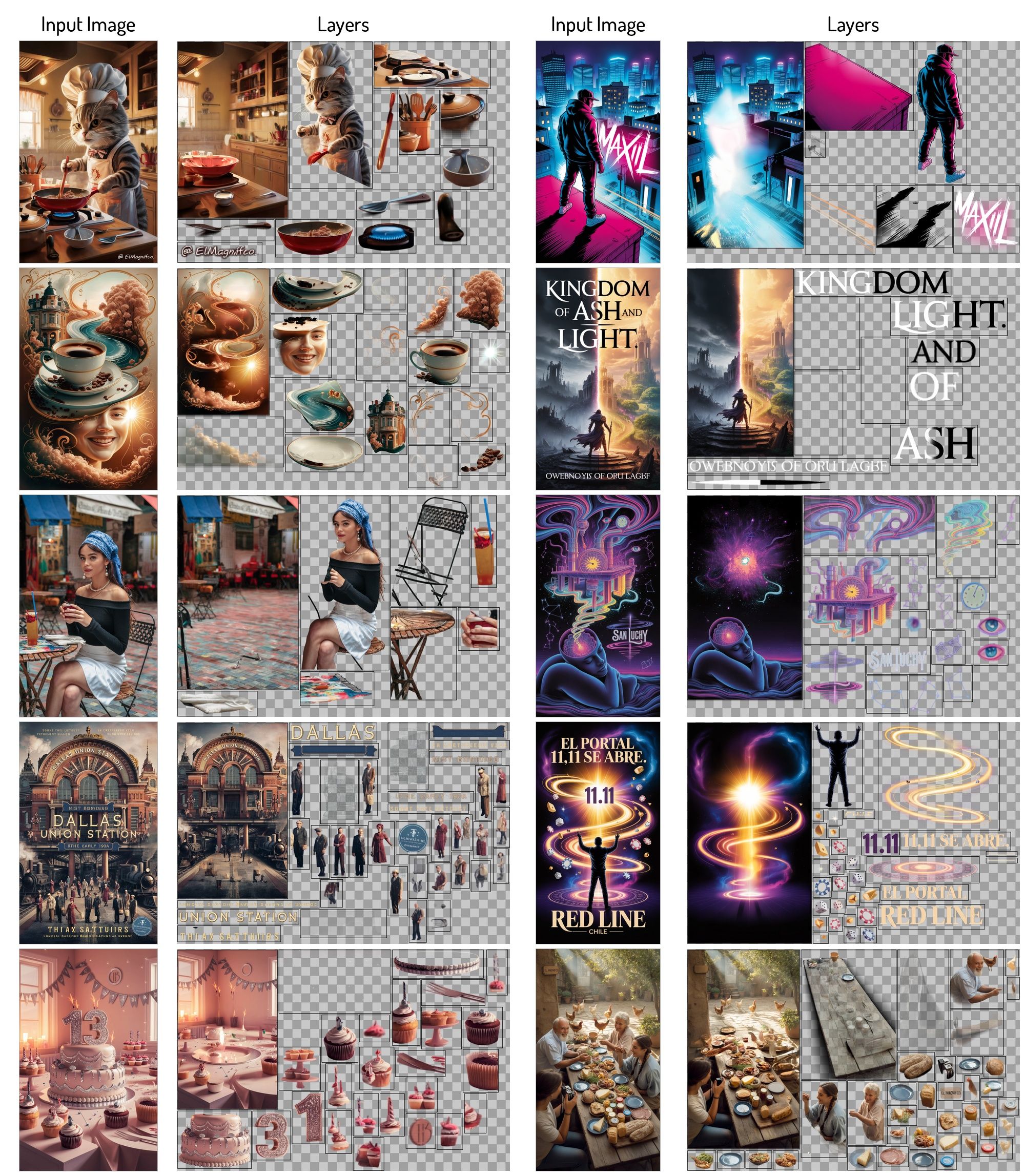}
\vspace{-2mm}
\caption{\footnotesize{{More Image-to-Layers Results on Designs Generated with Ideogram (2/2).}}}
\label{fig:i2l_ideogram_2}
\vspace{-2mm}
\end{figure*}

\begin{figure*}[!t]
\centering
\includegraphics[width=1\linewidth]{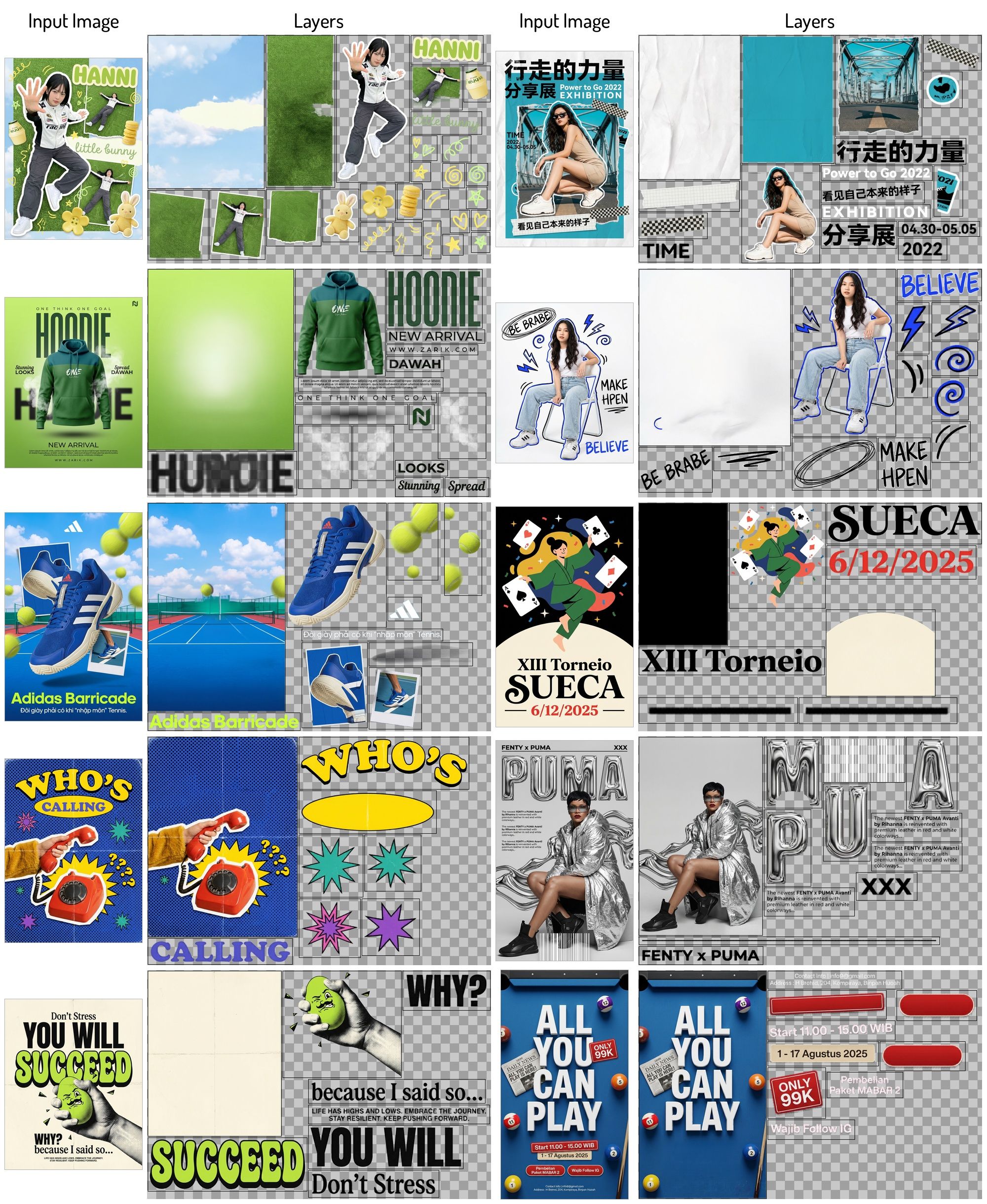}
\caption{\footnotesize{{More Image-to-Layers Results on Designs Generated with Nano-Banana-Pro (1/2).}}}
\vspace{-2mm}
\label{fig:i2l_pinterest_1}
\vspace{-2mm}
\end{figure*}

\begin{figure*}[!t]
\centering
\includegraphics[width=1\linewidth]{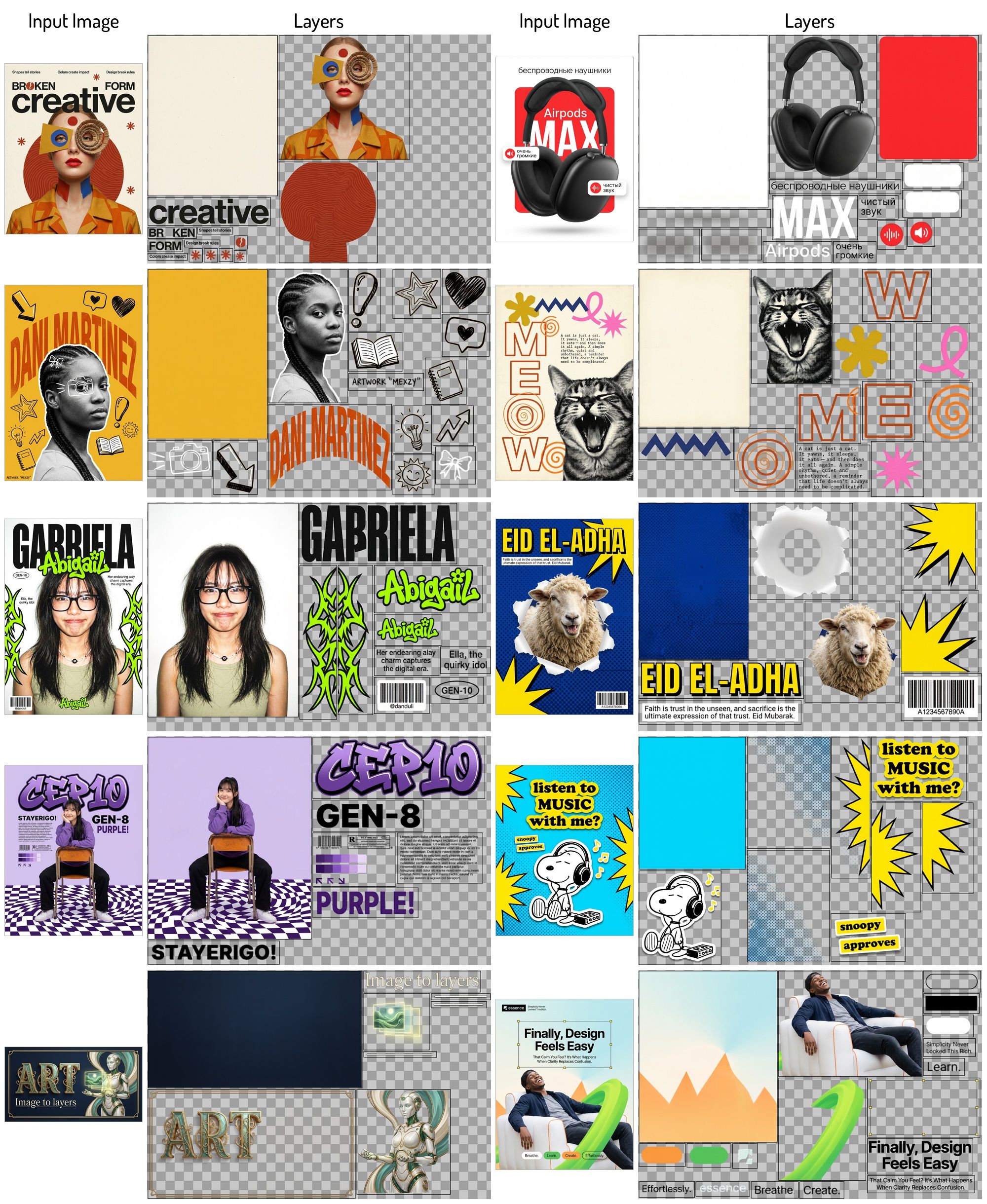}
\caption{\footnotesize{{More Image-to-Layers Results on Designs Generated with Nano-Banana-Pro (2/2).}}}
\vspace{-2mm}
\label{fig:i2l_pinterest_2}
\vspace{-2mm}
\end{figure*}

\begin{figure*}[!t]
\centering
\includegraphics[width=0.9\linewidth]{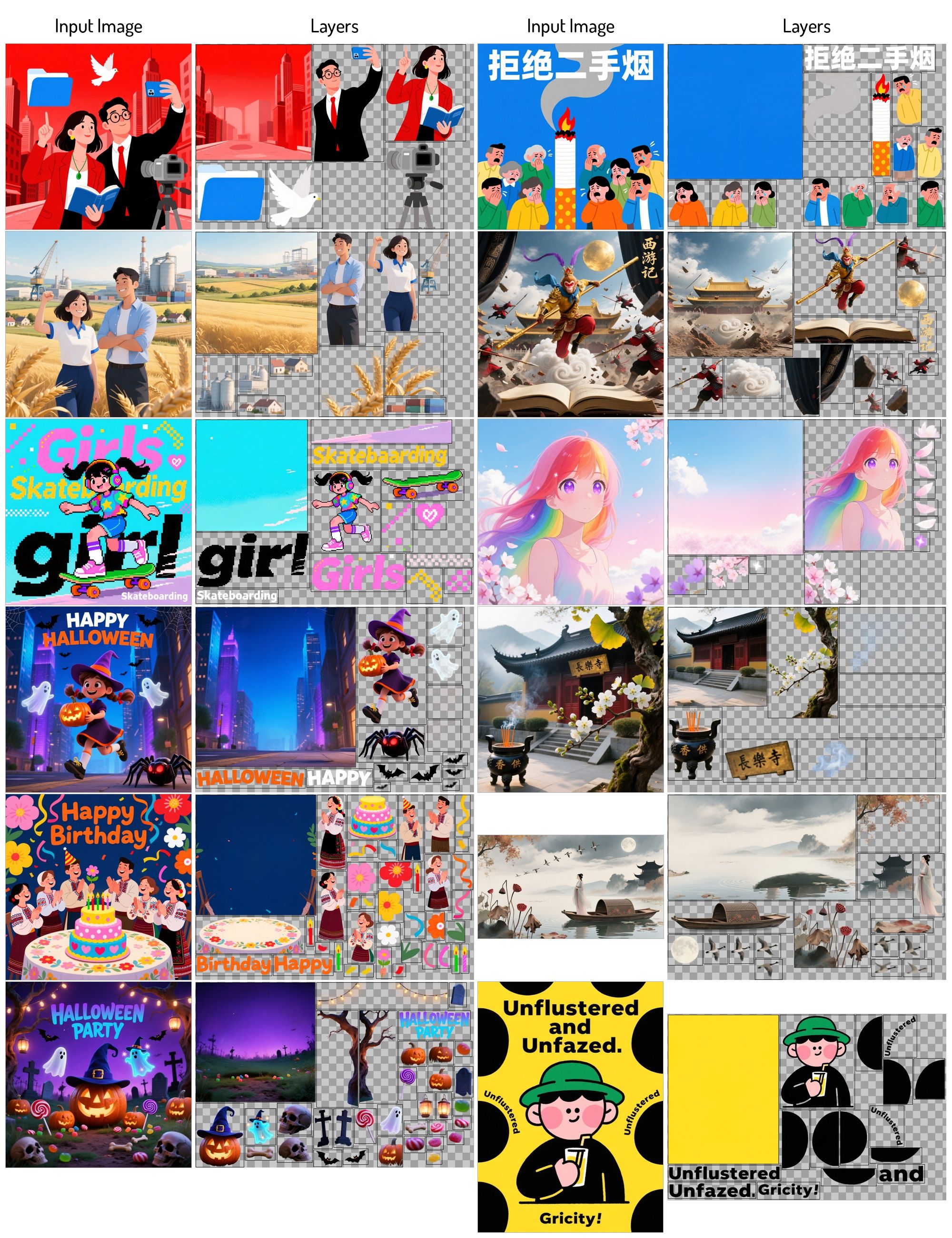}
\caption{\footnotesize{{More Image-to-Layers Results on Qwen-Image-Layered test set.}}}
\vspace{-2mm}
\label{fig:i2l_qwen}
\vspace{-2mm}
\end{figure*}

Fig.~\ref{fig:i2l_qwen_compare_1}, Fig.~\ref{fig:i2l_qwen_compare_2}, Fig.~\ref{fig:i2l_qwen_compare_3}, Fig.~\ref{fig:i2l_ideogram_1}, Fig.~\ref{fig:i2l_ideogram_2}, Fig.~\ref{fig:i2l_pinterest_1}, Fig.~\ref{fig:i2l_pinterest_2}, and Fig.~\ref{fig:i2l_qwen} provide further qualitative comparison results. We empirically find that our approach performs substantially better when required to decompose flat designs into an increasing number of transparent layers; our approach continues to perform well, while Qwen-Image-Layered struggled to assign meaningful objects to each layer. These visual results not only echo the above findings but also show that significant room for improvement remains, even though our model substantially outperforms Qwen-Image-Layered.
We also conduct an apple-to-apple user study on this test set, with results reported in Fig.~\ref{fig:compete_qwen_user_study}. Our approach achieves win rates of $79.5\%$, $68.9\%$, and $82.6\%$ for layer quality, integrity, and granularity, respectively.

\begin{figure*}[!t]
\centering
\includegraphics[width=\linewidth]{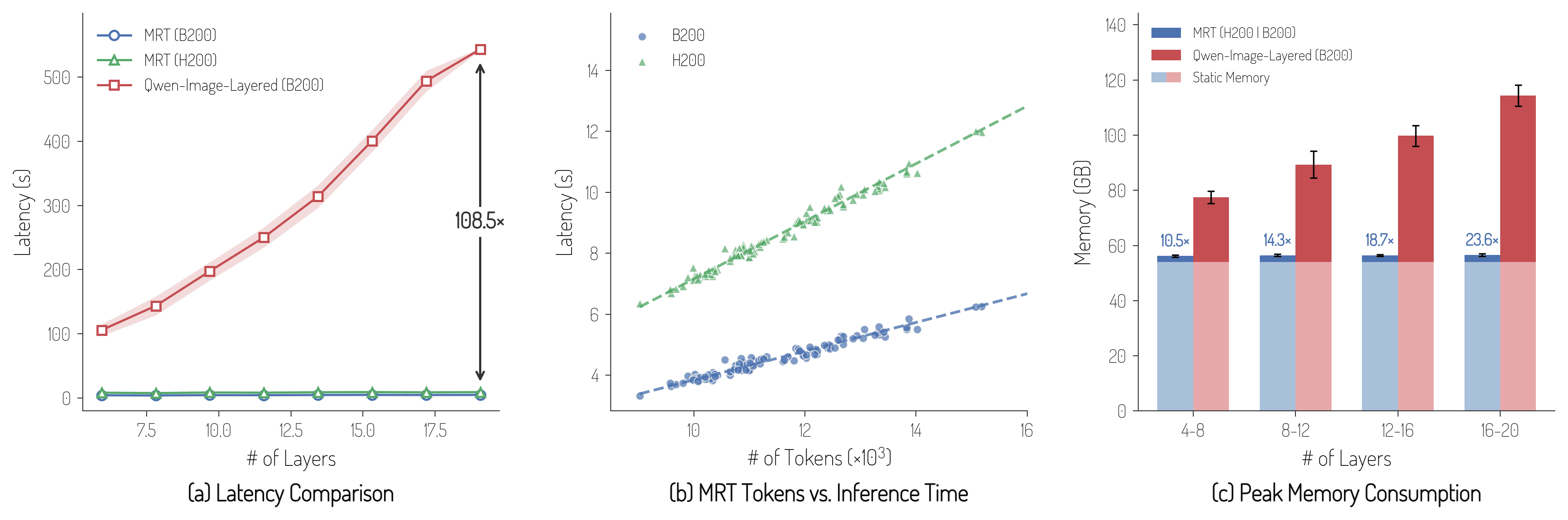}
\vspace{-5mm}
\caption{\footnotesize{\textbf{Inference efficiency comparison between MRT and Qwen-Image-Layered.} (a) Latency scaling with number of layers. MRT maintains near-constant latency ($\sim$5s) while Qwen-Image-Layered scales linearly, resulting in up to 108.5$\times$ speedup at $\sim$20 layers. (b) MRT inference time vs. token count on H200 and B200 GPUs, demonstrating linear scaling behavior. (c) Peak GPU memory consumption across varying layer configurations. The shaded region indicates the baseline memory allocated to model weights. MRT reduces memory consumption by  $10.5\times\to23.6\times$, with efficiency gains scaling proportionally with layer numbers. \textit{All reported results are conducted over 100 samples on single GPU with identical layer numbers.}}}
\vspace{-3mm}
\label{fig:inference_efficiency_comparison}
\end{figure*}

\vspace{1mm}
\noindent\textbf{Lower Latency.} As shown in Fig.~\ref{fig:inference_efficiency_comparison}, due to our regional diffusion transformer architecture, we achieve significant speedup compared to Qwen-Image-Layered, which uses the same number of full-resolution tokens to model each transparent layer regardless of their actual area within the canvas. We achieve similar latency speed-up as the statistics shown in ART~\cite{pu2025art} and we further applied various advanced cache techniques, model distillation, lower-precision, parallel inference to optimize the latency of our model to within \redtext{$\sim3$} seconds when running with $4\times$ H100 GPUs and \redtext{$\sim6$} seconds on a single H100 GPU when required to decompose a single 1K high-resolution image into nearly 20 transparent layers.

\vspace{1mm}
\noindent\textbf{Efficient Memory.}
Unlike Qwen-Image-Layered, which requires ${K}\times$ more visual tokens to extract ${K}\times$ different layers from a flat image, our approach is significantly more memory efficient, requiring far fewer tokens to decompose an image into many transparent layers. Fig.~\ref{fig:inference_efficiency_comparison} shows latency vs. number of layers, latency vs. number of tokens, and peak memory consumption vs. number of layers. Our method achieves clear advantages in both inference speed and memory usage; for example, generating more then $20$ layers with our MRT results in over $100\times$ acceleration.

\begin{figure*}[!t]
\centering
\includegraphics[width=\linewidth]{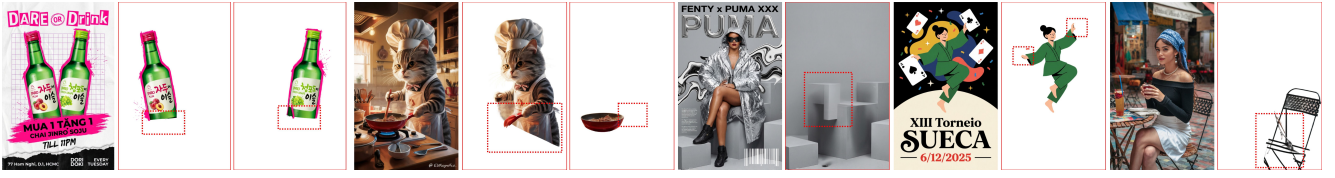}
\vspace{-5mm}
\caption{\footnotesize{\textbf{Illustrating the Challenges of the Image-to-Layers.} We show some representative failure cases when handling occluded layer completion. We find that our model fails to generate the occluded parts due to the regional crop design when the bounding boxes are tightly fit around only the visible pixels. We suspect another key reason is that these test cases differ from our training data distribution, and we leave this challenge to future work.}}
\vspace{-3mm}
\label{fig:i2l_failures}
\end{figure*}

\vspace{1mm}
\noindent\textbf{Challenges.}
We identify several remaining key challenges in the image-to-layer decomposition task:
(i) limited generalization to photorealistic images, where models struggle to maintain fidelity and realism on diverse real-world scenes;
(ii) ambiguity in layer granularity, arising from the ill-posed nature of layer definitions and the absence of clear ground-truth separation;
(iii) occluded layer completion, which remains difficult when layered occlusions involve semi-transparent or complex blending; and
(iv) background inpainting, where reconstructing plausible unseen regions is challenging under severe occlusion. We visualize representative failure cases in Fig.~\ref{fig:i2l_failures}.
The principal causes of failures in occluded layer completion are twofold: on the one hand, the layout detector may fail to predict accurate amodal bounding regions for occluded layers; on the other hand, the image-to-layer generation model may not faithfully reconstruct complex occluded pixels due to insufficient contextual cues and data diversity. These limitations highlight avenues for future research.

\begin{table}[t]
    \begin{minipage}[t]{1\linewidth}
    \centering
    \tablestyle{5pt}{1.2}
    \resizebox{1.0\linewidth}{!}
    {
    \begin{tabular}{l|ccc|ccc}
     & \multicolumn{3}{c|}{PSNR$\scriptstyle\text{merged}$ $\uparrow$} & \multicolumn{3}{c}{SSIM$\scriptstyle\text{merged}$ $\uparrow$} \\
    \cline{2-7}
    Layers & $[4,8)$ & $[8,16)$ & $[16,32)$ & $[4,8)$ & $[8,16)$ & $[16,32)$ \\
    \shline
    MRT (Ours) & 27.3440 & 25.9068 & 25.7229 & 0.9034 & 0.8762 & 0.8485 \\
    Qwen-Image-Layered & 25.8111 & 23.0645 & 22.1828 & 0.8706 & 0.8319 & 0.8065 \\
    \end{tabular}
    }
    \vspace{-2mm}
    \caption{
    \footnotesize{Comparison with Qwen-Image-Layered on the image-to-layers.}
    }
    \label{tab:i2l_qwen_compare}
    \end{minipage}
    \vspace{-5mm}
\end{table}

\begin{figure*}[!t]
\centering
\includegraphics[width=1\linewidth]{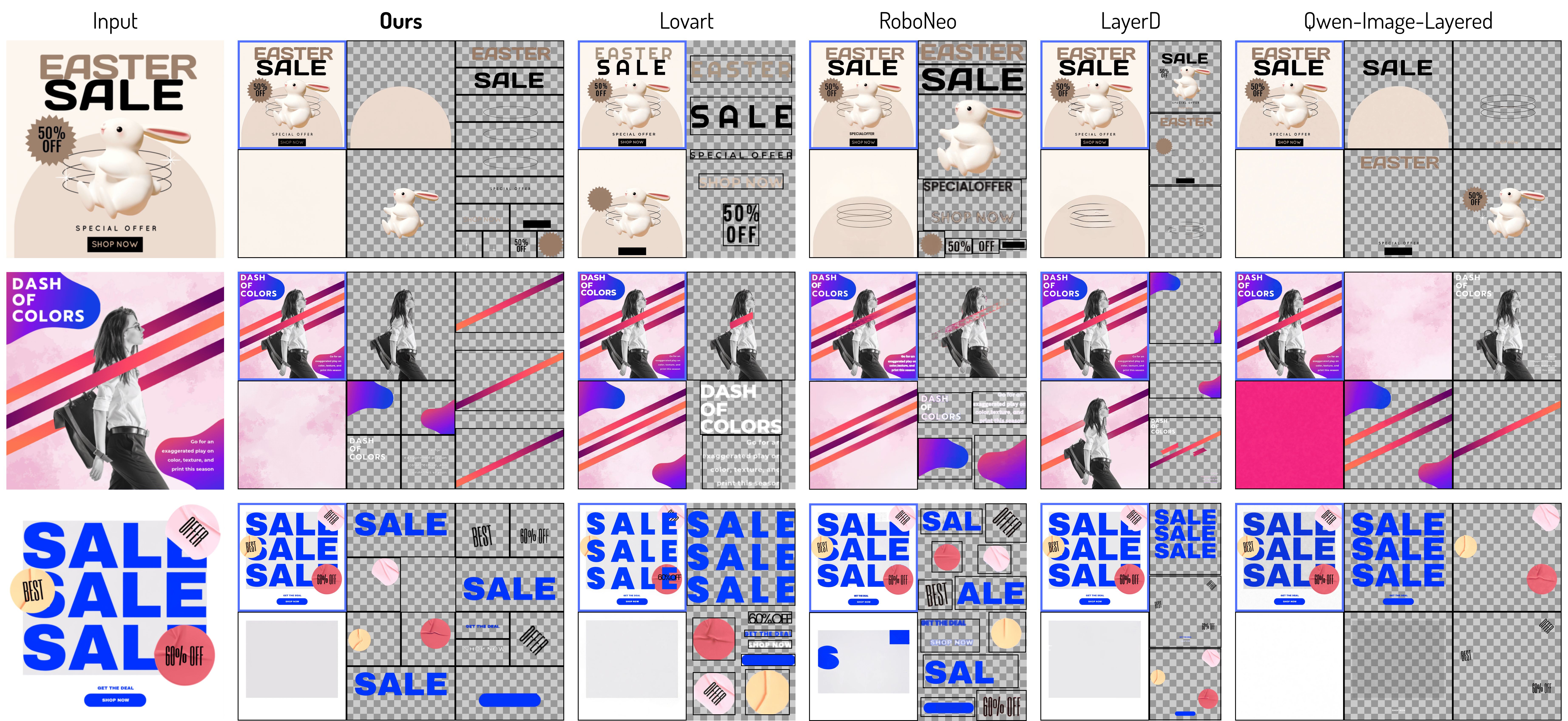}
\caption{\footnotesize{\textbf{Image-to-layers comparison.} Each panel's top-left shows the composed image with decomposed layers. Our method outperforms all baselines. Lovart shows poor decomposition quality, RoboNeo exhibits artifacts, LayerD and Qwen-Image-Layered produce overly grouped layers. Top-left: composed image with layers. (Best viewed zoomed in)}}
\vspace{-4mm}
\label{fig:i2l_qual_layerd}
\end{figure*}

\subsubsection{Layers-to-Layers: Layered Editing}
To the best of our knowledge, no prior work has studied the task of layered image editing.
To establish a comparison for this task, we instantiate a baseline using GPT-Image-1, which supports multi-conditional image inputs and transparent RGBA layer outputs.
We report results for our approach on two key tasks, detail how GPT-Image-1 is configured as a competitive baseline, and highlight the distinctive properties of our method.

\begin{figure}[!t]
\centering
\includegraphics[width=\linewidth]{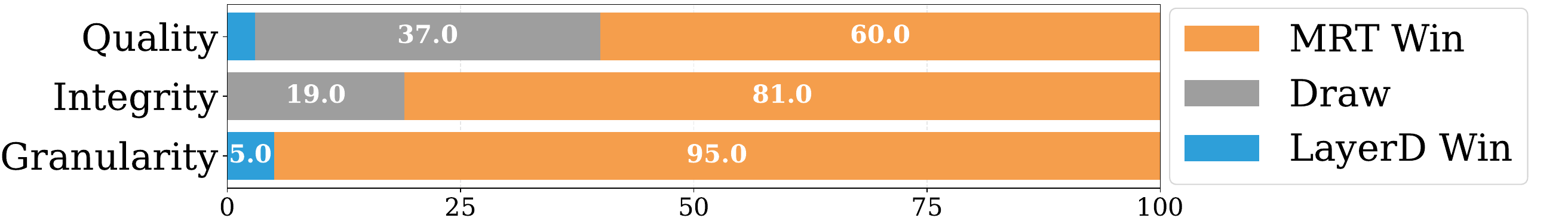}
\includegraphics[width=\linewidth]{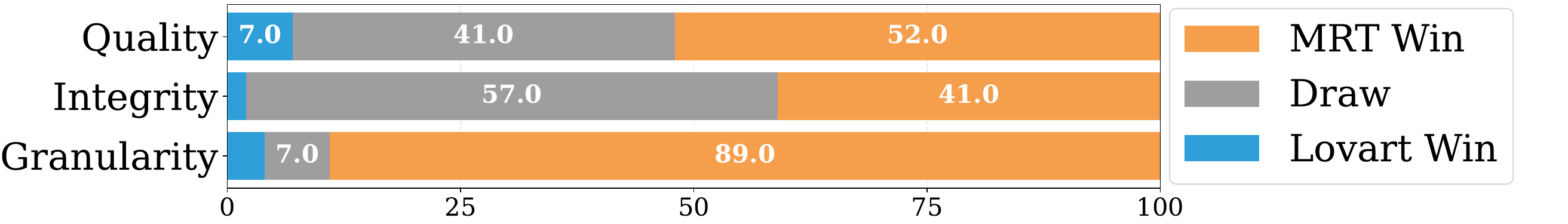}
\includegraphics[width=\linewidth]{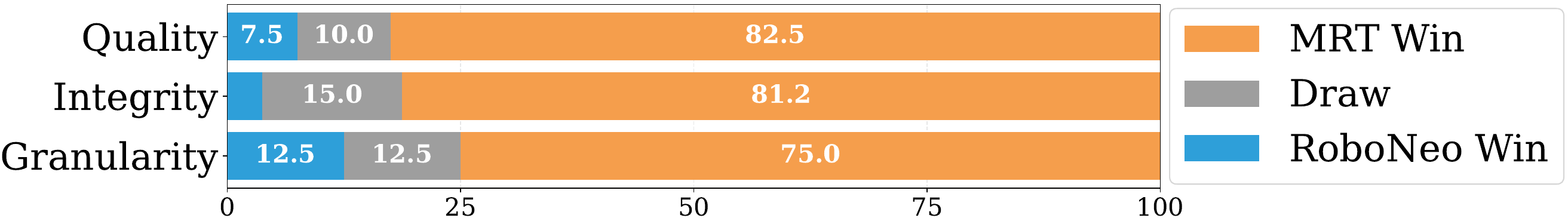}
\vspace{-5mm}
\caption{\footnotesize{\textbf{Comparison with SOTA and commercial systems on image-to-layers.} We conduct a blind user study where participants select the better result from paired samples. Blind user study shows our method significantly outperforms LayerD and commercial systems (Lovart, RoboNeo). Participants evaluate the results from three aspects including (i) {Quality}: semantic correctness and transparency, (ii) {Integrity}: faithful reconstruction of the input, and (iii) {Granularity}: appropriate decomposition level—avoiding overly grouped layers. Our approach demonstrates significant advantages across all evaluation dimensions according to user study.}}
\vspace{-5mm}
\label{fig:compete_i2l_user_study}
\end{figure}

\begin{figure*}[!t]
\centering
\includegraphics[width=0.95\textwidth]{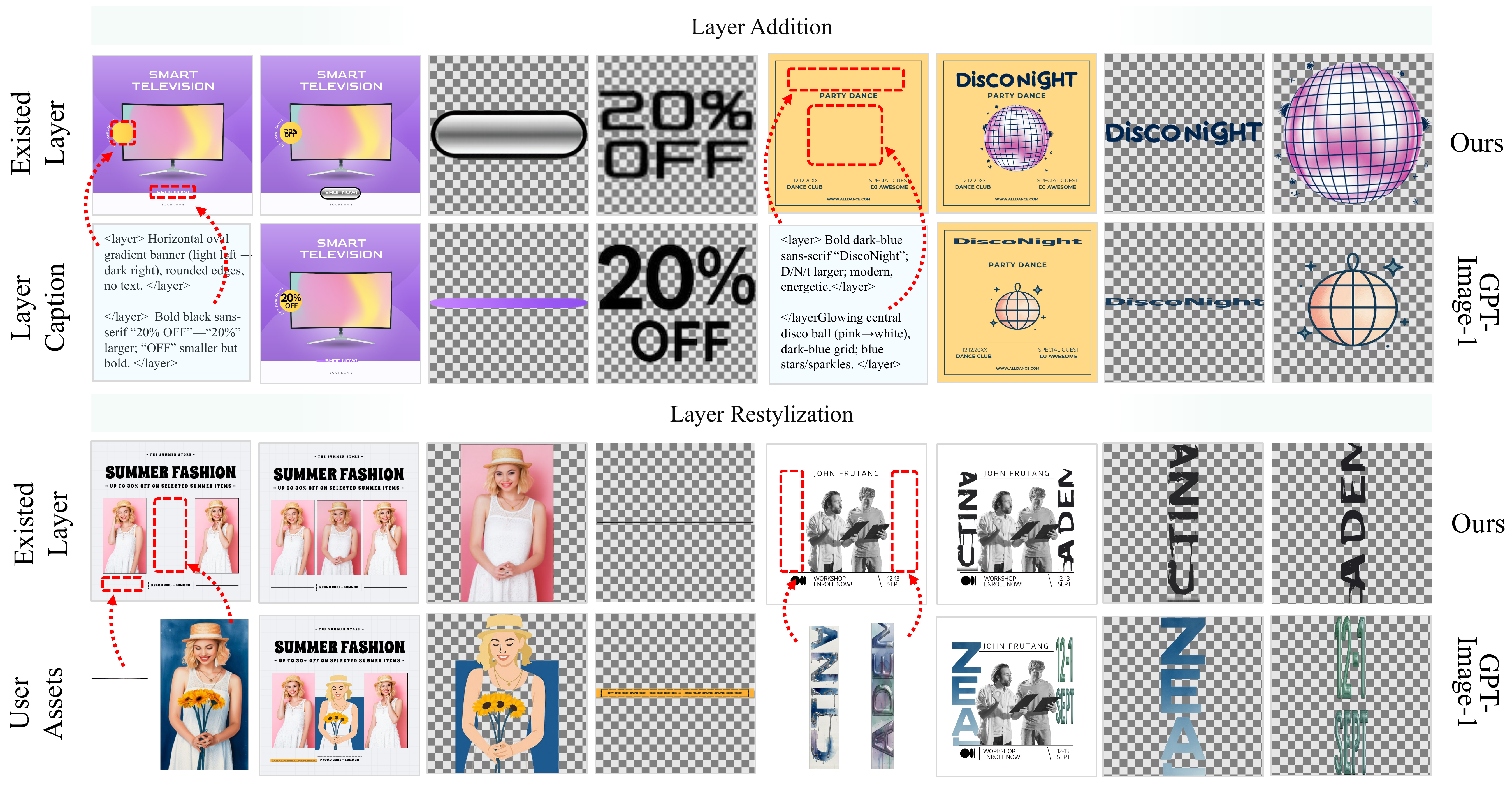}
\vspace{-3mm}
\caption{\footnotesize{\textbf{Qualitative comparison on layers-to-layers.} Layer addition (first two rows) and layer restylization (last two rows). For layer addition, our approch also better follow the layer-wise instructions than GPT-Image-1. For layer resylization, our method also outperforms GPT-Image-1 in terms of layer coherence and style consistency. The layers-to-layers task enables flexible user interaction with the generative model through iterative layer-wise editing.}}
\label{fig:l2l_add_qual}
\vspace{-3mm}
\end{figure*}

\vspace{1mm}
\noindent\textbf{Layer Addition.}
Layer Addition aims to insert new layers into an existing design conditioned on layer-wise captions. In this comparison, we simulate the user by providing two target bounding boxes on the template together with the corresponding layer-wise captions. Our model predicts the requested layers in parallel while maintaining cross-layer consistency.
For GPT-Image-1, we adopt an iterative generation procedure. We condition on the current composite image, draw red bounding boxes at the insertion locations, and input the corresponding layer-wise caption to GPT-Image-1, which outputs a transparent RGBA layer. We then insert the generated layer at the specified position and iterate the process for the remaining layers.
By generating multiple layers in single pass and conditioning on all layers, our method better captures inter-layer relationships and produces coherent insertions that preserve global composition and style in Fig.~\ref{fig:l2l_add_qual} and outperforms GPT-Image-1.

\vspace{1mm}
\noindent\textbf{Layer Restylization.}
For restylizing target layers, the user provides assets to be placed on the canvas; we restylize these assets into layers that harmonize with the overall composition.
For GPT-Image-1, we provide multi-image inputs: the merged image of existing layers annotated with a red bounding box to indicate the insertion location, together with the user-specified asset. After predicting one layer, we insert it at the specified position and iterate for the remaining targets.
Our method harmonizes all selected layers in a single pass, whereas GPT-Image-1 requires layer-by-layer generation, which increases latency and may propagate inconsistencies across multiple edits. Fig.~\ref{fig:l2l_add_qual} shows that our edits better preserve geometry while adapting appearance to the target style.

\begin{figure}[!t]
\centering
\includegraphics[width=0.9\linewidth]{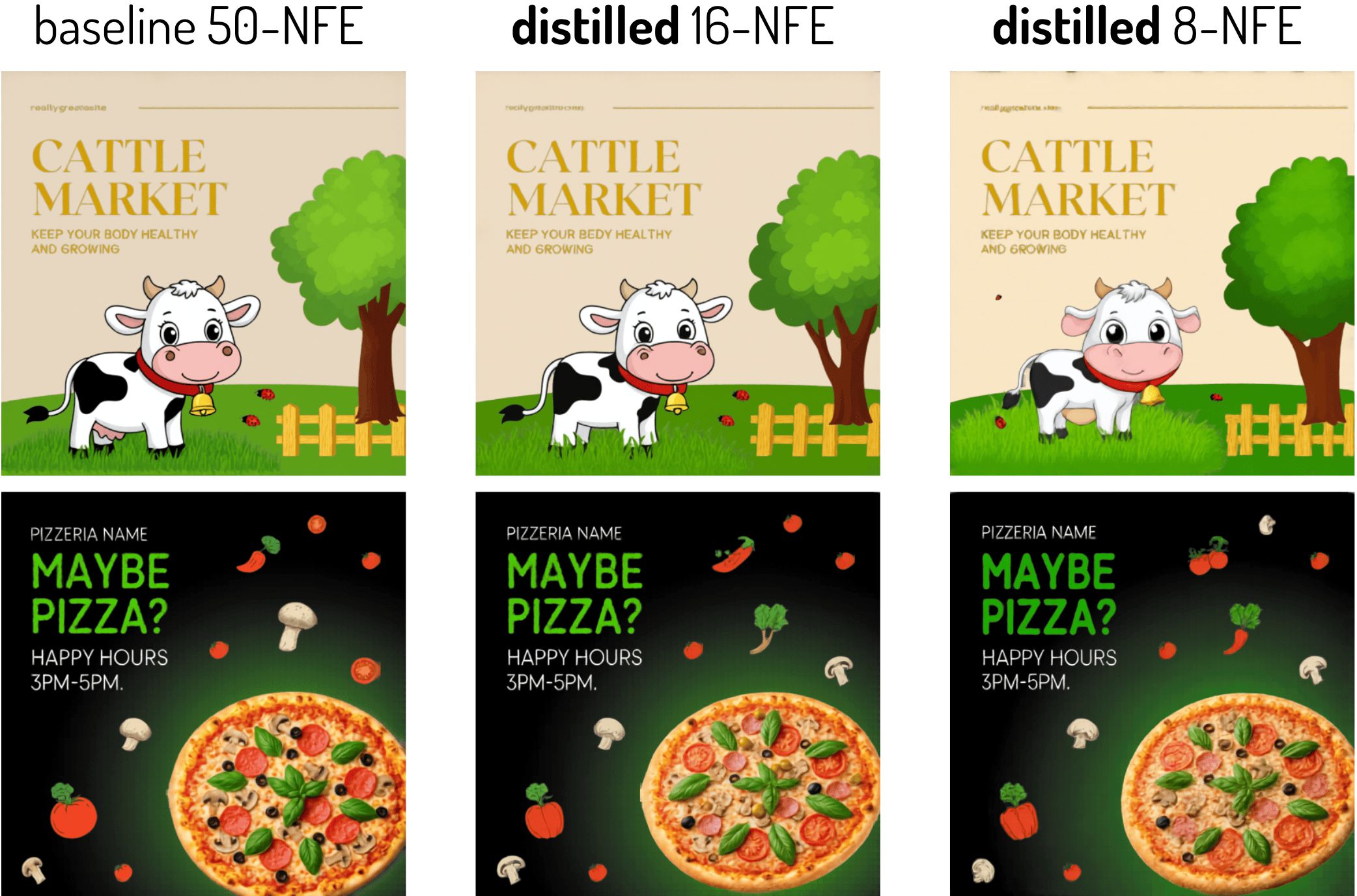}
\vspace{-1mm}
\caption{\footnotesize{Comparison between baseline and few-step distilled model.}}
\label{fig:steps_speed}
\end{figure}

\subsection{Ablation Study and Analaysis}

\vspace{1mm}
\noindent\textbf{Larger models and dataset improve quality.}
To demonstrate the importance of model and dataset scaling, we train text-to-layers models using FLUX.1 [dev] (13B) and Qwen-Image (20B) on the same 0.5M-sample dataset. Model scaling alone reduces FID from $17.79$ to $16.15$. Subsequently scaling the dataset to 10M samples further reduces FID to $15.63$ under a limited training budget, with additional gains expected from extended training. These results confirm that both model capacity and dataset scale are essential for high-quality generation.

\begin{table}[t]
\begin{minipage}[t]{1\linewidth}
\centering
\tablestyle{10pt}{1.2}
\resizebox{0.99\linewidth}{!}
{
\begin{tabular}{l|c|cc}
Training Data & FID$\scriptstyle \text{merged}$ $\downarrow$ & PSNR${_{\text{merged}}}$ $\uparrow$ & SSIM${_{\text{merged}}}$ $\uparrow$ \\
\shline
w/o overflow data & 15.68 & 21.81 & 0.8543 \\
w/ overflow data & 16.15 & 22.75 & 0.8711 \\
\end{tabular}
}
\vspace{-2mm}
\caption{
\footnotesize{\textbf{Overflow support on text-to-layers (T2L).}}
}
\label{tab:abl_overflow}
\end{minipage}
\begin{minipage}[t]{1\linewidth}
\centering
\tablestyle{5pt}{1.2}
\resizebox{0.99\linewidth}{!}
{
\begin{tabular}{l|c|c|ccc}
method & task mix ratio & FID$\scriptstyle \text{merged}$ $\downarrow$ & PSNR${_{\text{merged}}}$ $\uparrow$ & SSIM${_{\text{merged}}}$ $\uparrow$ \\
\shline
T2L & 100\%~/~0\%~/~0\% & 16.15 & 22.75 & 0.8711  \\
T2L+I2L & 80\%~/~20\%~/~0\% & 15.68 & 23.06 & 0.8924 \\
T2L+I2L+L2L & 70\%~/~15\%~/~15\% & 17.06 & 21.97 & 0.8606 \\
\end{tabular}
}
\caption{
\footnotesize{\textbf{Multiple task training}. T2L: text-to-layers. I2L: image-to-layers. L2L: layers-to-layers.}
}
\label{tab:ablation:unify_task}
\end{minipage}
    \begin{minipage}[t]{1\linewidth}
    \centering
    \tablestyle{5pt}{1.2}
    \resizebox{0.99\linewidth}{!}
    {
    \begin{tabular}{l|cc|cc}
    method & PSNR${_{\text{merged}}}$ $\uparrow$ & SSIM${_{\text{merged}}}$ $\uparrow$ & PSNR${_{\text{layer}}}$ $\uparrow$ & SSIM${_{\text{layer}}}$ $\uparrow$ \\
    \shline
    w/o text condition & 21.27 & 0.8697 & 26.03 & 0.9794 \\
    w/ text condition & 21.65 & 0.8805 & 27.24 & 0.9846 \\
    \end{tabular}
    }
    \vspace{-2mm}
    \caption{
    \footnotesize{{\textbf{Text condition on image-to-layers (I2L) task}}.}
    }
    \label{tab:abl_i2l_cap_predbox}
    \end{minipage}
    \begin{minipage}[t]{1\linewidth}
    \centering
    \tablestyle{5pt}{1.2}
    \resizebox{0.99\linewidth}{!}
    {
    \begin{tabular}{l|cc|cc}
   method & PSNR${_{\text{merged}}}$ $\uparrow$ & SSIM${_{\text{merged}}}$ $\uparrow$ & PSNR${_{\text{layer}}}$ $\uparrow$ & SSIM${_{\text{layer}}}$ $\uparrow$ \\
    \shline
    w/o merge aug. & 21.65 & 0.8805 & 27.24 & 0.9846 \\
    w/ merge aug. & 21.97 & 0.8864 & 26.96 & 0.9840 \\
    \end{tabular}
    }
    \vspace{-2mm}
    \caption{
    \footnotesize{\textbf{Layer grouping augmentation on image-to-layers (I2L) task}.}
    }
    \label{tab:abl_i2l_merge_aug_predbox}
    \end{minipage}
    \begin{minipage}[t]{0.99\linewidth}
    \centering
    \tablestyle{22pt}{1.2}
    \resizebox{1\linewidth}{!}
    {
    \begin{tabular}{l|cc}
    method & Denoise steps & FID$\scriptstyle \text{merged}$ $\downarrow$ \\
    \shline
    Baseline & 50 & 16.02 \\
    + DMD2 Distillation & 16 & 16.21 \\
    + DMD2 Distillation & 8 & 18.58 \\
    \end{tabular}
    }
    \vspace{-2mm}
    \caption{\footnotesize{\textbf{Multi-layer generator distillation.}}}
    \label{tab:few_step_sampling}
    \end{minipage}
    \vspace{-6mm}
\end{table}

\vspace{1mm}
\noindent\textbf{Overflow support w/o performance loss.}
Table~\ref{tab:abl_overflow} evaluates the impact of overflow-aware generation. Over 60\% of designs contain overflow layers while previous works all truncate these elements, severely limiting editability and reusability. Training with overflow data enables complete layer generation with minimal performance cost: our model achieves comparable FID, PSNR, and SSIM scores while uniquely preserving overflow elements.

\vspace{1mm}
\noindent\textbf{Multi-task training and performance trade-offs}
Table~\ref{tab:ablation:unify_task} shows unified multi-task training with random task sampling. Our framework integrates all three tasks without multi-stage fine-tuning while maintaining comparable performance across configurations, demonstrating minimal degradation from unification. We observe that introducing the layers-to-layers task slightly reduces overall performance, which we attribute to layer-to-layer dataset quality issues—a direction we leave for future work.

\vspace{1mm}
\noindent\textbf{Textual conditioning is not essential for image-to-layers.}
An important question is whether global captions are necessary for image-to-layers decomposition.
Table~\ref{tab:abl_i2l_cap_predbox} ablates caption conditioning and shows modest but consistent improvements across metrics. This reveals a noteworthy finding: while textual guidance aids boundary disambiguation and provides semantic cues for complex overlapping compositions, it is not essential for our framework.

\vspace{1mm}
\noindent\textbf{Layer grouping augmentation improves robustness.}
Table~\ref{tab:abl_i2l_merge_aug_predbox} validates layer grouping augmentation. Since our framework requires layout inputs, a distribution gap exists between precise training layout annotations and noisy test-time layouts from users or detectors. We address this by randomly merging layers during training to increase layout diversity. This strategy yields consistent improvements even on \designbenchmark with high-quality layout annotations, with larger gains expected under noisy layout conditions.

\vspace{1mm}
\noindent\textbf{Distilled multi-layer generator brings significant acceleration.}
By incorporating DMD2 distillation~\cite{yin2024one,yin2024improved}, we accelerate our multi-layer generation from $50$ to $8$ denoising steps, achieving a $6\times$ speedup with minimal performance degradation. FID scores remain comparable in Table~\ref{tab:few_step_sampling} and visual quality is largely preserved in Fig.~\ref{fig:steps_speed}, demonstrating the effectiveness of distillation for few-step generation in multi-layer image diffusion models.

\vspace{1mm}
\noindent\textbf{Additional ablations.}
We provide additional ablation studies on caption length, multilingual design generation, and fine-tuning with PrismLayers data in the supplementary material.
\section{Conclusion}
\label{conclusion}
In this paper, we have presented the first systematic study examining the performance frontier of multi-layer transparent image generation at scale.
We introduced the Masked Region Transformer, a large-scale diffusion framework that unifies text-to-layers, image-to-layers, and layers-to-layers generation within a shared masked region paradigm. Trained on over 10M multilingual design samples, our 20B-parameter model incorporates key technical innovations: an overflow-aware canvas layer for complete boundary handling, and distribution matching distillation for real-time generation. Together, these contributions enable efficient synthesis of high-fidelity, semi-transparent, fully editable visual layers.

{
    \small
    \bibliographystyle{ieeenat_fullname}
    \bibliography{main}
}

\clearpage
\setcounter{page}{1}
\maketitlesupplementary

\setcounter{section}{0}
\setcounter{figure}{0}
\setcounter{table}{0}

\section{Additional ablation experiments}
\label{sec:supp_ablation}

\begin{table}[t]
    \begin{minipage}[t]{1\linewidth}
    \centering
    \tablestyle{30pt}{1.2}
    \resizebox{1.0\linewidth}{!}
    {
    \begin{tabular}{l|cc}
    \multirow{2}{*}{Training Caption Length} & \multicolumn{2}{c}{FID$\scriptstyle\text{merged}$ $\downarrow$} \\
    \cline{2-3}
     & Short Cap. & Long Cap. \\
    \shline
    Short Cap. & 17.64 & 18.56 \\
    Long Cap. & 17.95 & 16.15 \\
    Mixed (50\% short + 50\% long) & 16.13 & 15.93 \\
    \end{tabular}
    }
    \vspace{-2mm}
    \caption{
    \footnotesize{Effect of caption length during training. We train models with short captions, long captions, or a mixture of both, and evaluate FID on VC5K test set using short and long captions respectively.}
    }
    \label{tab:abl_caption_length}
    \end{minipage}
\end{table}

\begin{table}[t]
    \begin{minipage}[t]{1\linewidth}
    \centering
    \tablestyle{20pt}{1.2}
    \resizebox{1.0\linewidth}{!}
    {
    \begin{tabular}{l|cccc}
    method & Denoise steps & Latency(s) & Speed up & FID$\scriptstyle \text{merged}$ $\downarrow$ \\
    \shline
    Baseline & 50 & 14.4 & - & 16.02 \\
    + Distill & 16 & 4.5 & 3.2x & 16.21 \\
    + Distill & 8 & 2.3 & 6.26x & 18.58 \\
    \end{tabular}
    }
    \vspace{-2mm}
    \caption{
    \footnotesize{Multi-layer generator distillation with inference time.}
    }
    \label{tab:infer_speed}
    \end{minipage}
\end{table}

\subsection{Mixed Training with Variable Caption Length}
Table~\ref{tab:abl_caption_length} demonstrates the importance of caption diversity during training. Models trained with mixed caption lengths achieve the best generalization, with FID of 16.13 on short captions and 15.93 on long captions. Training exclusively on one caption type creates a domain gap: short-caption-only training degrades to 18.56 FID on long captions, while long-caption-only training achieves 16.15 FID, showing better robustness but still suboptimal on short captions.

\begin{table}[t]
    \begin{minipage}[t]{1\linewidth}
    \centering
    \tablestyle{10pt}{1.2}
    \resizebox{0.99\linewidth}{!}
    {
    \begin{tabular}{l|cccc}
    \#layer numbers & 2$\sim$7 & 8$\sim$11 & 12$\sim$14 & 15$\sim$50 \\
    \shline
    PSNR${_{\text{merged}}}$ $\uparrow$ & 22.51 & 21.99 & 21.36 & 20.65 \\
    SSIM${_{\text{merged}}}$ $\uparrow$ & 0.8932 & 0.8869 & 0.8780 & 0.8610 \\
    \end{tabular}
    }
    \vspace{-2mm}
    \caption{
    \footnotesize{Effect of layer numbers on image-to-layers (I2L) generation quality. We evaluate the model's performance across different ranges of layer numbers in the generated results.}
    }
    \label{tab:abl_layer_numbers_i2l}
    \end{minipage}
\end{table}

\subsection{Effect of Layer Numbers on Image-to-layer}
Table~\ref{tab:abl_layer_numbers_i2l} demonstrates our method's scalability across different layer counts for the image-to-layers task. The model handles compositions ranging from 2 to 50 layers effectively, maintaining stable performance across this wide range. This flexibility enables decomposition of both simple designs and complex multi-element compositions without architectural modifications.

\subsection{Analysis of Distilled Models}

\begin{figure*}[t]
    \centering
    \vspace{-2mm}
    \includegraphics[width=\linewidth]{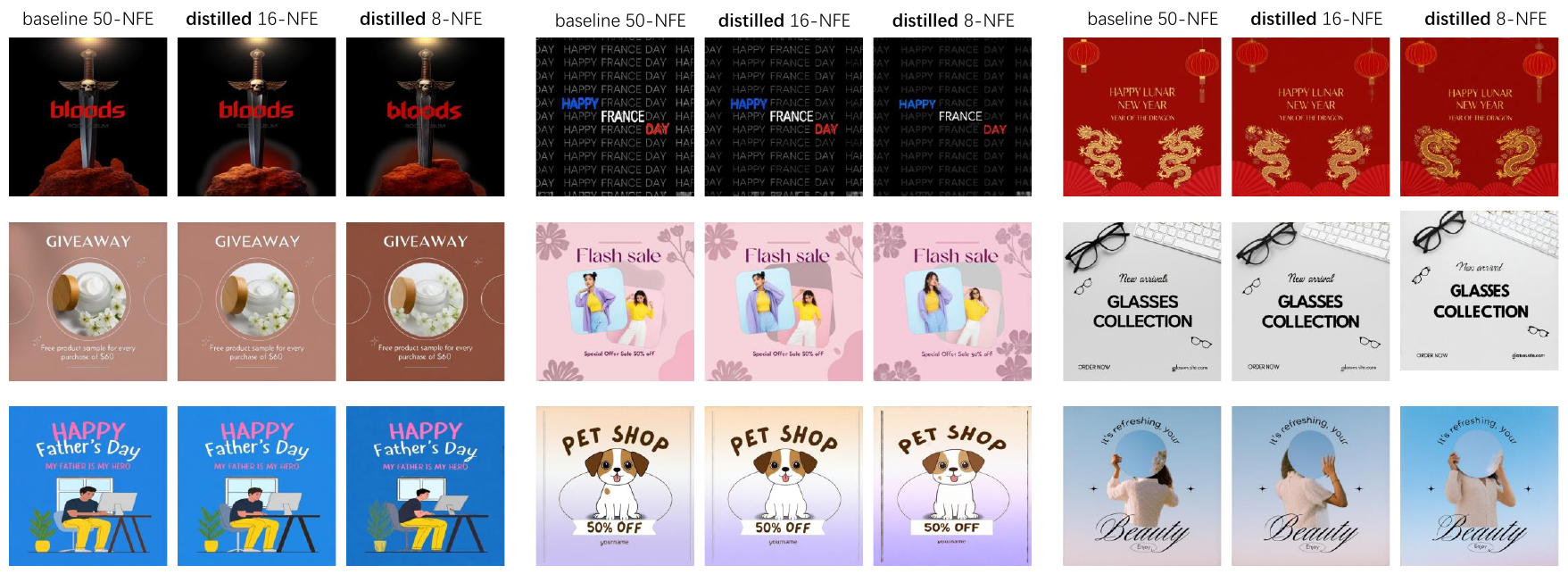}
    \vskip -3mm
    \caption{\footnotesize{\textbf{Generation quality of distilled models.} We achieve up to 6x speed up without sacrificing the quality and fidelity of images.}}
    \label{fig:supp_t2l_dmd}
\end{figure*}

To evaluate the real-world efficiency of our approach, we conducted inference speed benchmarks on a single NVIDIA H200 GPU. We compared the standard baseline method (operating at 50 denoising steps) against our distilled MRT model at reduced inference steps (16 and 8 steps). As shown in Table~\ref{tab:infer_speed}, the baseline model requires 14.4 seconds to complete the generation process. In contrast, applying DMD2 distillation significantly accelerates inference. Specifically, our model achieves a $3.2\times$ speed-up (4.5s) at 16 steps with negligible degradation in generation quality (FID increases only slightly from 16.02 to 16.21). Furthermore, reducing the inference budget to just 8 steps yields a massive $6.26\times$ speed-up (2.3s), showing that our method successfully balances high-fidelity generation with interactive-level latency. We also present the generated samples and compare the original and distilled models in Fig.~\ref{fig:supp_t2l_dmd}.

\section{Attention Analysis of Image-to-Layer Model}
\begin{figure*}[t]
    \centering
    \includegraphics[width=0.75\linewidth]{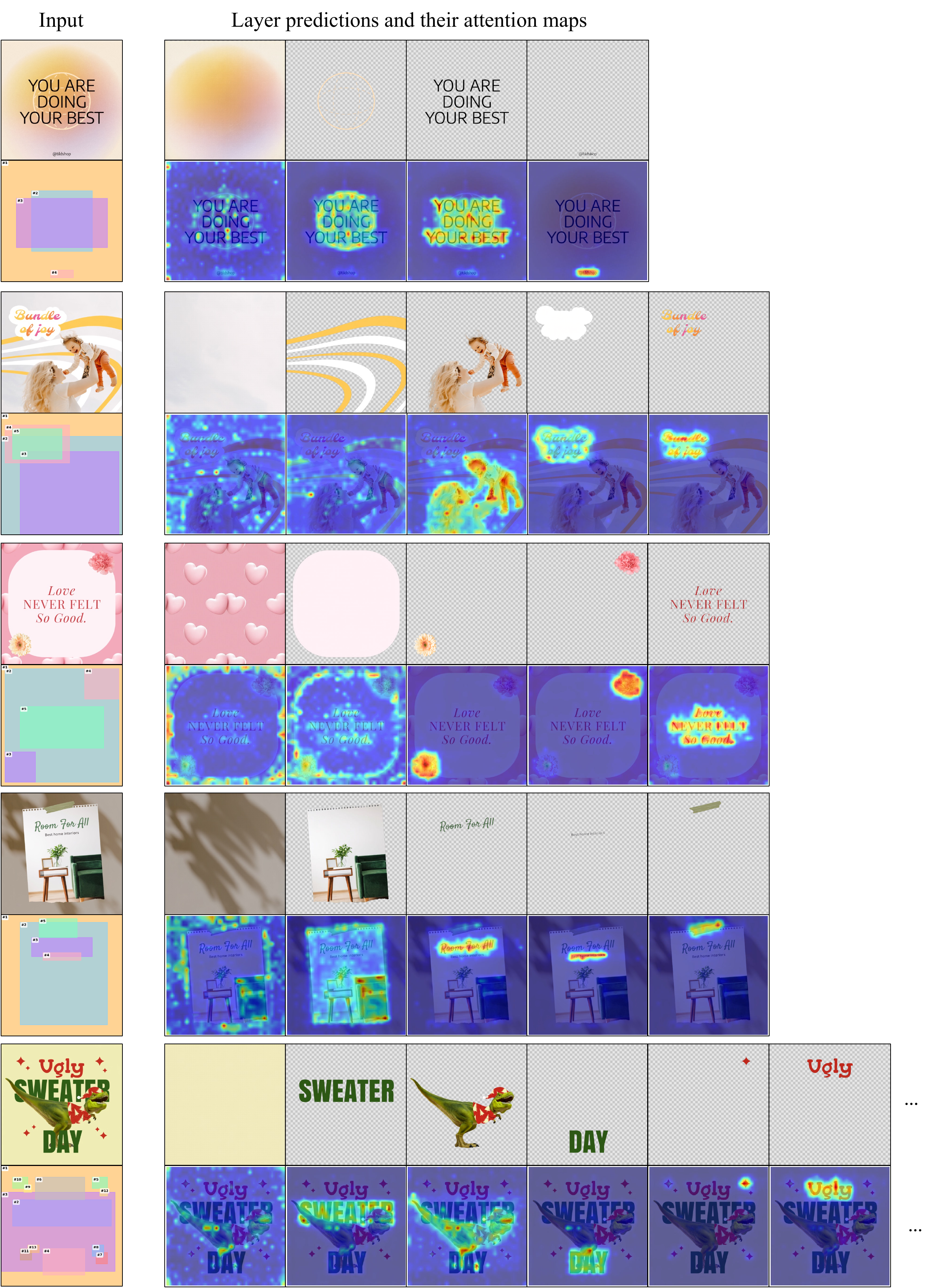}
    \vspace{-2mm}
    \caption{\footnotesize{\textbf{Attention map visualizations of image-to-layers task.} We demonstrate the interpretability of our model by visualizing the internal attention weights during the layer generation process. Left: The input composite image and its corresponding layout. Right: The decomposition results. The top row displays the predicted transparent layers, while the bottom row shows the corresponding attention maps overlaid on the input image. Red regions indicate high activation values. The results highlight that the model’s attention is semantically selective, accurately aligning with specific visual elements (\emph{e.g.}, text, foreground objects, background patterns) to generate high-quality, disentangled layers.}}
    \label{fig:supp_i2l_attn}
\end{figure*}

To validate that our model learns meaningful semantic representations rather than merely memorizing layout priors, we visualize the pixel-wise attention maps generated during the decomposition process. Fig.~\ref{fig:supp_i2l_attn} illustrates the correspondence between the generated transparent layers and their associated attention activations. As observed, the attention mechanism exhibits strong spatial localization capabilities. For each predicted layer, the attention weights (visualized as heatmaps) highly correlate with the semantic boundaries of the target elements. For instance, when reconstructing high-frequency components such as text (\emph{e.g.}, ``Bundle of Joy'' in the second case, ``Love NEVER FELT...'' in the third one) or fine-grained graphical elements, the attention is tightly focused on the relevant character strokes and shapes, effectively suppressing background noise. Conversely, for background patterns or larger geometric shapes, the attention acts more broadly to capture the texture and spatial extent of the region. This visualization confirms that the model successfully disentangles the composite image by attending to distinct visual features guided by the layout, ensuring that the resulting RGBA layers possess clean alpha mattes and coherent textures.

\section{User study details}

\subsection{User Study on Text-to-Layer Task}
To evaluate the generation quality of our models on the \texttt{text-to-layer} task, we conducted a user study comparing our method (MRT) with the baseline (ART). We employed a blind, pairwise comparison setup. For each sample, participants were first shown the input text prompt, followed by the corresponding results generated by MRT and ART displayed side-by-side. To eliminate positional bias, the display order (left or right) of these two results was randomized for each evaluation. Participants were asked to cast a three-way forced-choice vote—"Method A is better," "Method B is better," or "Tie"—across four distinct dimensions: (1) elements (layout), (2) visual appeal (aesthetics), (3) correctness of the text (typography), and (4) coherence and quality of each layer (harmonization).

\begin{figure*}[t]
    \centering
    \includegraphics[width=0.95\linewidth]{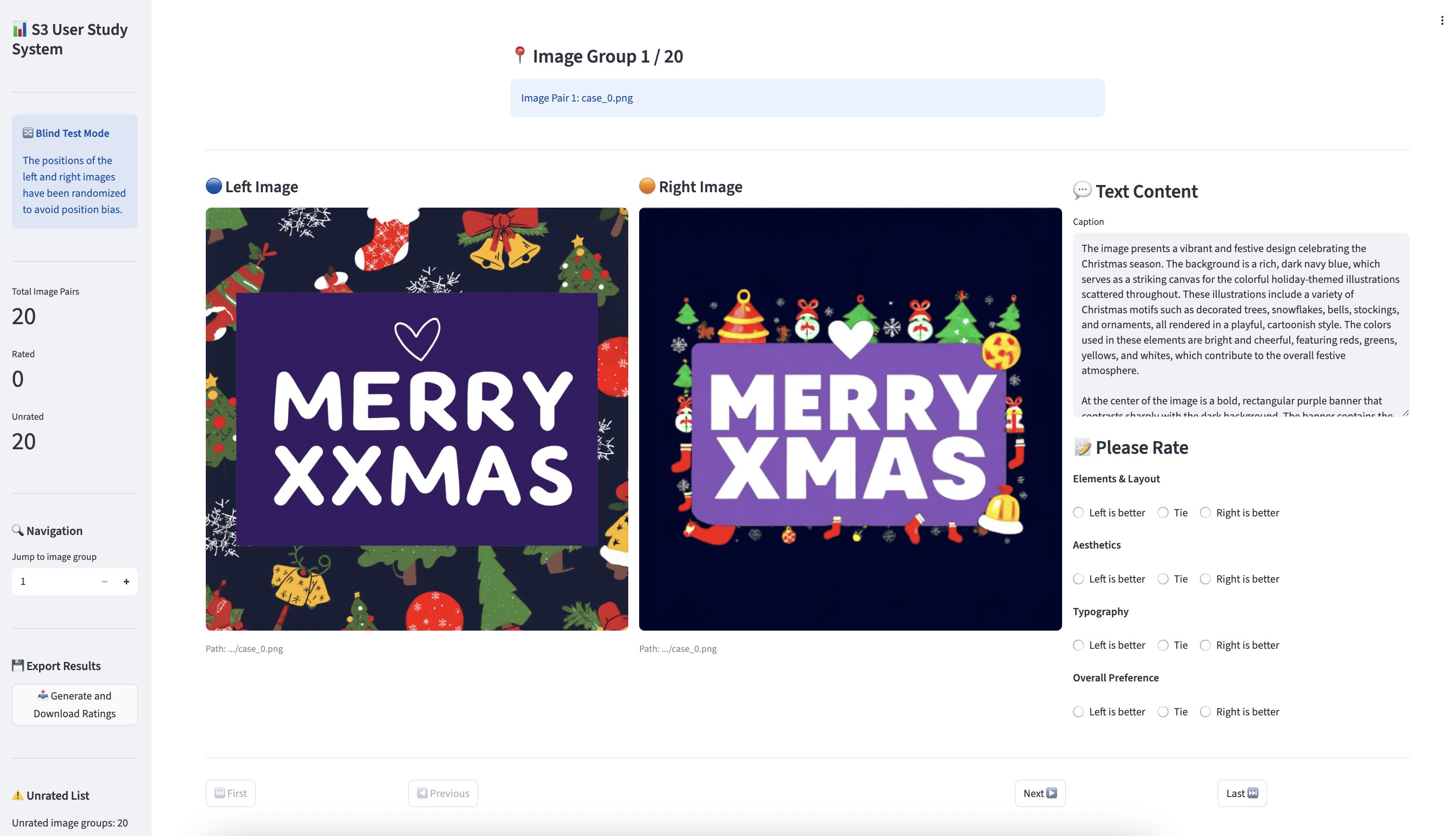}
    \vspace{-2mm}
    \caption{\footnotesize{\textbf{User study interface for text-to-layers evaluation.} Two generated results are displayed side-by-side with the text caption shown on the right. Participants vote across four dimensions: elements (layout), aesthetics, typography, and overall preference.}}
    \label{fig:user_ui_t2l}
\end{figure*}

The web-based evaluation interface is shown in Fig.~\ref{fig:user_ui_t2l}, where two generated results are displayed side-by-side with the text caption provided on the right panel.

\subsection{User Study on Image-to-Layer Task}
For the \texttt{image-to-layers} task, we conducted a comprehensive user study by performing three separate pairwise comparisons between our method and three state-of-the-art baselines: (1) Ours vs. LayerD, (2) Ours vs. Lovart, and (3) Ours vs. Roboneo. Each comparison was run as an independent blind test. Participants in each study were presented with a three-image layout: the original input image was displayed as a central reference, while our method's result and the corresponding baseline's result were shown side-by-side. To eliminate positional bias, the display order (left or right) of our result and the baseline's result was fully randomized in every trial. Participants were asked to make a three-way forced-choice vote ("Method A is better," "Method B is better," or "Tie") based on three key metrics: (1) granularity, (2) layer integrity, and (3) layer quality.

\begin{figure*}[t]
    \centering
    \includegraphics[width=0.95\linewidth]{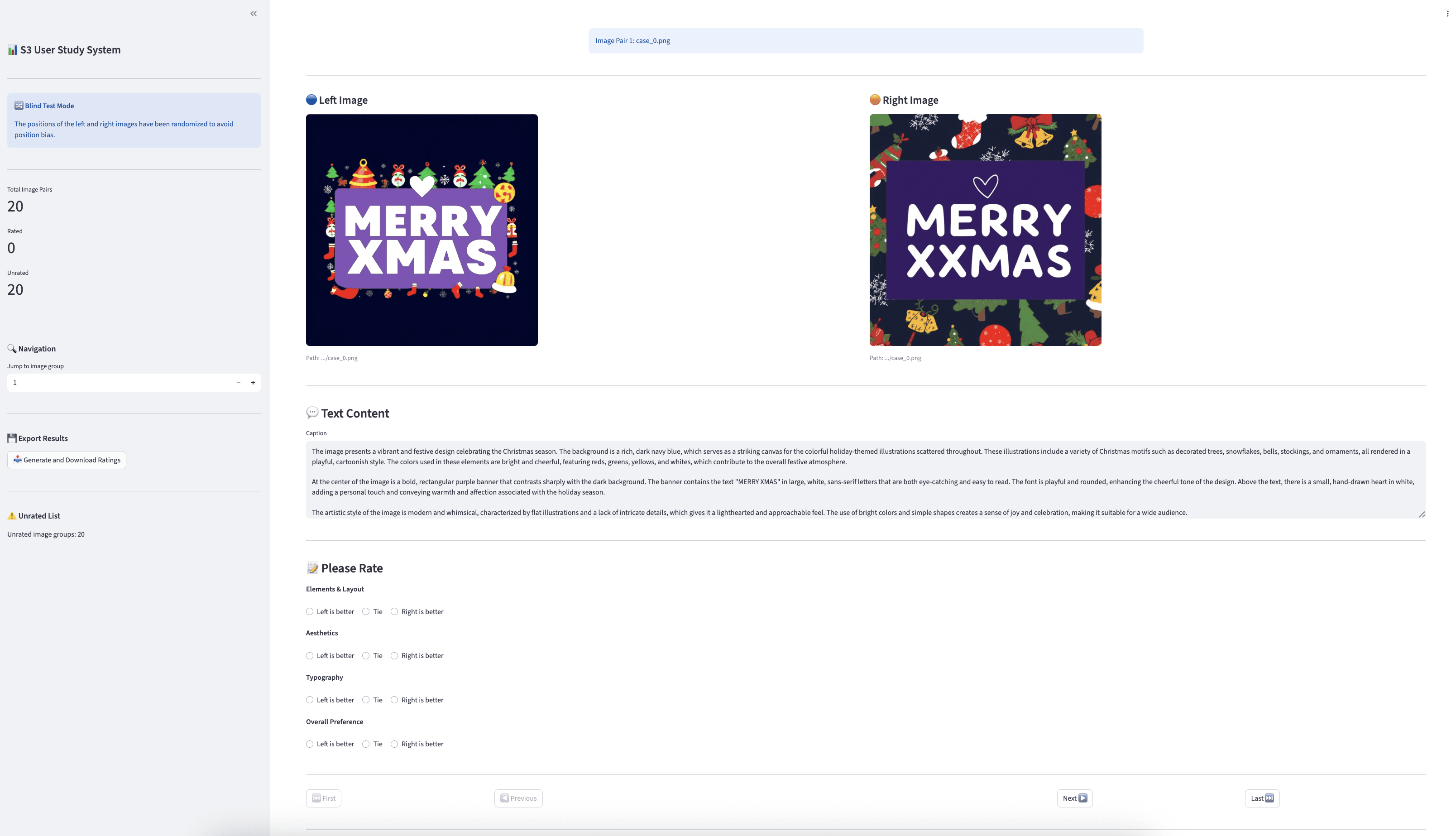}
    \vspace{-2mm}
    \caption{\footnotesize{\textbf{User study interface for image-to-layers evaluation.} The reference input image is displayed at the center with decomposition results from two methods shown on both sides. Participants evaluate based on three metrics: granularity, layer integrity, and layer quality.}}
    \label{fig:user_ui_i2l}
\end{figure*}

The evaluation interface is illustrated in Fig.~\ref{fig:user_ui_i2l}, where the reference input image is shown at the center with decomposition results from two methods displayed on both sides.

\section{Limitations}

Although our model demonstrates strong performance in the image-to-layer task, it faces challenges when applied to real-world photographs. Specifically, our method often fails to correctly handle shadows, resulting in segmented object layers that exclude shadow regions and leaving the shadows on the background layer, which leads to visual inconsistency. We attribute this limitation primarily to our training data: our model was trained exclusively on design datasets, which are planar and lack physical effects such as shadows, reflections, and refractions that commonly appear in natural scenes. Despite this domain gap, we were pleasantly surprised to find that our method can still generalize reasonably well to real images, even without any supervision on real-world multi-layer data. As shown in our illustrations, most objects are successfully separated, which we believe stems from the strong visual understanding capability inherited from the Qwen-Image backbone, demonstrating the robustness, adaptability, and scalability of our approach. In future work, we plan to extend our method to real-world image scenarios by collecting and training on datasets that include realistic visual effects such as shadows and reflections. We believe such extensions will further enhance the model’s ability to produce coherent and physically plausible layer decompositions.

\section{Visualizations and Qualitative Analysis}

\subsection{Diverse Text-to-Layer Generation}
\begin{figure*}[t]
    \centering
    \includegraphics[width=\linewidth]{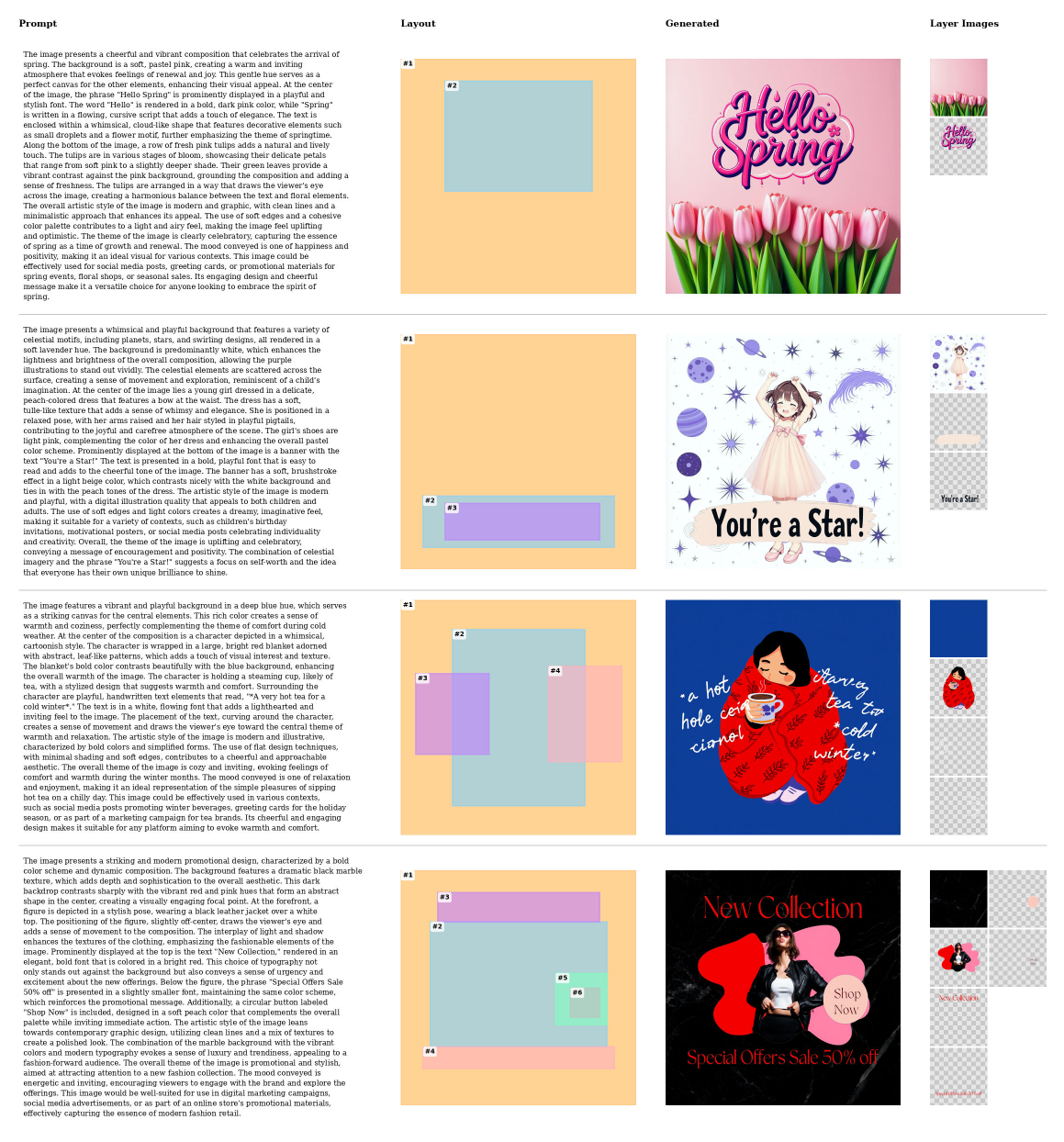}
    \caption{\footnotesize{\textbf{Text-to-layers generation examples.} We visualize diverse text-to-layers generation results from our method, showing the input text prompts and corresponding multi-layer outputs. Each example displays individual transparent RGBA layers along with the merged composition. Our approach generates coherent multi-layer designs that maintain spatial consistency, stylistic harmony, and accurate layer boundaries, demonstrating strong alignment between text prompts and layered compositions.}}
    \label{fig:supp_t2l_vis_1}
\end{figure*}

\begin{figure*}[t]
    \centering
    \includegraphics[width=\linewidth]{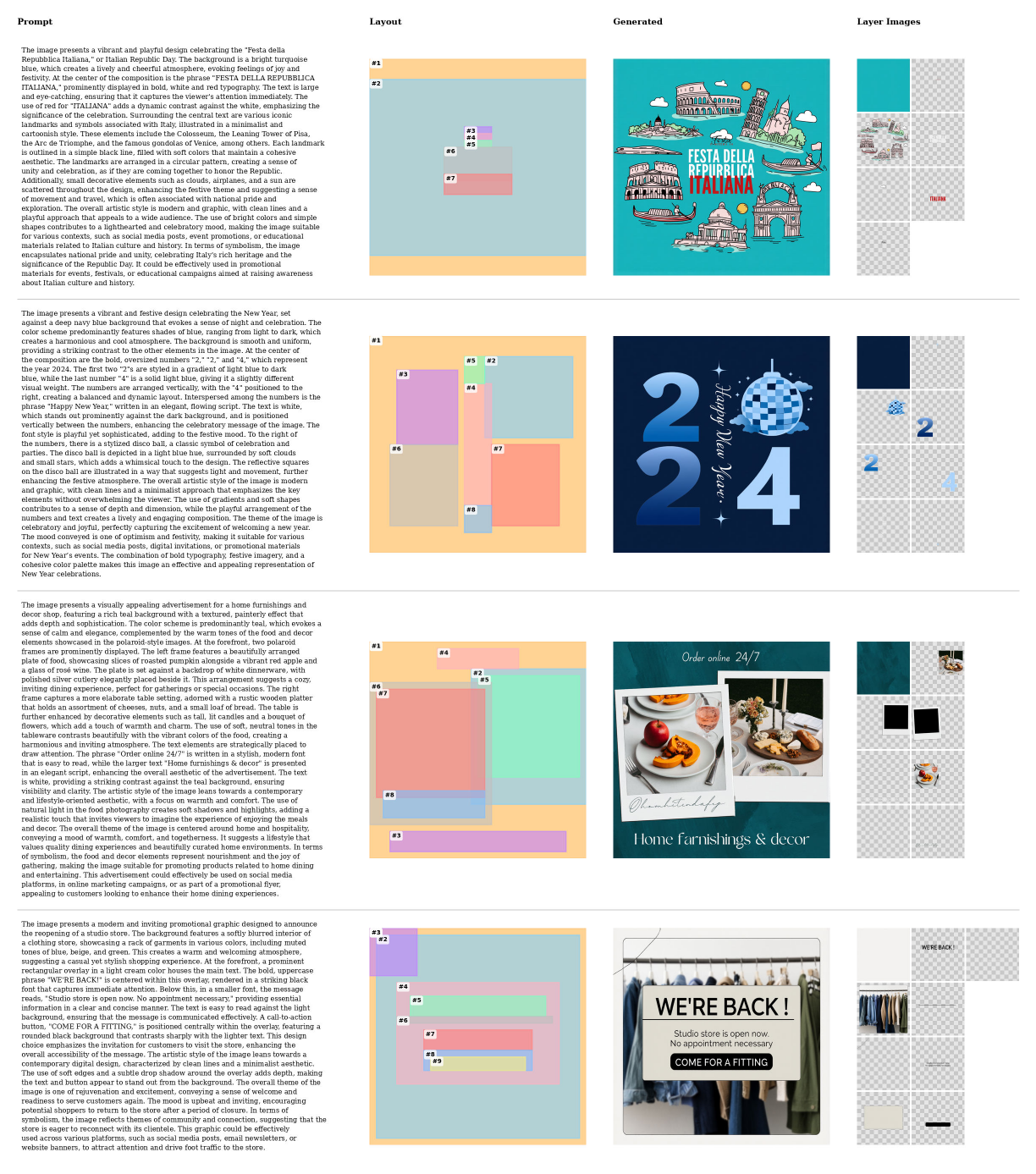}
    \caption{\footnotesize{\textbf{Additional text-to-layers generation examples.} More examples demonstrating our method's capability to generate multi-layer designs from text descriptions. These results showcase the diversity of generated layouts, layer compositions, and visual styles.}}
    \label{fig:supp_t2l_vis_2}
\end{figure*}

\begin{figure*}[t]
    \centering
    \includegraphics[width=\linewidth]{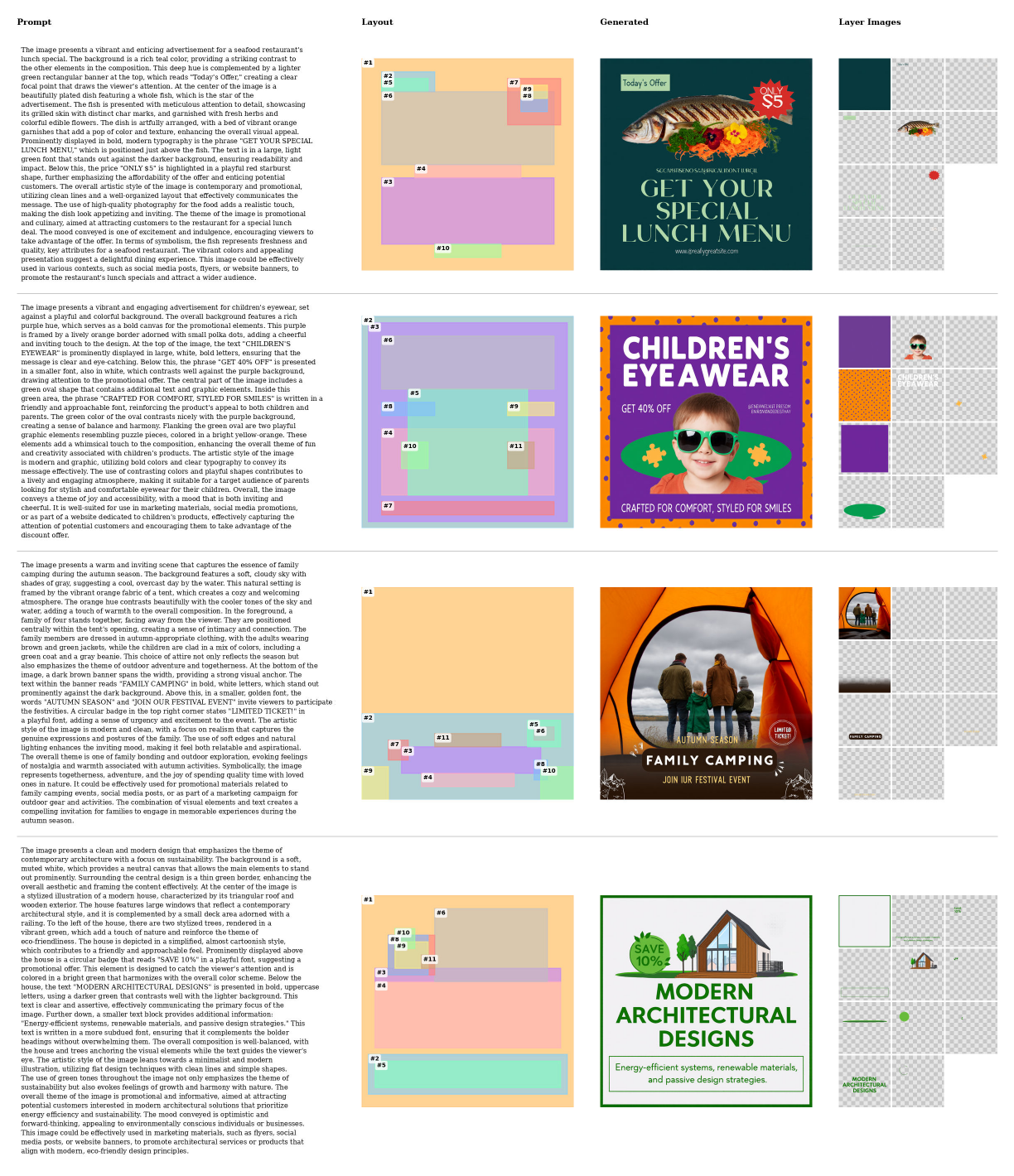}
    \caption{\footnotesize{\textbf{Additional text-to-layers generation examples.}}}
    \label{fig:supp_t2l_vis_3}
\end{figure*}

\begin{figure*}[t]
    \centering
    \includegraphics[width=\linewidth]{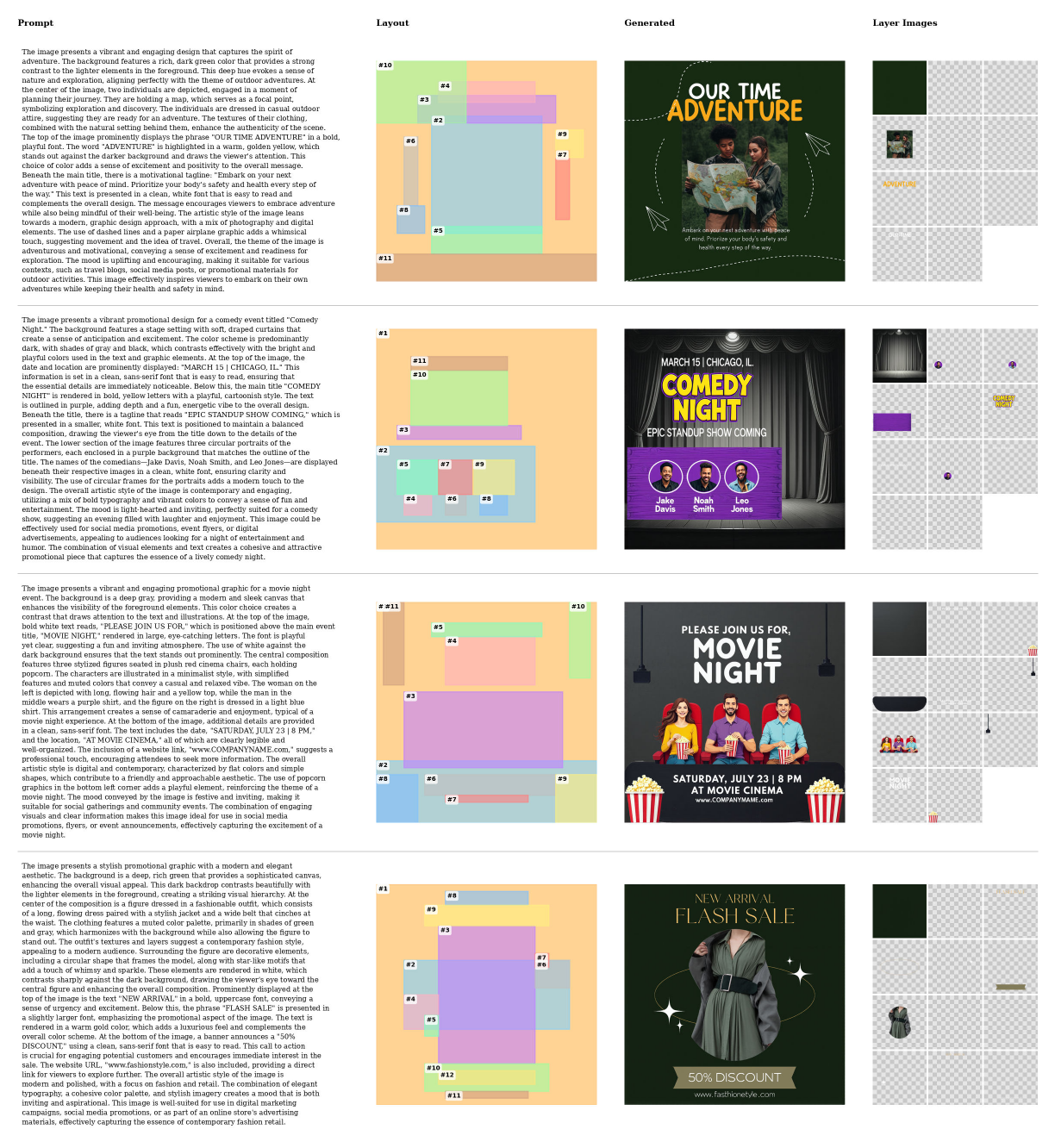}
    \caption{\footnotesize{\textbf{Additional text-to-layers generation examples.}}}
    \label{fig:supp_t2l_vis_4}
\end{figure*}

\begin{figure*}[t]
    \centering
    \includegraphics[width=\linewidth]{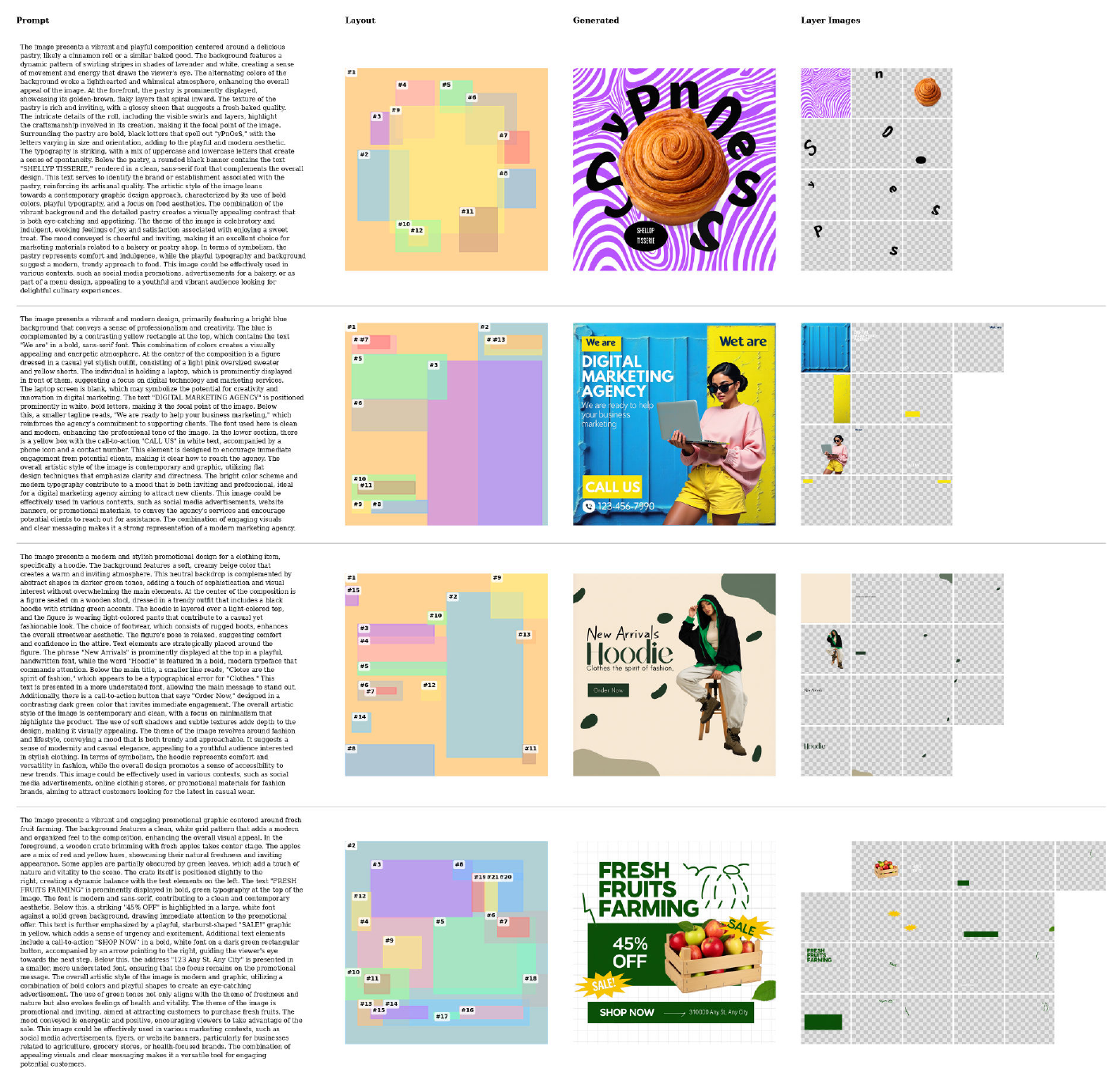}
    \caption{\footnotesize{\textbf{Additional text-to-layers generation examples.}}}
    \label{fig:supp_t2l_vis_5}
\end{figure*}

\begin{figure*}[t]
    \centering
    \includegraphics[width=\linewidth]{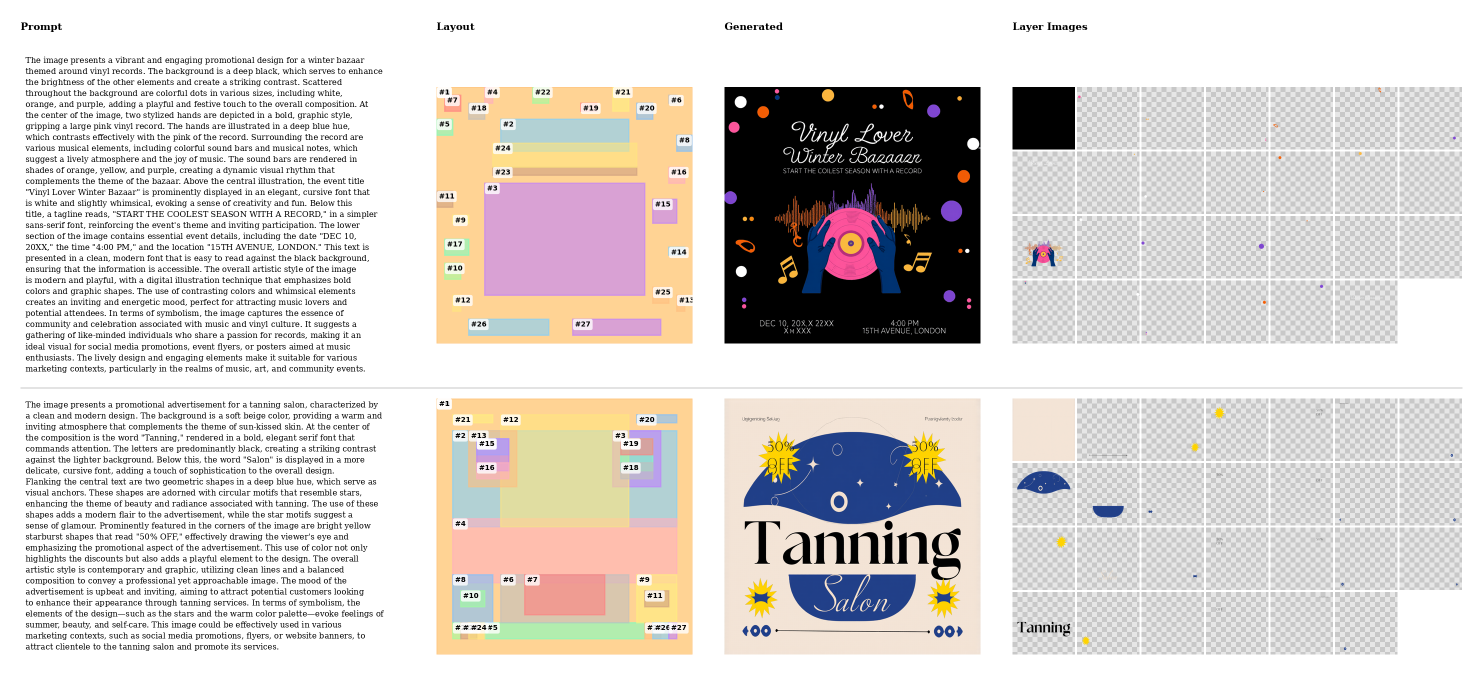}
    \caption{\footnotesize{\textbf{Additional text-to-layers generation examples.} Our unified framework handles various design complexities, from simple compositions to intricate multi-element designs with over 25 layers, while maintaining generation quality.}}
    \label{fig:supp_t2l_vis_pl_3}
\end{figure*}

\begin{figure*}[t]
    \centering
    \includegraphics[width=\linewidth]{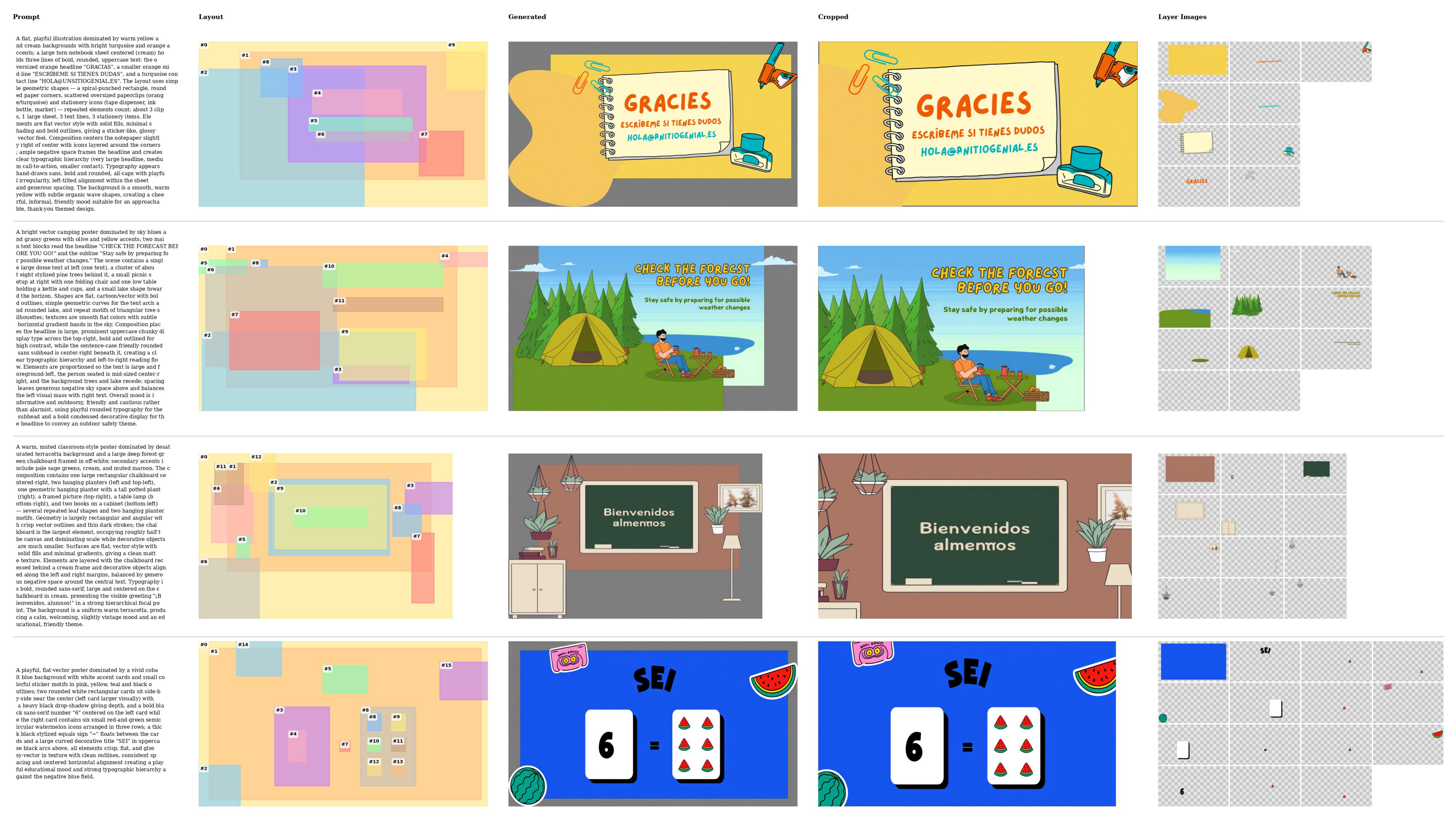}
    \caption{\footnotesize{\textbf{Text-to-layers with overflow layer generation.} Additional examples highlighting our method's unique capability to generate overflow layers that extend beyond the background boundary. As discussed in Section~\ref{sec:approach} and shown in Fig.~\ref{fig:overflow_layers}, over 60\% of designs contain overflow layers. Our approach generates complete full-size RGBA layers on the canvas, preserving editability and reusability that previous methods sacrifice by truncating pixels at background boundaries.}}
    \label{fig:supp_t2l_vis_overflow}
\end{figure*}

\begin{figure*}[t]
    \centering
    \includegraphics[width=1.0\linewidth]{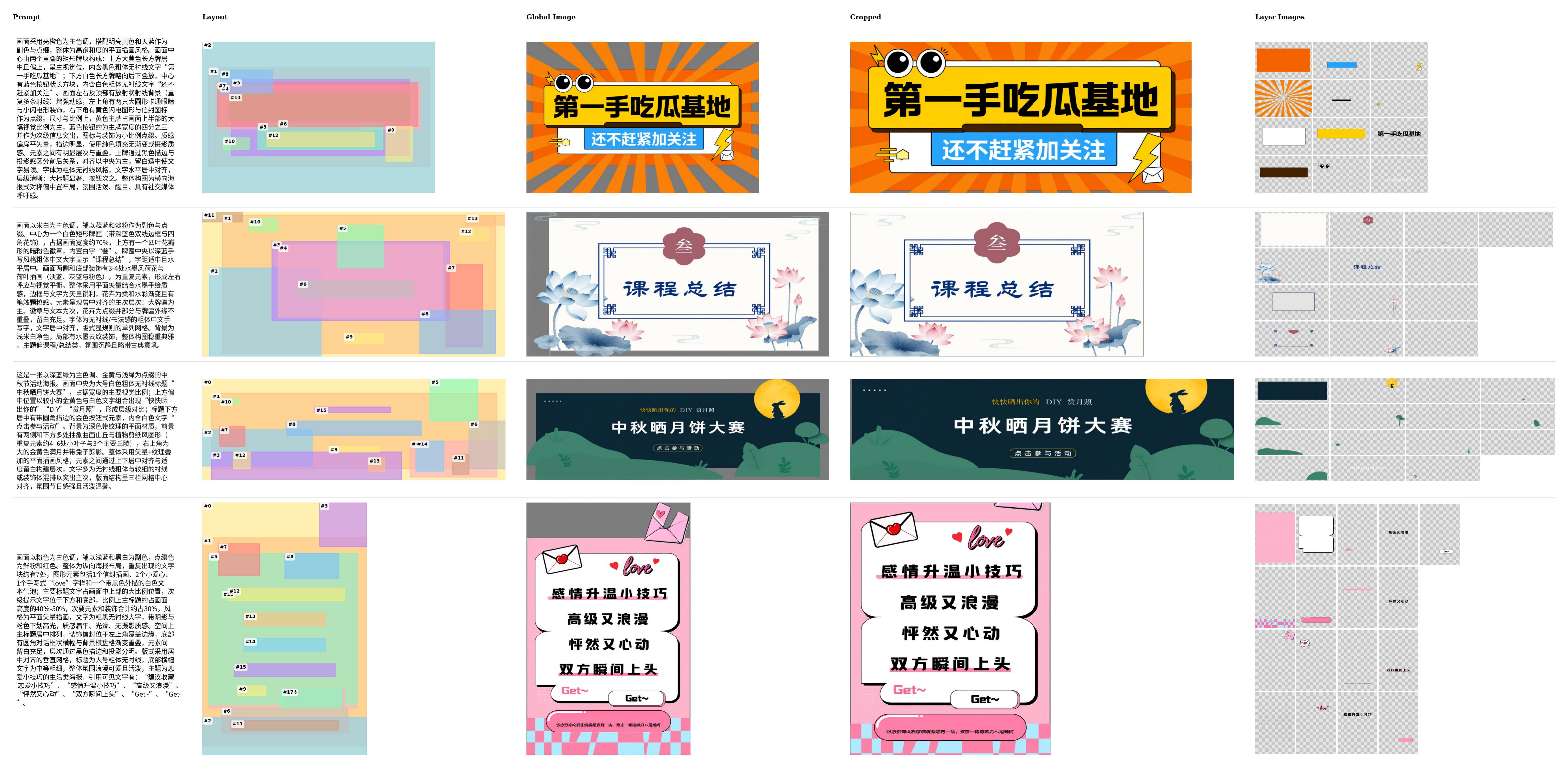}
    \caption{\footnotesize{\textbf{Text-to-layers with multilingual support.} Examples demonstrating our model's capability to generate designs with multilingual text layers. Our dataset includes diverse languages (as shown in Fig.~\ref{fig:statistics}), enabling generation of visually rendered text in multiple languages including Chinese. This showcases the model's ability to handle typography across different writing systems while maintaining design quality.}}
    \label{fig:supp_t2l_vis_multilingual}
\end{figure*}

We visualize the qualitative results of our Text-to-Layer task in Fig.~\ref{fig:supp_t2l_vis_1} through Fig.~\ref{fig:supp_t2l_vis_overflow}. Our Masked Region Transformer demonstrates exceptional versatility in generating high-fidelity multi-layer designs solely from textual descriptions. As shown in Fig.~\ref{fig:supp_t2l_vis_1} through Fig.~\ref{fig:supp_t2l_vis_pl_3}, the model successfully synthesizes coherent compositions ranging from simple layouts to complex designs with over 25 layers and even more, maintaining strict spatial consistency and stylistic harmony. A key advantage of our approach is the native support for diverse typography; Fig.~\ref{fig:supp_t2l_vis_overflow} illustrates our unique overflow generation capability. Unlike prior methods that truncate content at the canvas edge, our model generates complete, full-size RGBA layers that extend beyond the visible background boundary, thereby preserving full editability and reusability for downstream compositional tasks. Furthermore, Fig.~\ref{fig:supp_t2l_vis_multilingual} highlights the model's capability to render accurate visual text across multiple languages, including Chinese, ensuring practical utility for global design applications.

\subsection{Comparative Analysis of Image-to-Layer}
\begin{figure*}[t]
    \centering
    \includegraphics[width=0.8\linewidth]{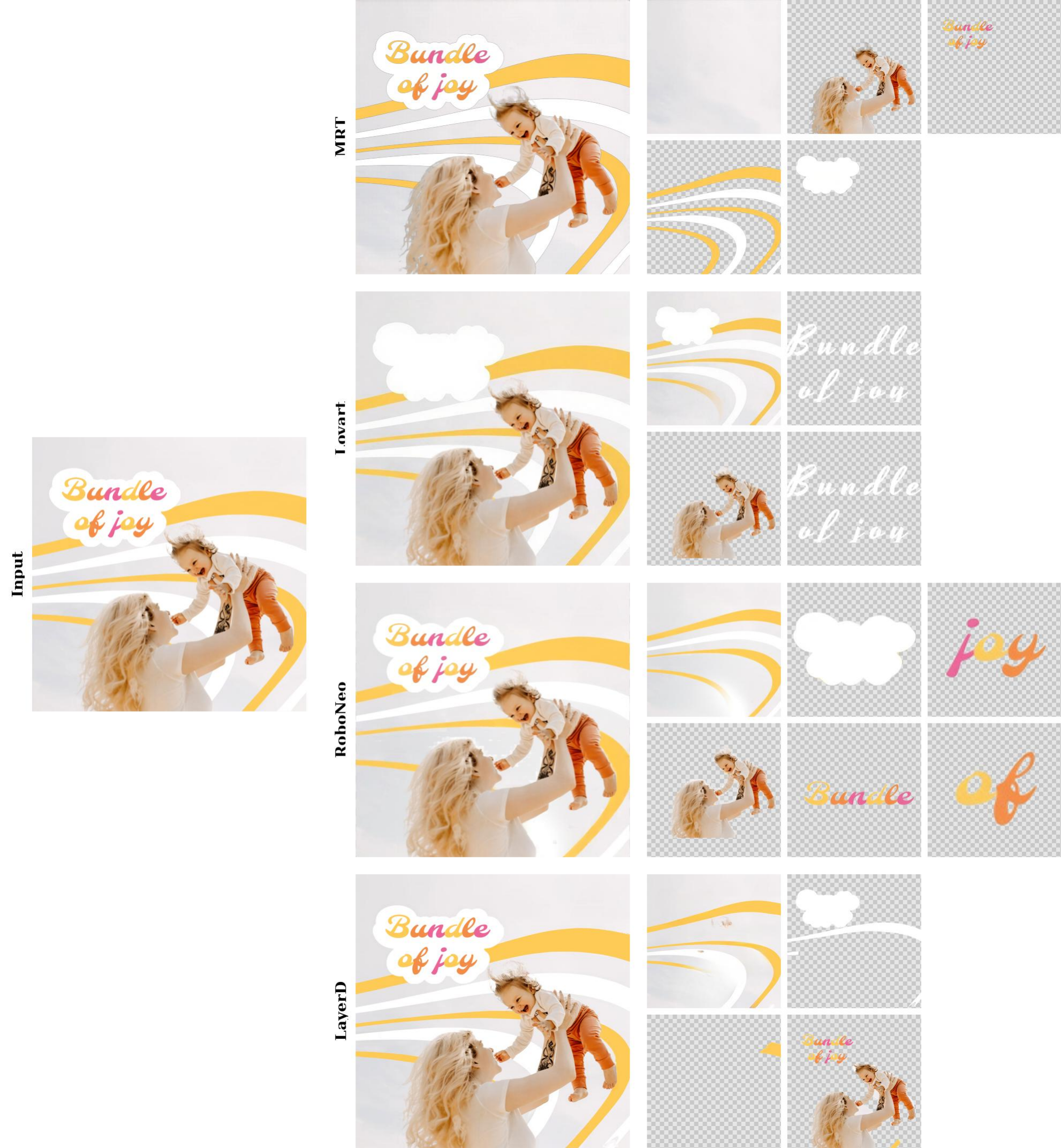}
    \caption{\footnotesize{\textbf{Qualitative comparison on image-to-layers task.} We compare our method with LayerD, Lovart, and RoboNeo on decomposing a graphic design into transparent layers. Each panel shows: the input image (top-left), followed by our result and baseline results with their decomposed layers. Our method produces cleaner layer boundaries, better granularity, and more complete RGBA layers compared to the baselines.}}
    \label{fig:supp_i2l_comp_0}
\end{figure*}

\begin{figure*}[t]
    \centering
    \includegraphics[width=\linewidth]{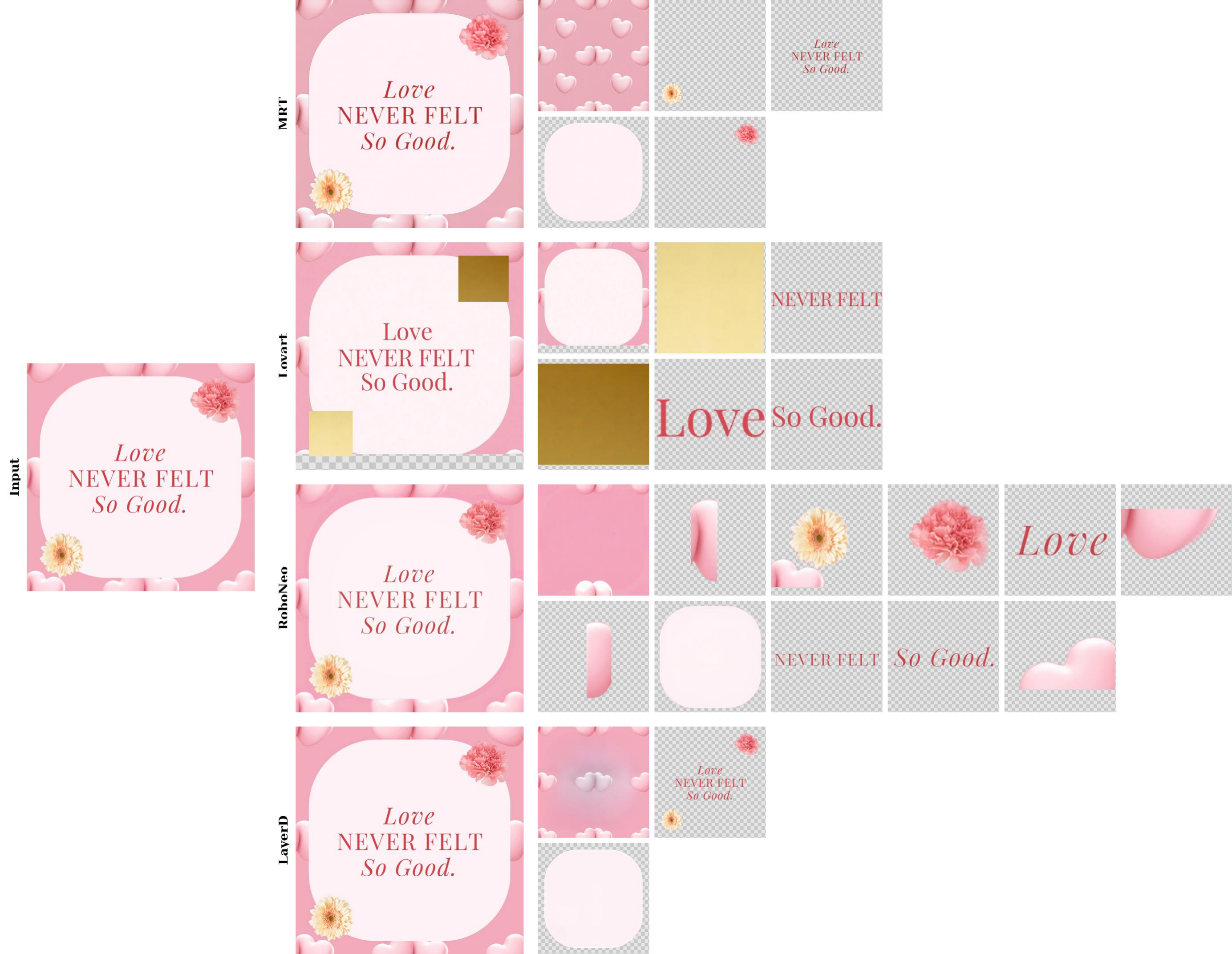}
    \caption{\footnotesize{\textbf{Additional qualitative comparison on image-to-layers task.} Our method demonstrates superior layer decomposition quality with better semantic correctness and transparency handling. The decomposed layers from our method maintain higher integrity and can faithfully reconstruct the input image, while baselines show issues with layer artifacts, improper grouping, or incomplete decomposition.}}
    \label{fig:supp_i2l_comp_1}
\end{figure*}

\begin{figure*}[t]
    \centering
    \includegraphics[width=0.8\linewidth]{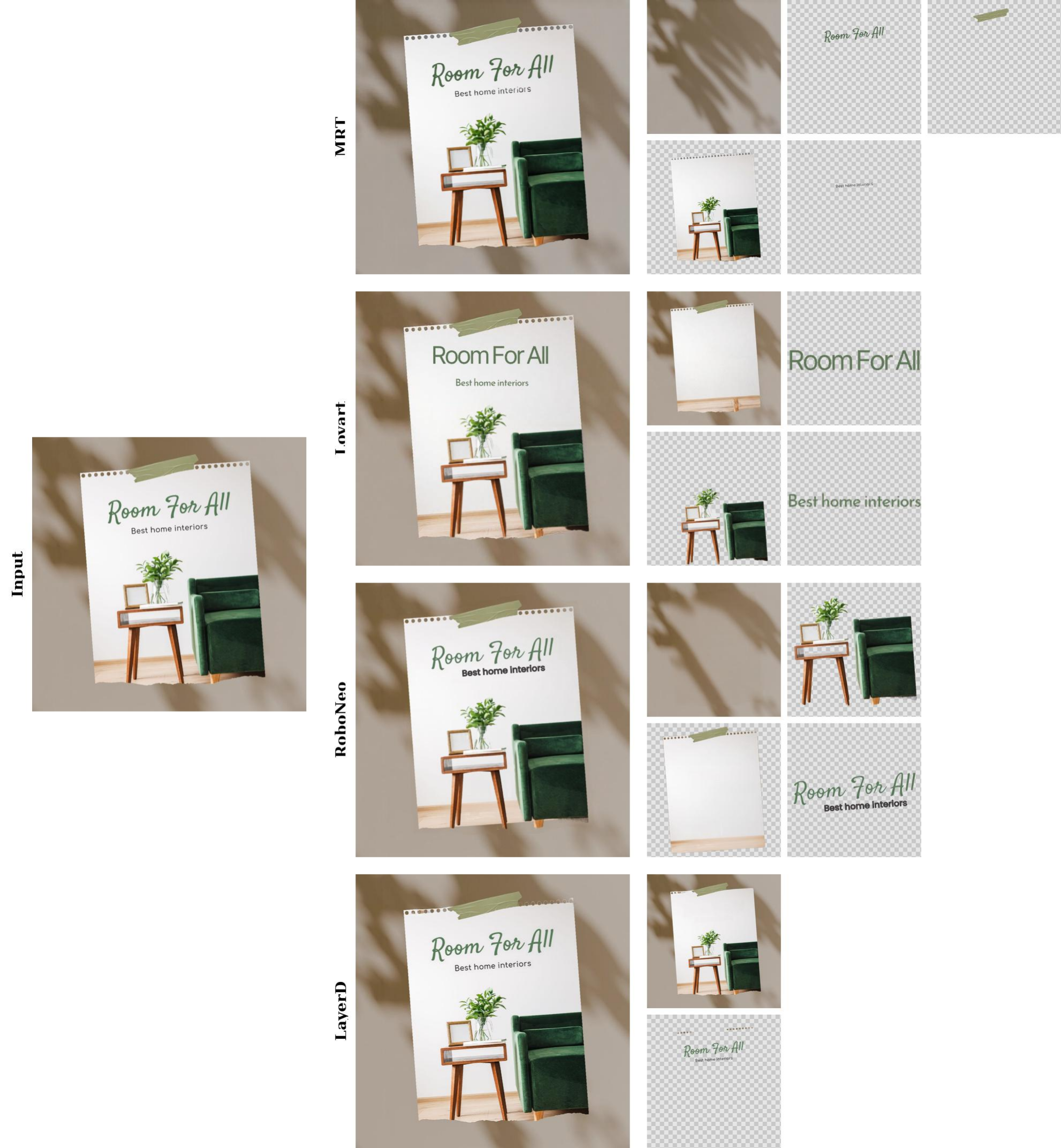}
    \caption{\footnotesize{\textbf{Additional qualitative comparison on image-to-layers task.} This example further demonstrates our method's advantages in layer quality, integrity, and appropriate granularity. Our approach successfully decomposes complex compositions while avoiding the overly grouped layers produced by LayerD or the artifacts present in commercial system outputs.}}
    \label{fig:supp_i2l_comp_2}
\end{figure*}

\begin{figure*}[t]
    \centering
    \includegraphics[width=\linewidth]{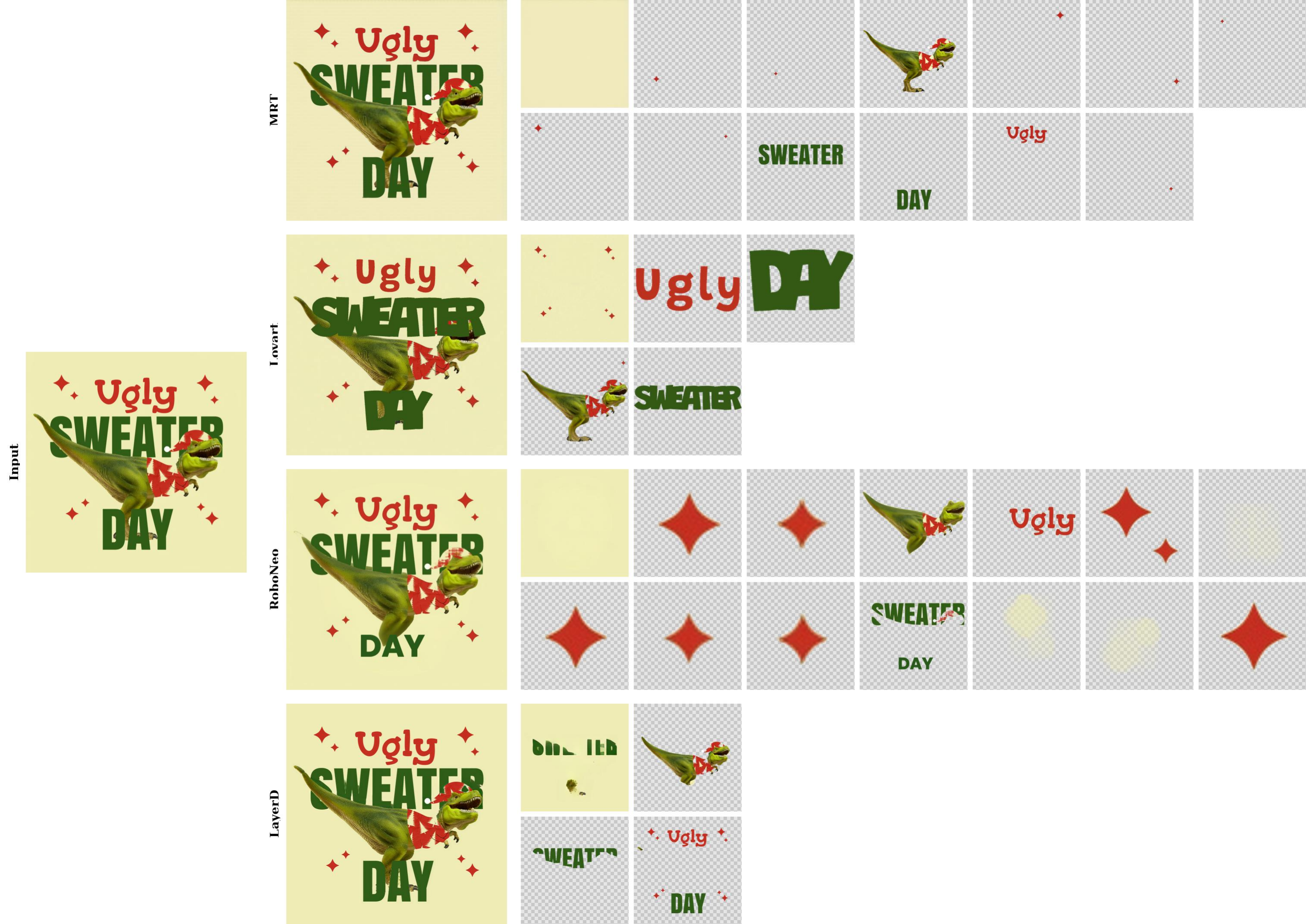}
    \caption{\footnotesize{\textbf{Additional qualitative comparison on image-to-layers task.} Our method consistently outperforms baselines across different design styles and complexities. The visualization shows that our approach produces high-quality transparent layers with accurate alpha channels and proper semantic decomposition, essential for downstream editing tasks.}}
    \label{fig:supp_i2l_comp_3}
\end{figure*}

\begin{figure*}[t]
    \centering
    \includegraphics[width=\linewidth]{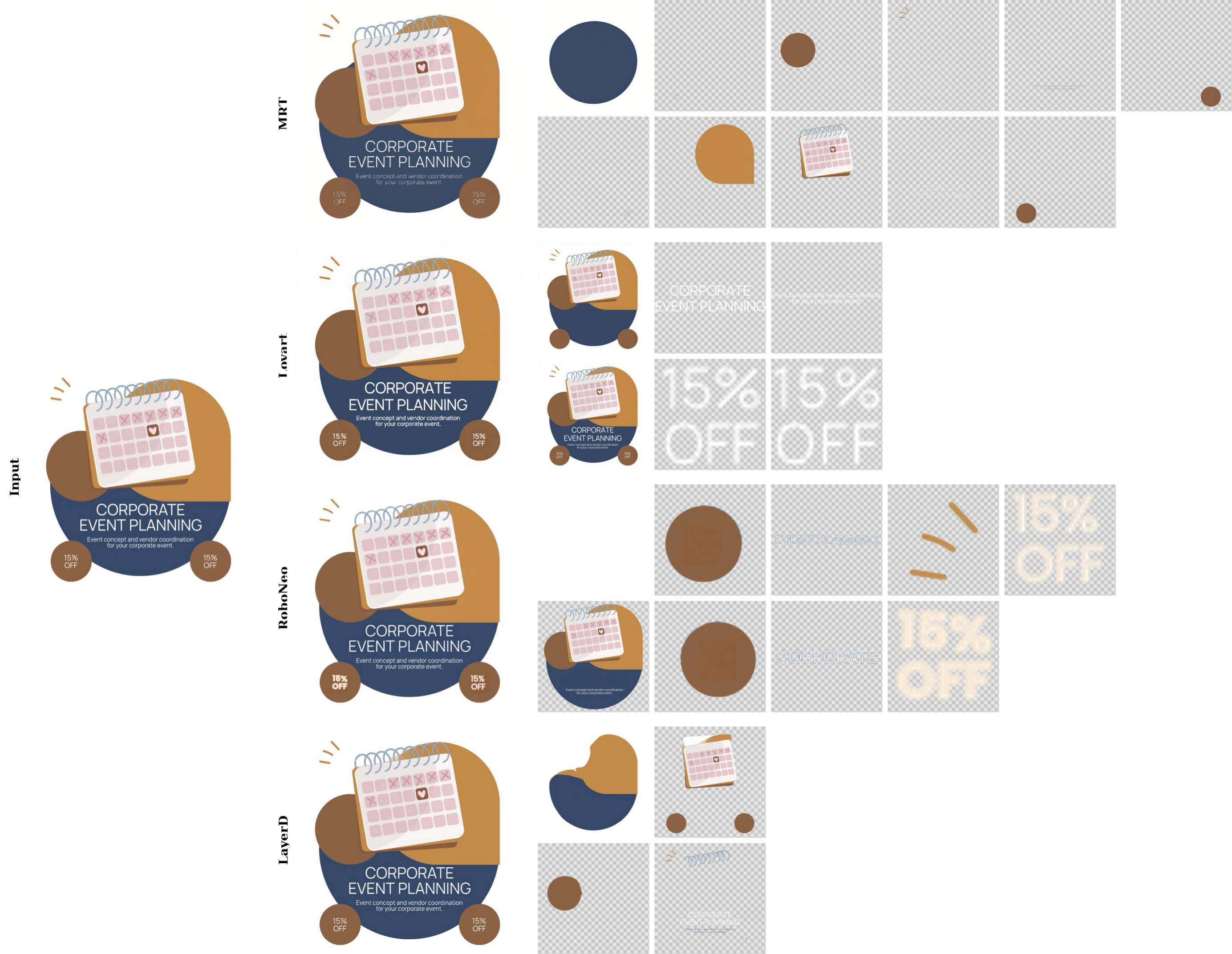}
    \caption{\footnotesize{\textbf{Additional qualitative comparison on image-to-layers task.} This case highlights our method's ability to handle complex multi-element designs. While commercial systems like RoboNeo suffer from severe artifacts and LayerD produces overly grouped layers that compromise fine-grained editing flexibility, our method maintains both quality and appropriate decomposition granularity.}}
    \label{fig:supp_i2l_comp_4}
\end{figure*}

\begin{figure*}[t]
    \centering
    \includegraphics[width=0.8\linewidth]{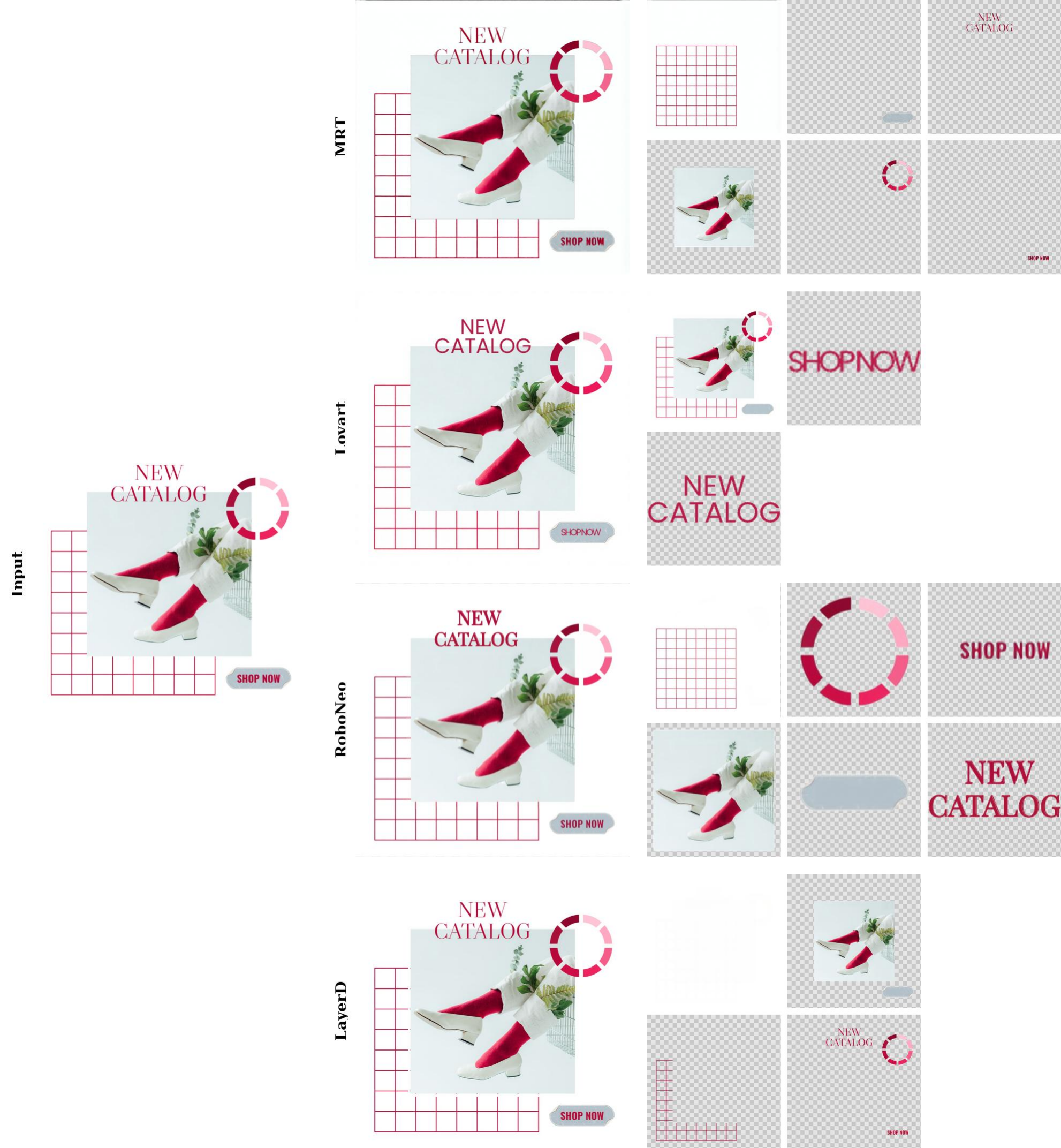}
    \caption{\footnotesize{\textbf{Additional qualitative comparison on image-to-layers task.} Our method excels at decomposing designs with overlapping elements and complex visual hierarchies. The comparison demonstrates superior performance across all three evaluation dimensions: quality (semantic correctness and transparency), integrity (faithful reconstruction), and granularity (appropriate decomposition level).}}
    \label{fig:supp_i2l_comp_5}
\end{figure*}

\begin{figure*}[t]
    \centering
    \includegraphics[width=\linewidth]{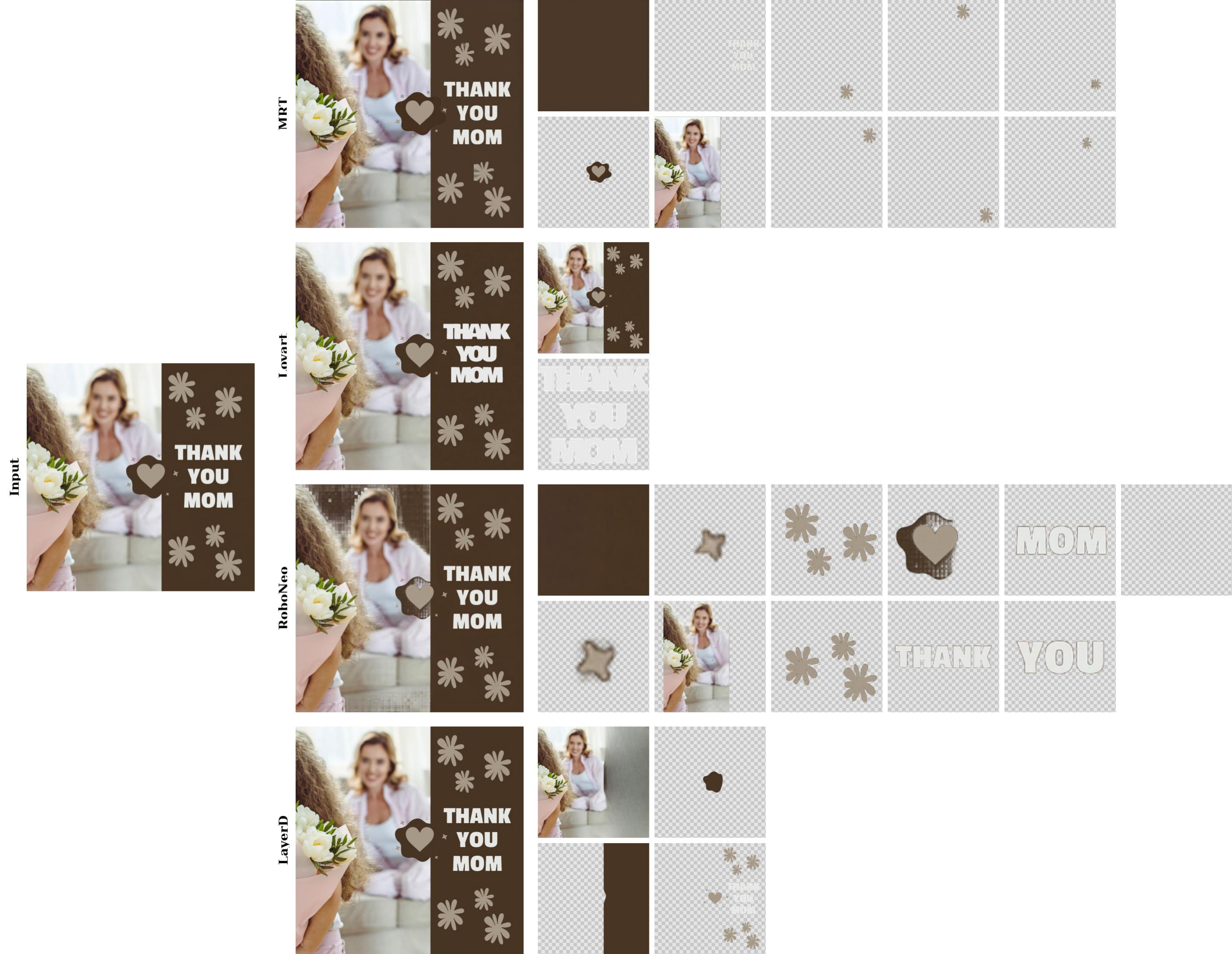}
    \caption{\footnotesize{\textbf{Additional qualitative comparison on image-to-layers task.} This example showcases our method's robustness across different design categories. Our decomposed layers maintain sharp boundaries, clean transparency, and semantic coherence, enabling practical editing workflows that commercial and academic baselines struggle to support.}}
    \label{fig:supp_i2l_comp_6}
\end{figure*}

\begin{figure*}[t]
    \centering
    \includegraphics[width=\linewidth]{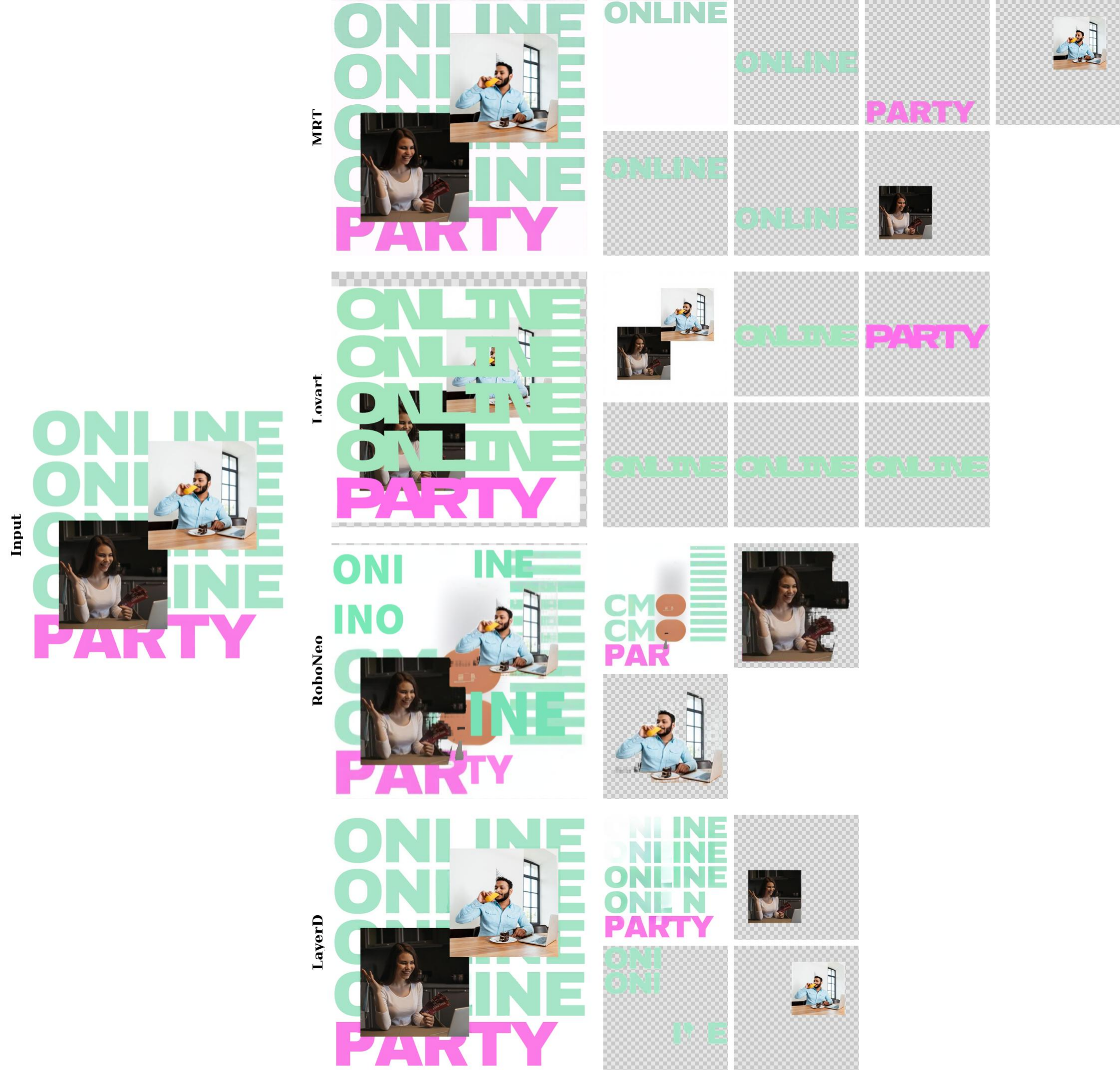}
    \caption{\footnotesize{\textbf{Additional qualitative comparison on image-to-layers task.} Final comparison case demonstrating consistent quality advantages of our method. The decomposition preserves layer reusability and editability while maintaining visual fidelity, confirming the effectiveness of our masked region transformer framework for the image-to-layers task.}}
    \label{fig:supp_i2l_comp_7}
\end{figure*}

In Fig.~\ref{fig:supp_i2l_comp_0} through Fig.~\ref{fig:supp_i2l_comp_7}, we provide a comprehensive qualitative comparison between our approach and state-of-the-art baselines, including LayerD, Lovart, and RoboNeo. The results consistently demonstrate that our method establishes a new standard for layer decomposition quality. While commercial systems like RoboNeo often introduce visual artifacts or fail to produce clean transparency, and academic baselines like LayerD tend to produce overly grouped layers that limit editing flexibility, our Masked Region Transformer achieves a superior balance. Our method excels in generating precise alpha mattes, maintaining semantic integrity, and achieving appropriate decomposition granularity (e.g., separating distinct visual elements rather than merging them). This is particularly evident in complex overlapping regions, where our model successfully disentangles elements that other methods fail to separate.

\subsection{Scalability on Layer Counts in Image-to-Layer}
\begin{figure*}[t]
  \centering
  \begin{minipage}[t]{0.49\textwidth}
    \centering
    \includegraphics[width=\linewidth]{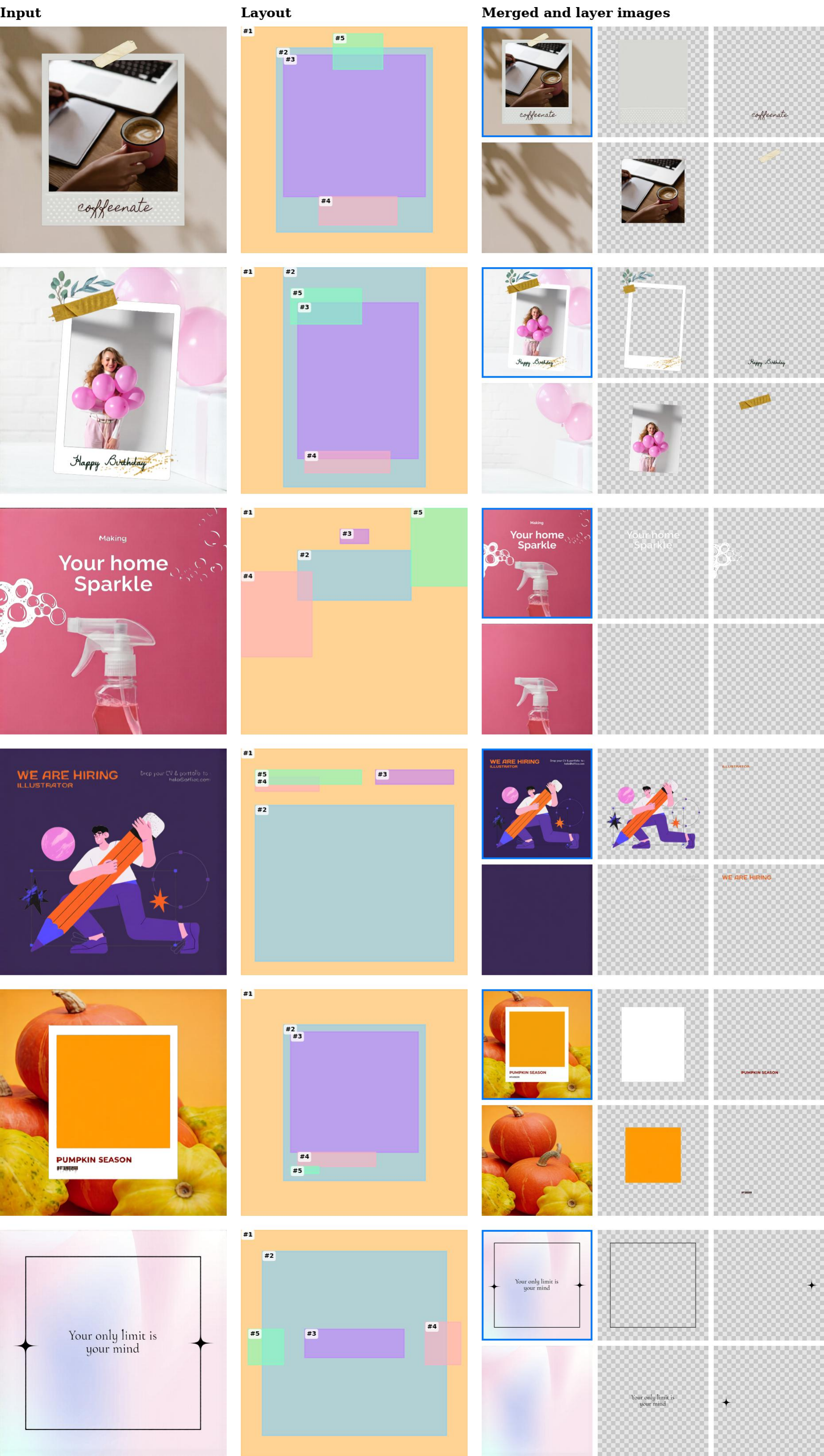}
  \end{minipage}\hfill
  \begin{minipage}[t]{0.49\textwidth}
    \centering
    \includegraphics[width=\linewidth]{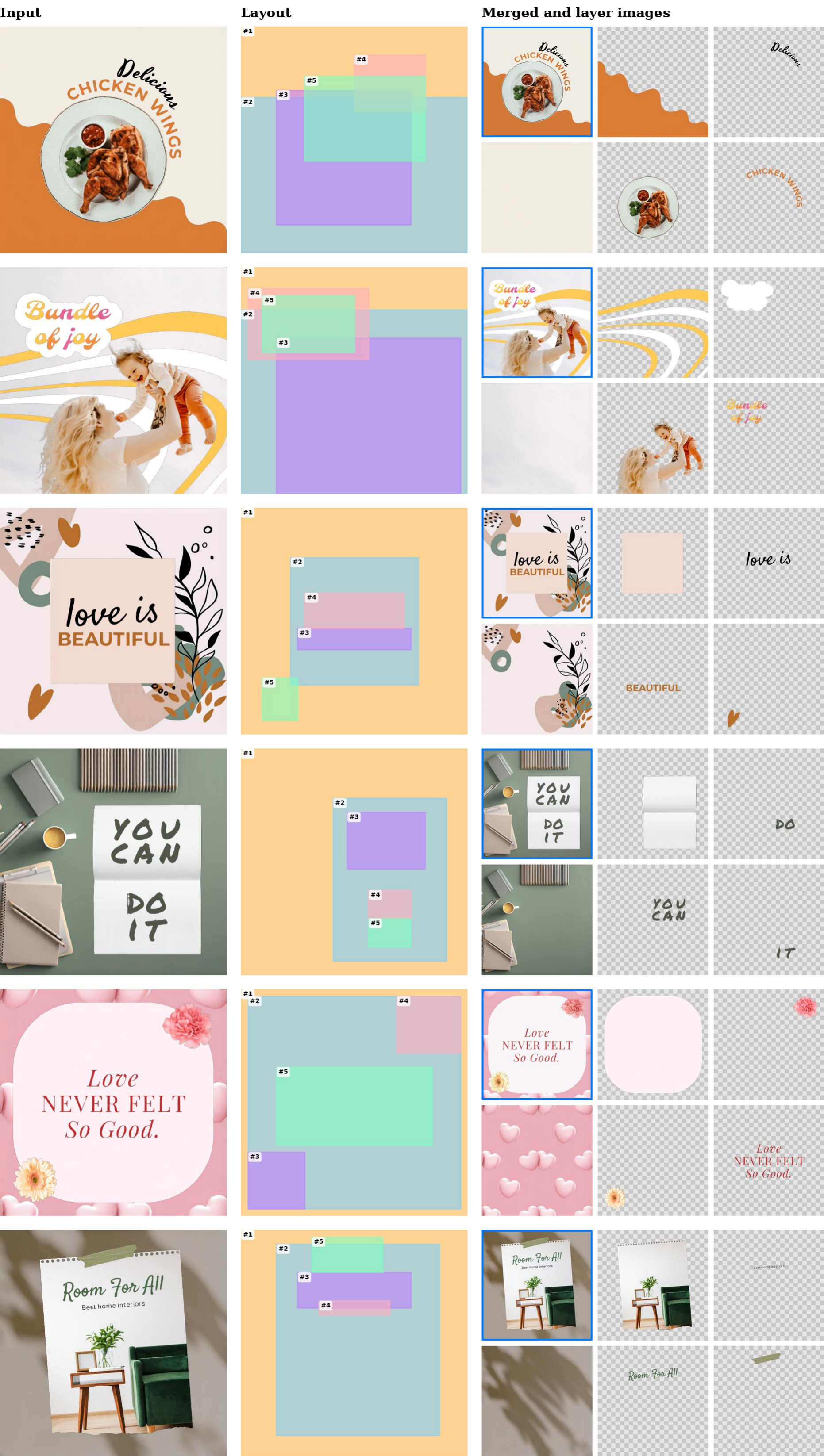}
  \end{minipage}
    \vspace{-2mm}
  \caption{\footnotesize{\textbf{Image-to-layers visualization with 6 layers.} We visualize the layer-by-layer decomposition process showing individual RGBA layers with transparency. Each layer is displayed separately along with its alpha mask, and the merged composition demonstrates faithful reconstruction of the input design. This visualization demonstrates our method's ability to generate clean, reusable layers with accurate spatial boundaries and alpha channels.}}
  \label{fig:supp_i2l_vis_6}
\end{figure*}

\begin{figure*}[t]
  \centering
  \begin{minipage}[t]{0.49\textwidth}
    \centering
    \includegraphics[width=\linewidth]{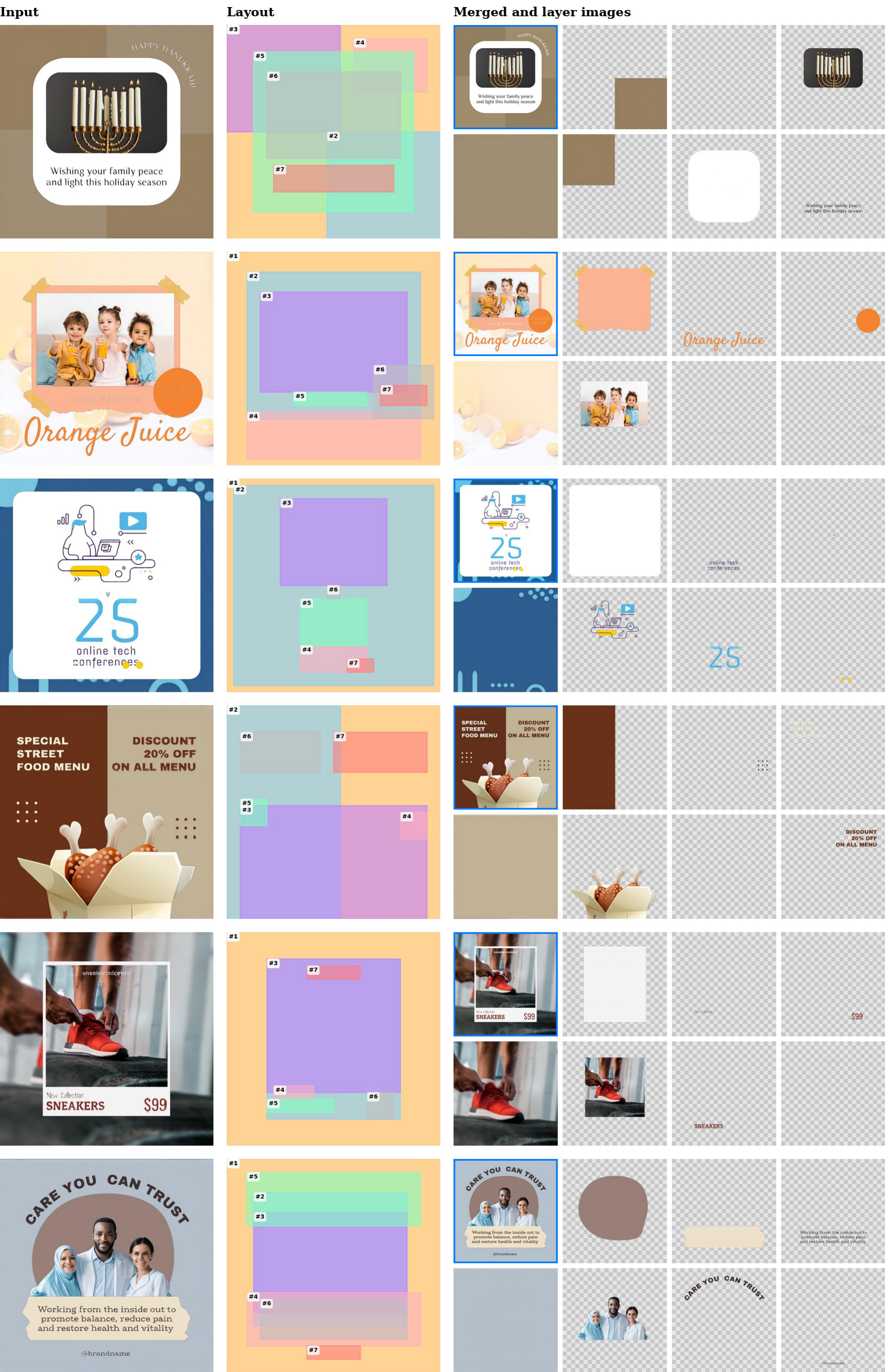}
  \end{minipage}\hfill
  \begin{minipage}[t]{0.49\textwidth}
    \centering
    \includegraphics[width=\linewidth]{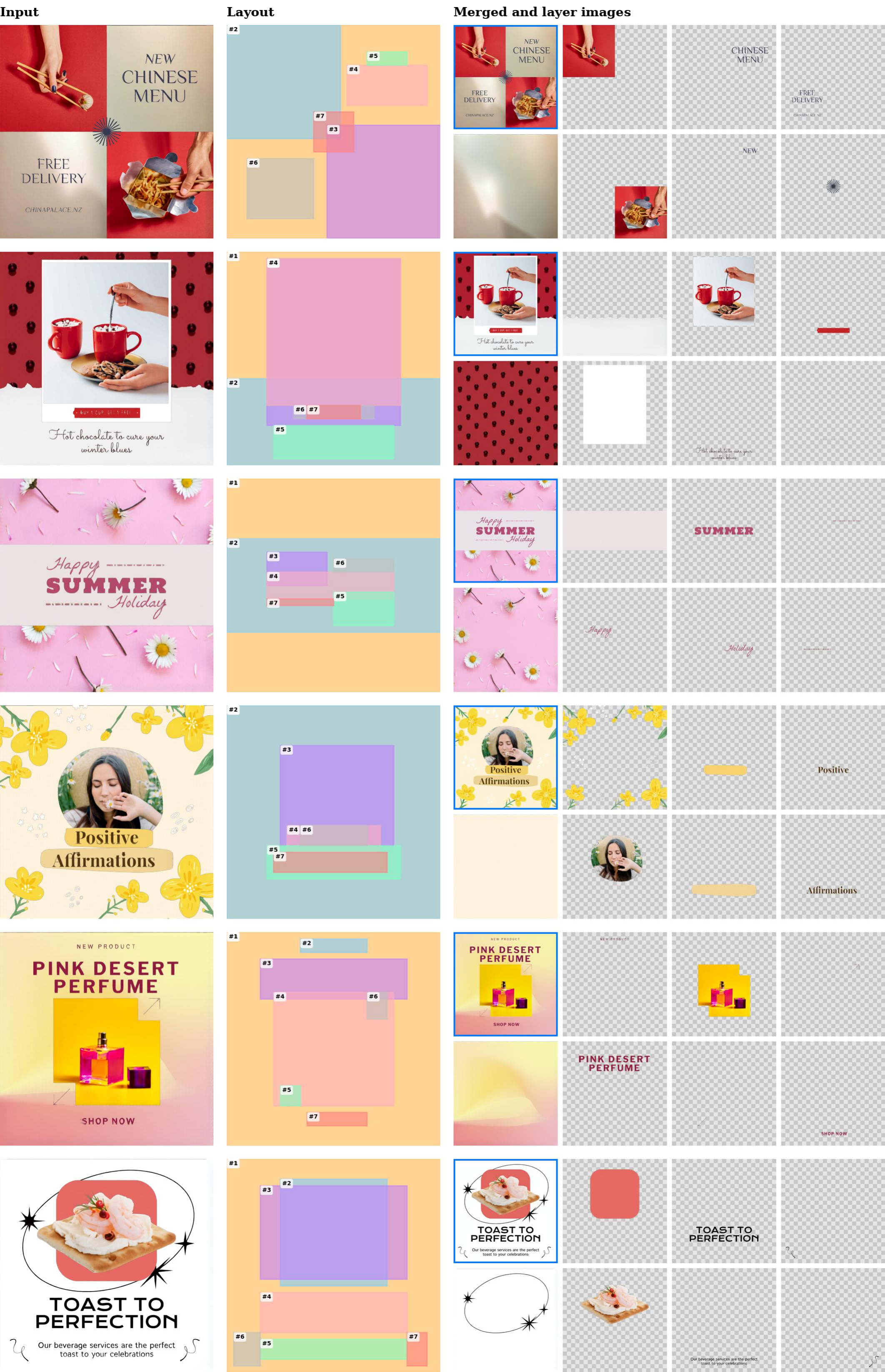}
  \end{minipage}
    \vspace{-2mm}
  \caption{\footnotesize{\textbf{Image-to-layers visualization with 8 layers.} Decomposition result showing increased layer complexity with 8 distinct transparent layers. Our method successfully handles more complex compositions, maintaining layer quality and proper decomposition granularity across the extended layer hierarchy. Each layer preserves semantic meaning and can be independently edited.}}
  \label{fig:supp_i2l_vis_8}
\end{figure*}

\begin{figure*}[t]
    \centering
    \vspace{-6mm}
    \includegraphics[width=0.63\linewidth]{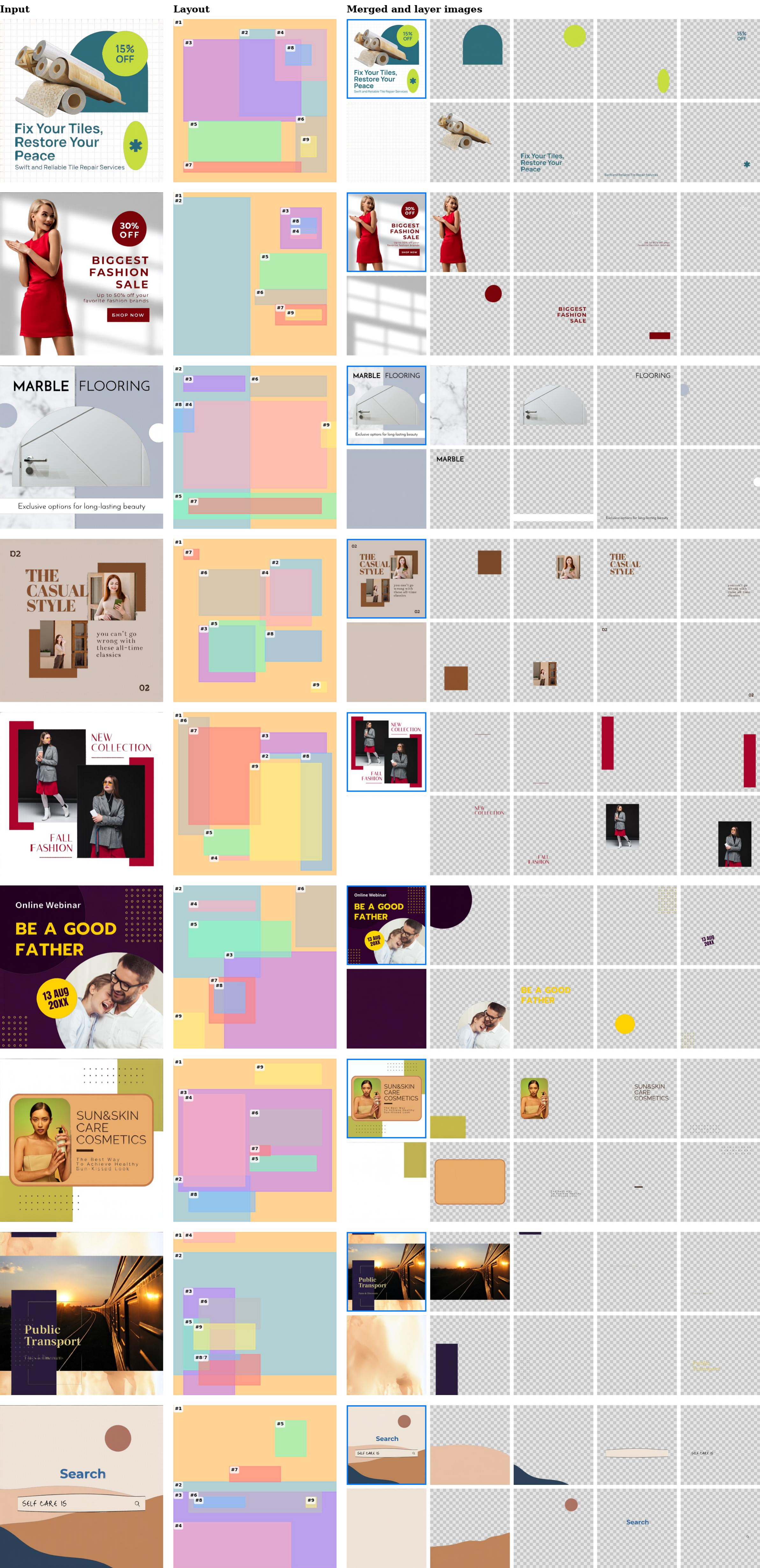}
    \vspace{-2mm}
    \caption{\footnotesize{\textbf{Image-to-layers visualization with 10 layers.} Further demonstrating scalability to compositions with 10 transparent layers. Our masked region transformer maintains stable performance across different layer counts, producing coherent decompositions without architectural modifications. The visualization shows consistent layer quality from background to foreground elements.}}
    \label{fig:supp_i2l_vis_10}
\end{figure*}

\begin{figure*}[t]
    \centering
    \vspace{-6mm}
    \includegraphics[width=0.7\linewidth]{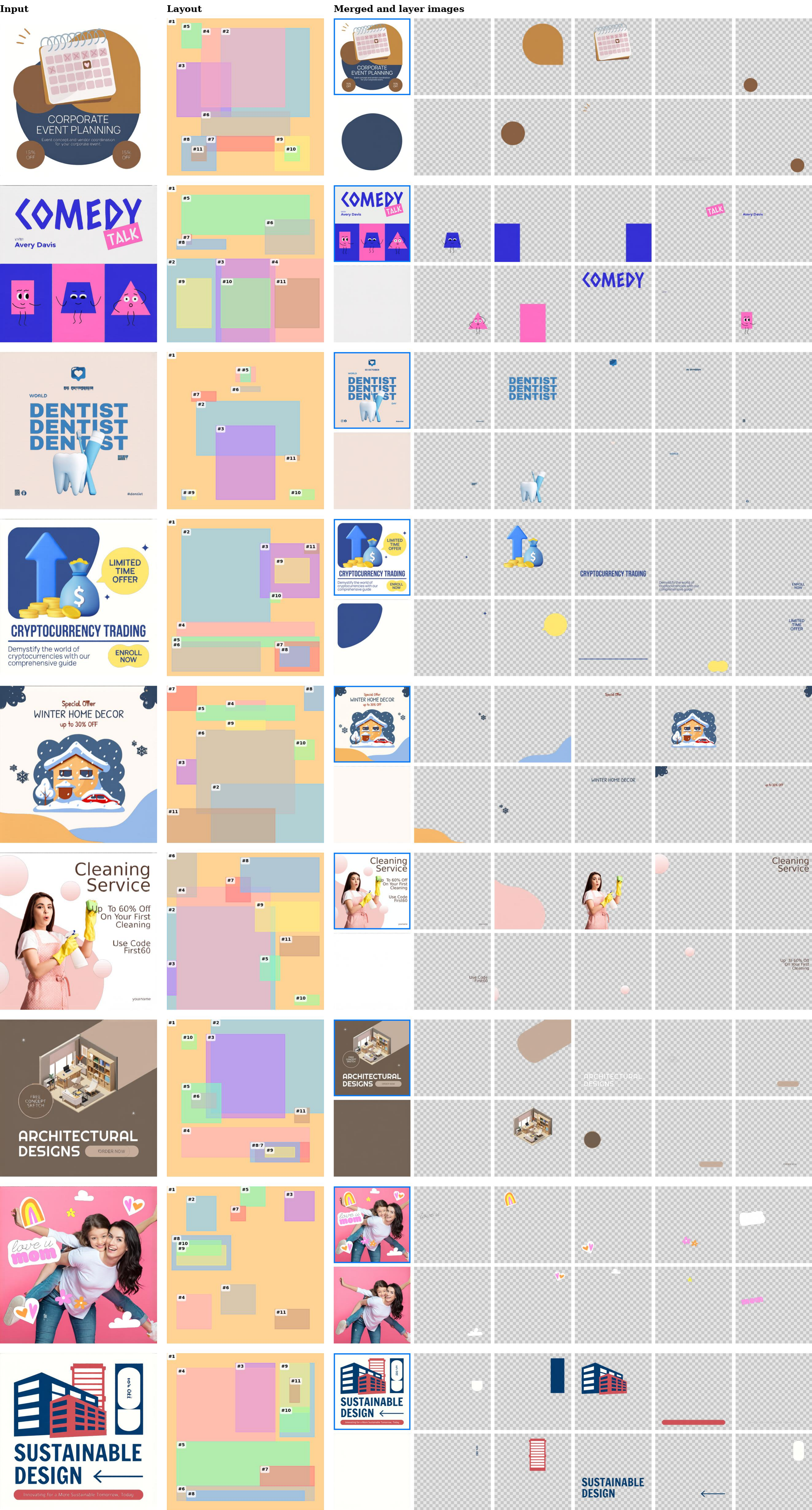}
    \vspace{-2mm}
    \caption{\footnotesize{\textbf{Image-to-layers visualization with 12 layers.} Decomposition of a complex design into 12 transparent layers, demonstrating our method's capability to handle high layer counts while maintaining decomposition quality. Each layer retains sharp boundaries and proper alpha masks, essential for professional editing workflows.}}
    \label{fig:supp_i2l_vis_12}
\end{figure*}

\begin{figure*}[t]
    \centering
    \vspace{-6mm}
    \includegraphics[width=0.75\linewidth]{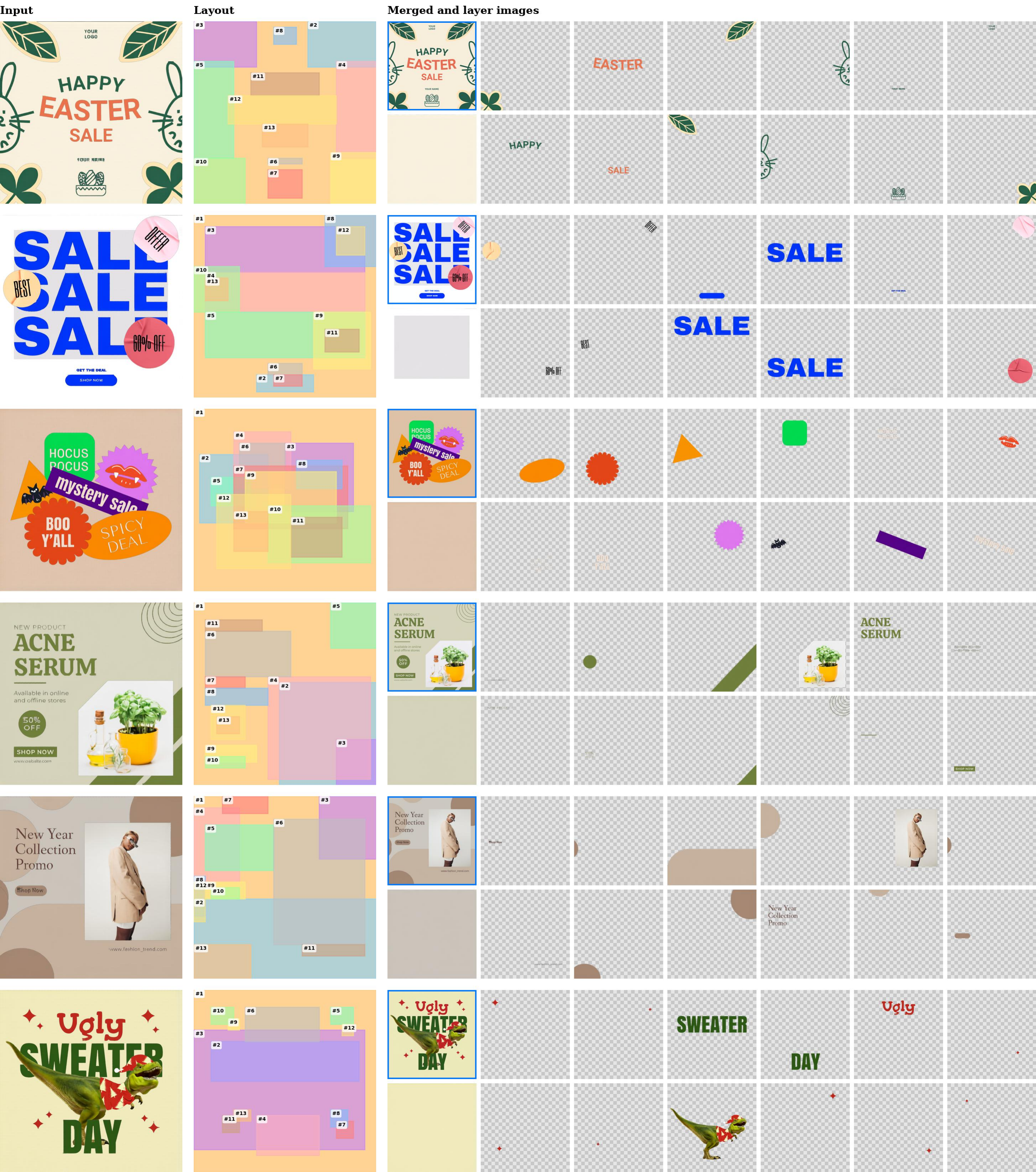}
    \vskip 1mm
    \includegraphics[width=0.85\linewidth]{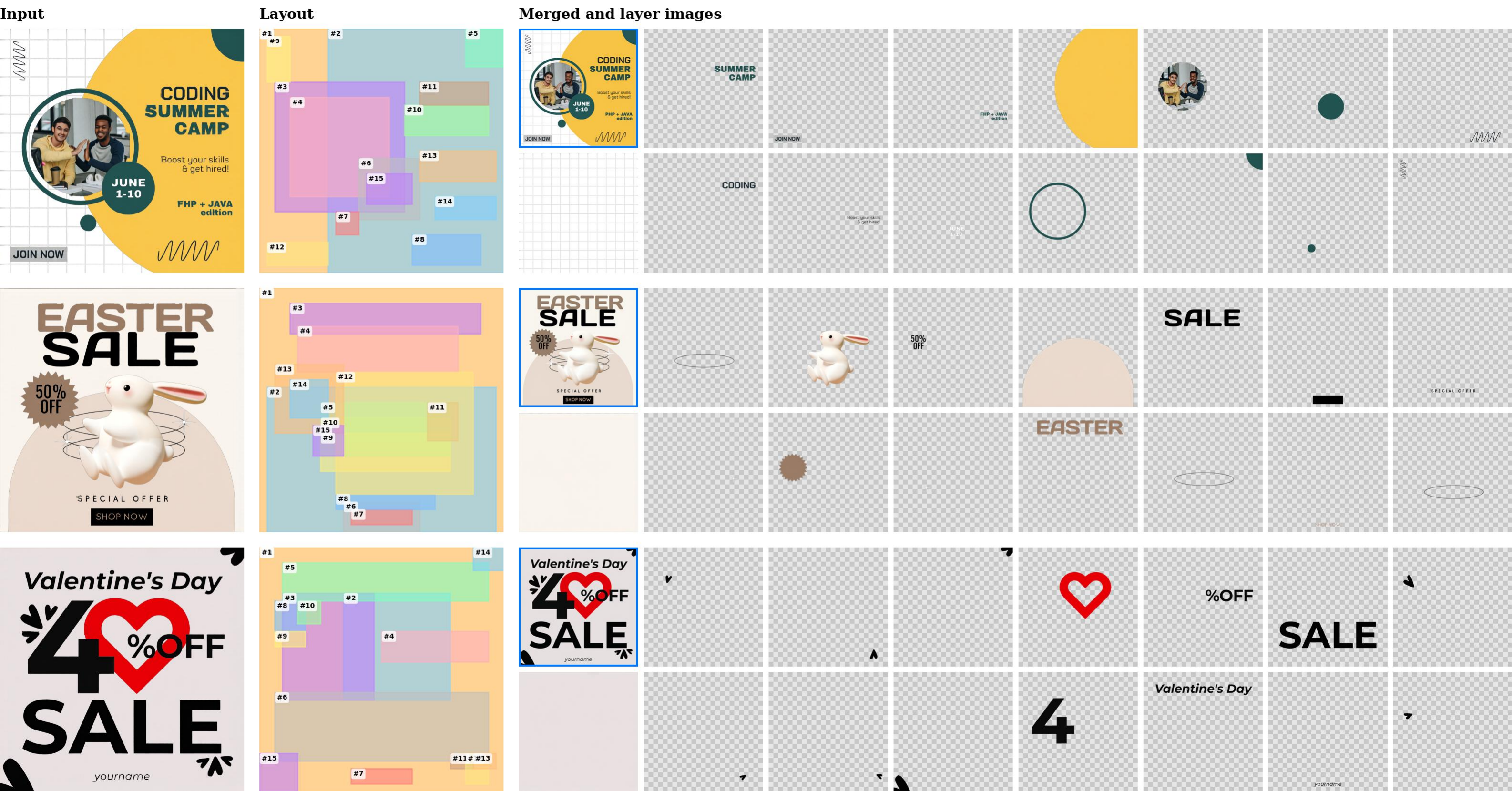}
    \vspace{-2mm}
    \caption{\footnotesize{\textbf{Image-to-layers visualization with 14 and 16 layers.} Two examples showcasing our method's scalability to very high layer counts (14 and 16 layers respectively). As shown in Table~\ref{tab:abl_layer_numbers_i2l}, our approach maintains stable performance across a wide range of layer numbers from 2 to 50 layers, demonstrating flexibility in handling both simple and complex multi-element compositions.}}
    \label{fig:supp_i2l_vis_14_16}
\end{figure*}

To evaluate the robustness of our framework, we visualize image-to-layers decomposition results across varying degrees of complexity in Fig.~\ref{fig:supp_i2l_vis_6} through Fig.~\ref{fig:supp_i2l_vis_14_16}, ranging from 6 layers up to 16 layers. These visualizations confirm that our architecture scales effectively without performance degradation. The model maintains consistent quality in boundary detection and content preservation in cases of a wide range of layer counts. This stability across diverse layer counts validates the efficacy of our masked attention mechanism, proving that the model can handle the structural complexity of professional-grade graphic designs.

\subsection{Context-Aware Layer Addition}
\begin{figure*}[t]
    \centering
    \includegraphics[width=1.0\linewidth]{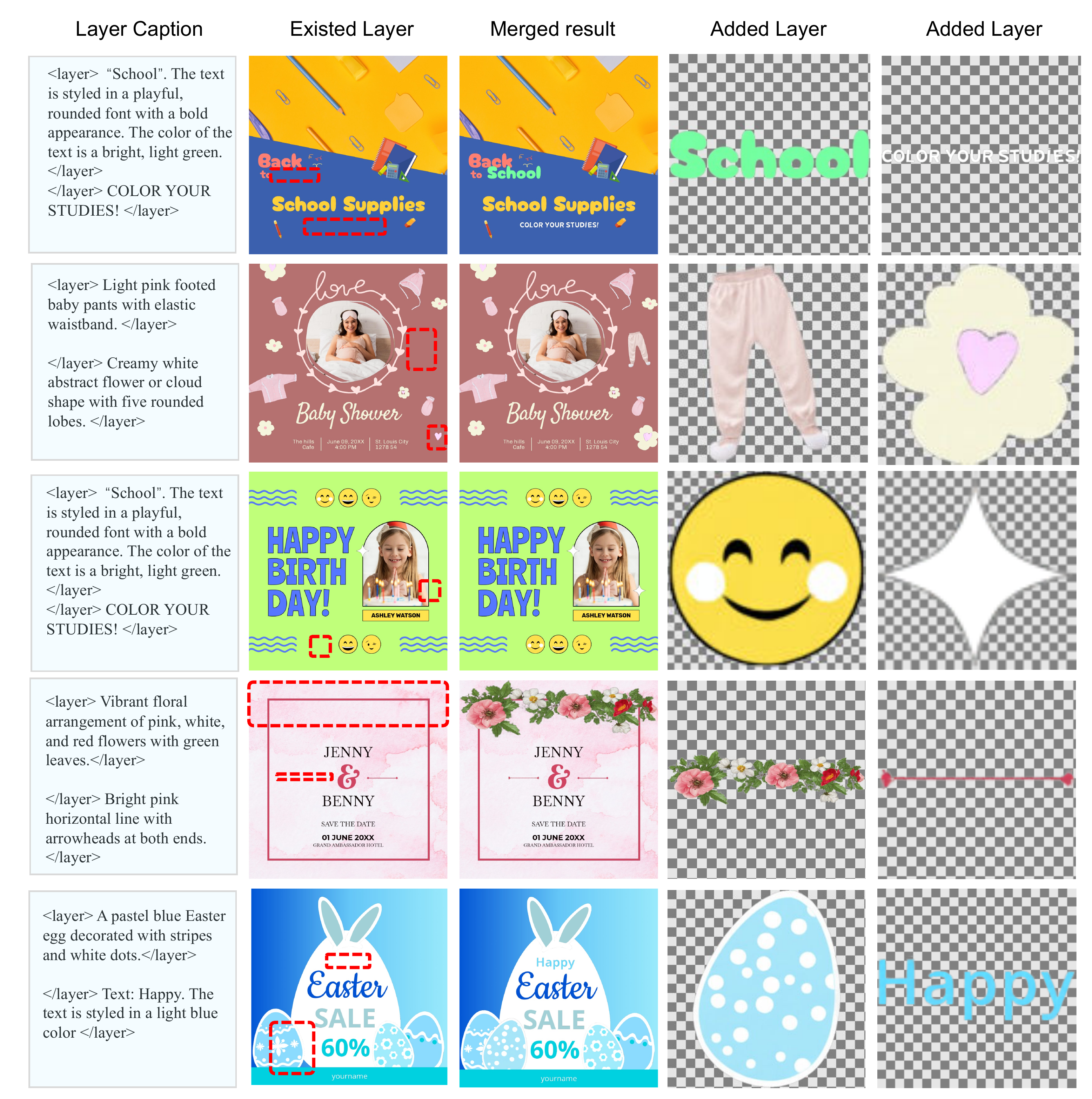}
    \caption{\footnotesize{\textbf{Additional examples for layer addition task.} We demonstrate the layers-to-layers capability by adding new layers to existing compositions based on text prompts. Our method generates new layers that maintain cross-layer consistency and harmonize with the existing design's spatial layout and visual style. By generating multiple layers in a single pass and conditioning on all existing layers, our approach better captures inter-layer relationships and produces coherent insertions that preserve global composition.}}
    \label{fig:supp_l2l_add}
\end{figure*}

Fig.~\ref{fig:supp_l2l_add} demonstrates the capabilities of our layers-to-layers task, specifically focusing on layer addition. Here, we simulate a user editing workflow where new elements—such as text or decorative objects—are inserted into an existing design based on text prompts and specified bounding boxes. The results show that our model does not merely paste isolated objects; instead, it generates new layers that are contextually aware, matching the lighting, perspective, and artistic style of the existing layers. By conditioning on the full composition, the added layers harmonize seamlessly with the original design, preserving the global aesthetic while fulfilling the user's semantic requirements.

\subsection{Layer Restylization and Harmonization}
\begin{figure*}[t]
    \centering
    \includegraphics[width=0.5\linewidth]{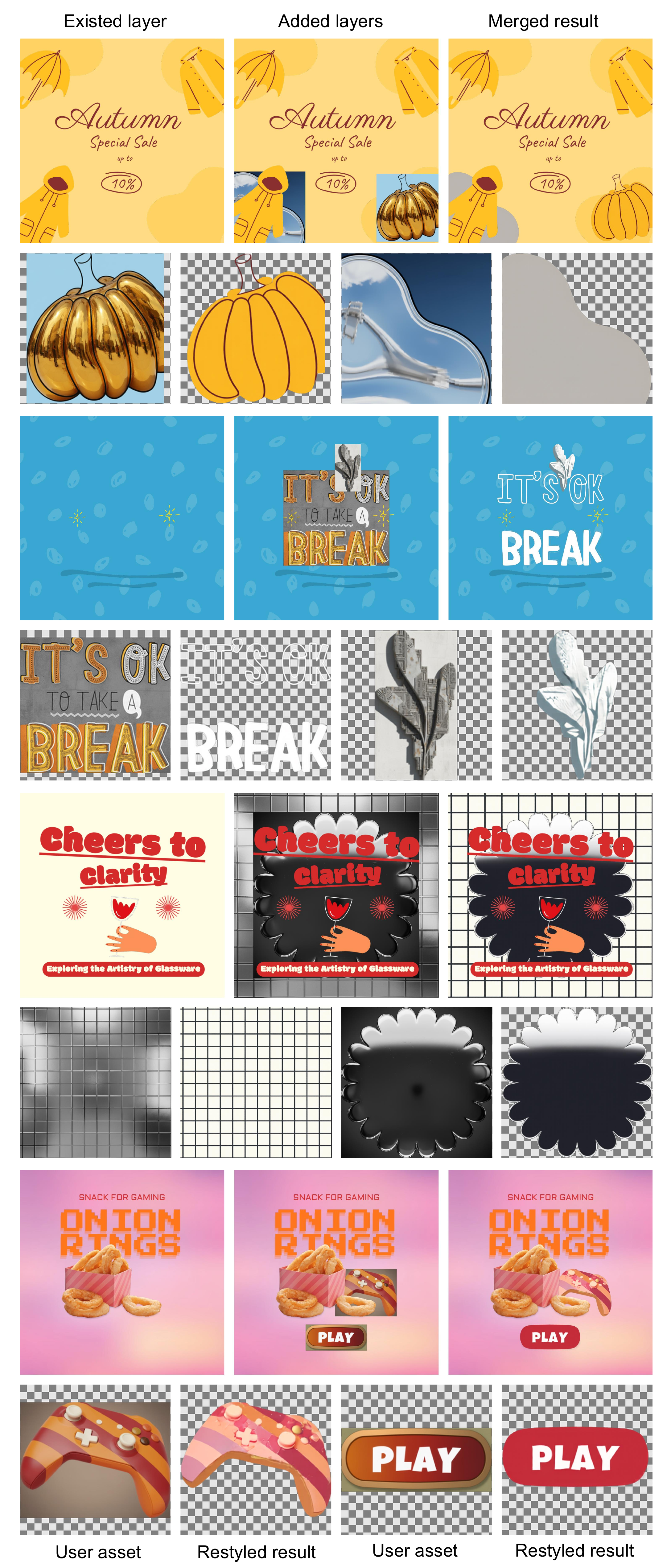}
    \caption{\footnotesize{\textbf{Additional examples for layer restylization task.} We visualize the transformation of user-provided assets into stylistically harmonized layers that match the overall composition. Our method performs this restylization in a single pass for all target layers, preserving geometric structure while adapting appearance to align with the existing design's visual identity. The results demonstrate effective style transfer while maintaining layer coherence and compositional harmony.}}
    \label{fig:supp_l2l_restyle}
\end{figure*}

In Fig.~\ref{fig:supp_l2l_restyle}, we showcase the layer restylization capability, where user-provided assets are transformed to align with a target design's visual identity. Our model effectively transfers style while preserving the geometric structure of the input asset. The visualization demonstrates that our single-pass generation approach ensures cross-layer consistency, successfully adapting the color palette, texture, and artistic rendering of external assets to match the pre-existing composition. This capability is essential for unifying disparate elements into a cohesive graphic design.

\subsection{Layout Generalization in Text-to-Layer}
\begin{figure*}[t]
    \centering
    \includegraphics[width=0.8\linewidth]{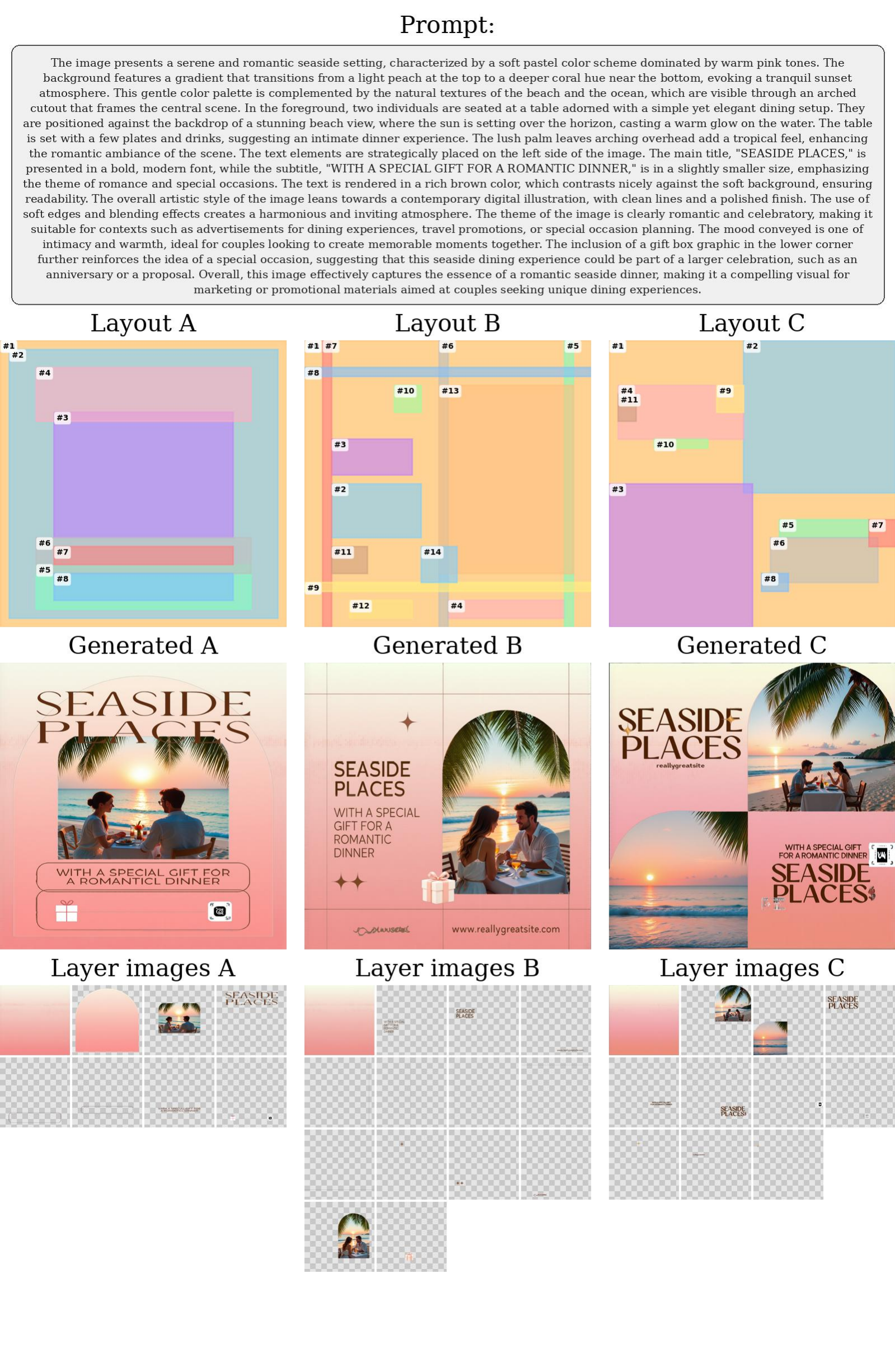}
    \vspace{-16mm}
    \caption{\footnotesize{\textbf{Text-to-layers: Merged image vs. layout visualization.} Additional example demonstrating our model's ability to generate well-composed multi-layer designs from text prompts. The side-by-side comparison shows how textual descriptions are translated into visual compositions (left) with structured layer hierarchies (right), highlighting the model's capability to learn both aesthetic and structural design principles.}}
    \label{fig:supp_t2l_layout_2}
\end{figure*}

\begin{figure*}[t]
    \centering
    \includegraphics[width=0.8\linewidth]{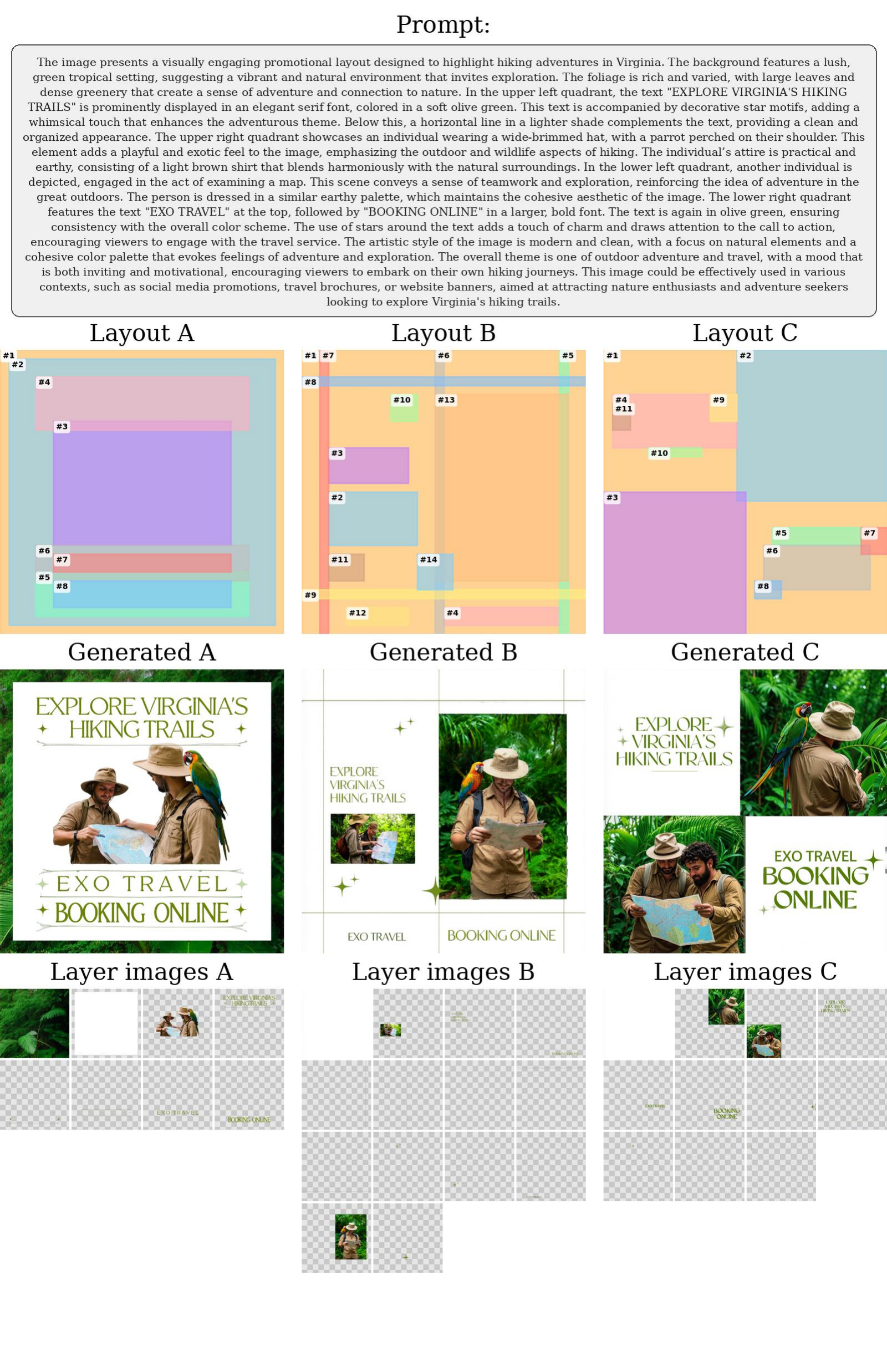}
    \vspace{-16mm}
    \caption{\footnotesize{\textbf{Text-to-layers: Merged image vs. layout visualization.} Another example showing the relationship between the generated merged design and its underlying layer layout structure. The layout visualization reveals how our model organizes multiple layers with appropriate spatial relationships, z-ordering, and compositional balance to create aesthetically pleasing designs from text descriptions.}}
    \label{fig:supp_t2l_layout_3}
\end{figure*}

Fig.~\ref{fig:supp_t2l_layout_2} and Fig.~\ref{fig:supp_t2l_layout_3} present an analysis of the interplay between text prompts and spatial controls. In these experiments, the text prompt contains implicit or explicit descriptions of element positions, while we simultaneously provide varying spatial layouts (bounding boxes) that may conflict with these textual descriptions. Remarkably, the results demonstrate that our model exhibits strong adherence to the user-provided layout, effectively overriding the spatial biases present in the text prompt while retaining the semantic content. This confirms that our framework successfully disentangles semantic generation from spatial arrangement, allowing users to enforce arbitrary layouts—such as moving a title from the top to the bottom—without compromising the generated content's quality or the prompt's semantic fidelity.

\subsection{Layout-Guided Image Decomposition}
\begin{figure*}[t]
    \centering
    \includegraphics[width=0.7\linewidth]{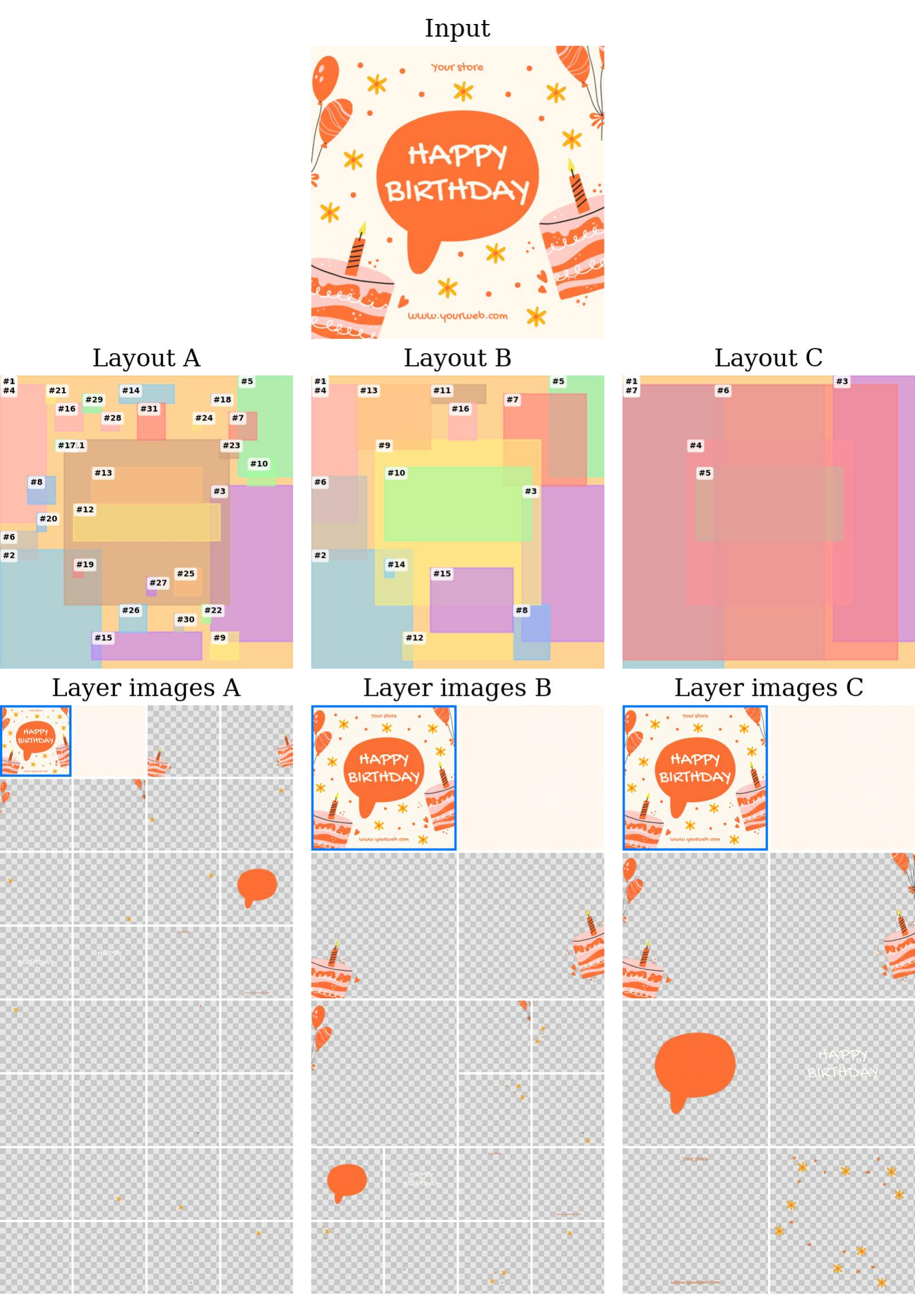}
    \vspace{-5mm}
    \caption{\footnotesize{\textbf{Image-to-layers: Merged image vs. layout visualization.} We visualize the input image alongside the extracted layer layout structure for the image-to-layers decomposition task. This demonstrates how our method decomposes raster images into semantically meaningful layers with well-defined spatial boundaries. The layout representation shows bounding boxes and z-order that guide the decomposition process.}}
    \label{fig:supp_i2l_layout_1}
\end{figure*}

\begin{figure*}[t]
    \centering
    \includegraphics[width=0.7\linewidth]{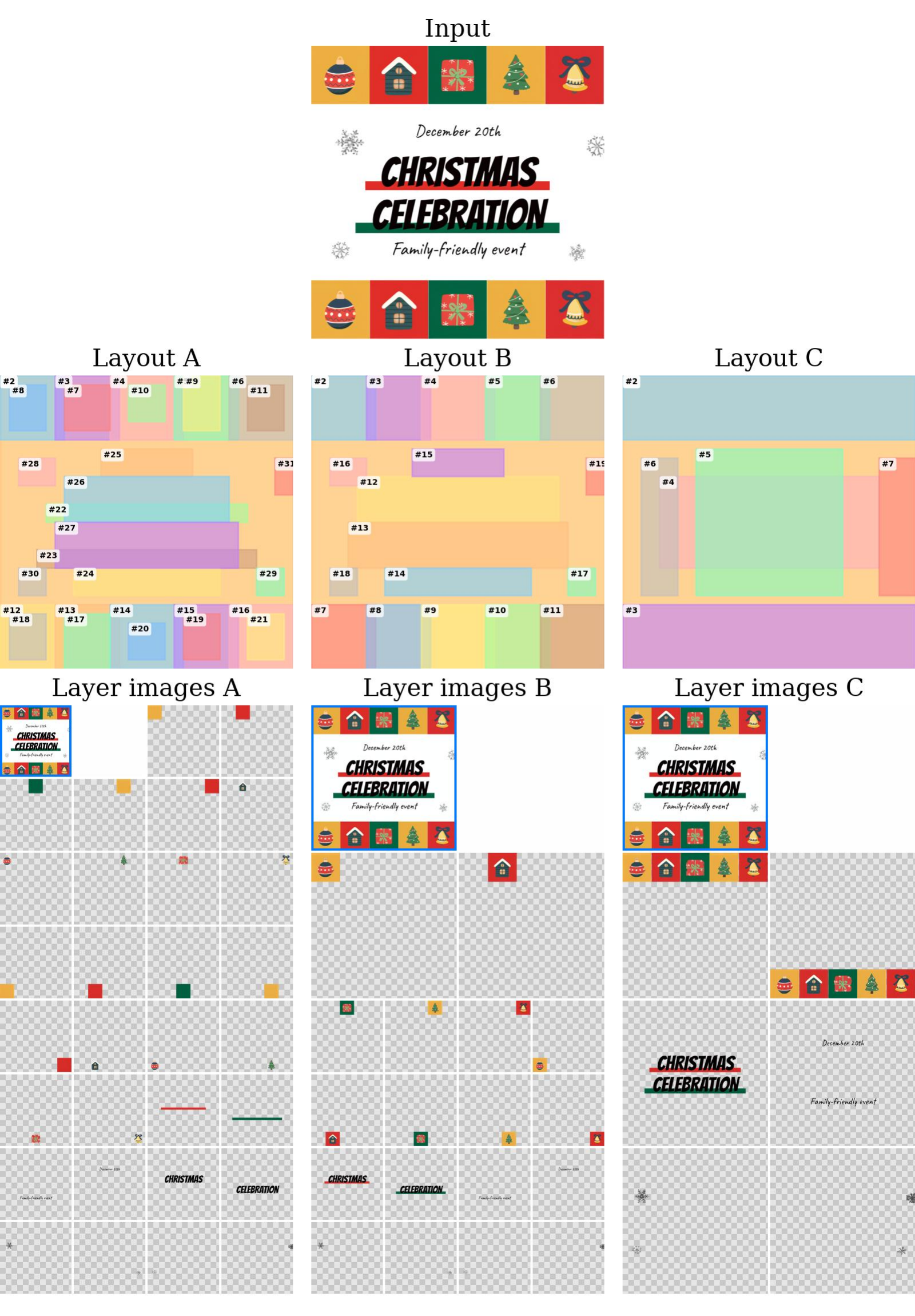}
    \vspace{-5mm}
    \caption{\footnotesize{\textbf{Image-to-layers: Merged image vs. layout visualization.} Another example illustrating the correspondence between input raster images and their layer layouts. Our method leverages layout information (either from automatic detectors or manual annotations) to perform accurate layer decomposition. The layer grouping augmentation strategy helps improve robustness to noisy or ambiguous layout specifications.}}
    \label{fig:supp_i2l_layout_2}
\end{figure*}

\begin{figure*}[t]
    \centering
    \vspace{-8mm}
    \includegraphics[width=0.7\linewidth]{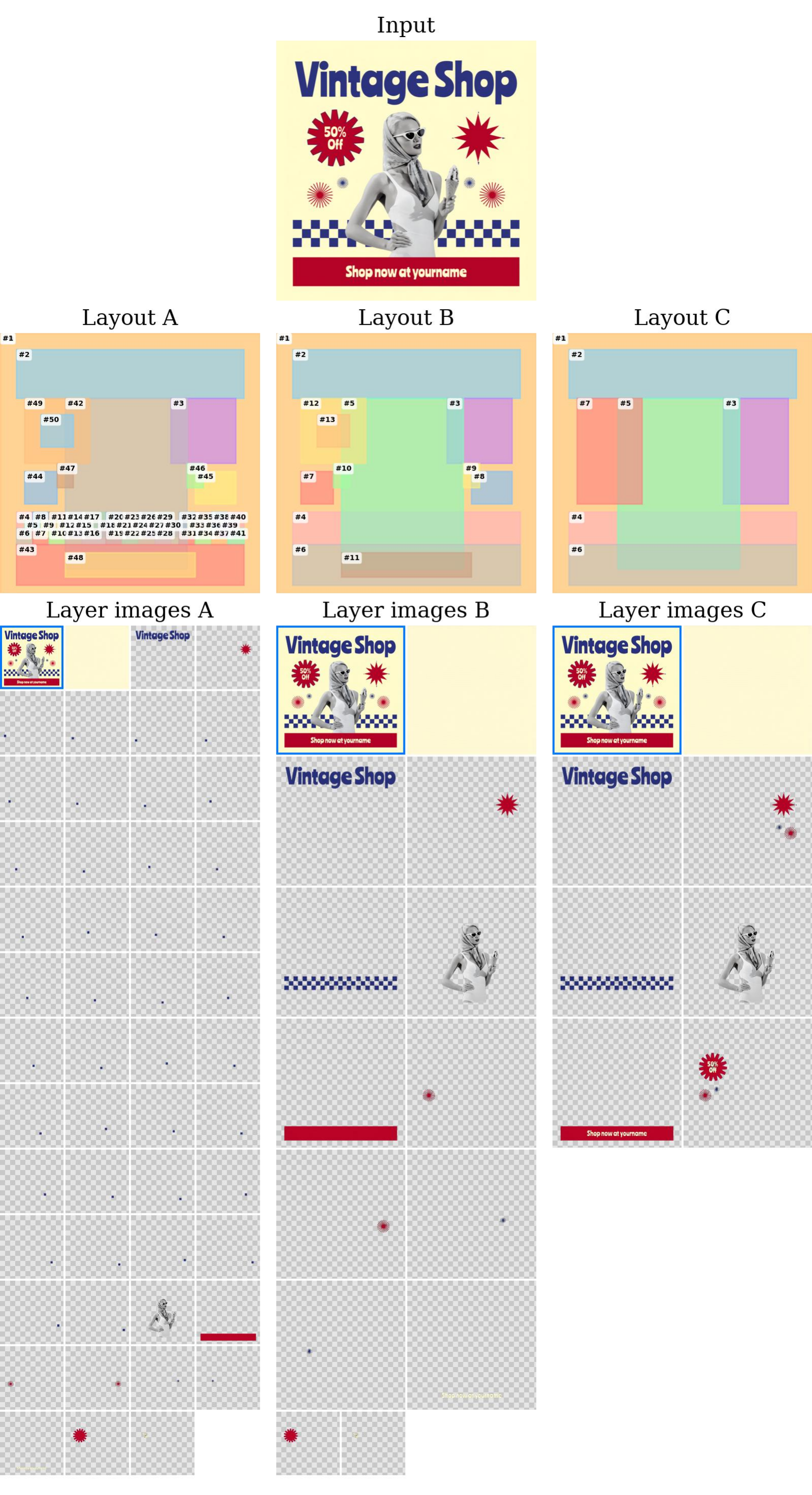}
    \vspace{-5mm}
    \caption{\footnotesize{\textbf{Image-to-layers: Merged image vs. layout visualization.} Final example showing the input-layout relationship in image-to-layers decomposition. This visualization confirms our method's ability to handle diverse design categories and layout complexities, producing high-quality transparent layers that can be independently edited while maintaining faithful reconstruction of the original composition.}}
    \label{fig:supp_i2l_layout_3}
\end{figure*}

For the Image-to-Layer task, Fig.~\ref{fig:supp_i2l_layout_1} through Fig.~\ref{fig:supp_i2l_layout_3} visualize the input raster images alongside their corresponding layout structures (bounding boxes and Z-order) used during inference. These examples illustrate how the model utilizes layout information—whether derived from automatic detectors or manual annotation—as a structural prior to guide the decomposition process. The visualizations show that the model accurately resolves ambiguities in the raster image by leveraging the provided spatial cues, resulting in semantically meaningful layers that strictly conform to the specified boundaries. This highlights the model's ability to produce controllable and predictable decompositions essential for professional editing workflows.

\subsection{Generalization to Natural Scenes}
\begin{figure*}[t]
    \centering
    \includegraphics[width=0.75\linewidth]{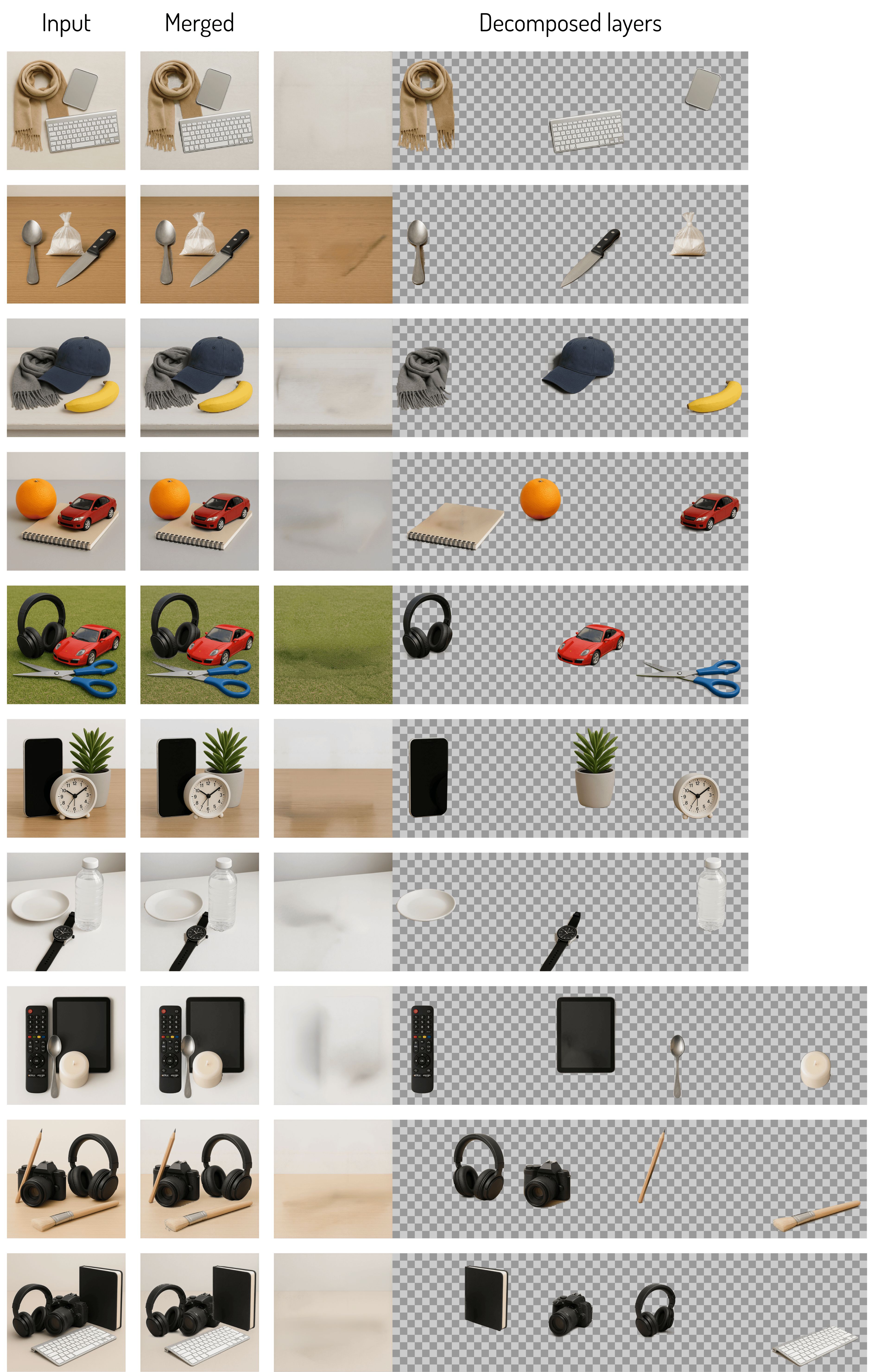}
    \vspace{-2mm}
    \caption{\footnotesize{\textbf{Image-to-layers on real-world photographs: Limitation analysis.} We demonstrate our method's generalization to out-of-domain natural images. Despite being trained exclusively on design datasets, our model can decompose real photographs into layers. However, as discussed in the Limitations section, the model faces challenges with physical effects like shadows—often excluding shadow regions from object layers and leaving them on the background. This limitation stems from the domain gap between planar designs and real-world scenes with lighting effects. Nevertheless, the strong visual understanding from the Qwen-Image backbone enables reasonable generalization, with most objects successfully separated.}}
    \label{fig:supp_i2l_real}
\end{figure*}

Although our model is trained exclusively on graphic design datasets (posters, flyers, etc.), Fig.~\ref{fig:supp_i2l_real} demonstrates its zero-shot generalization capability to real-world natural images. The model successfully segments objects from photographs into transparent layers, leveraging the strong visual understanding inherited from the Qwen-Image backbone. However, we observe a specific limitation due to the domain gap: unlike flat graphic designs, real-world scenes contain complex physical lighting effects. Consequently, the model often fails to associate cast shadows with their respective objects, leaving shadows on the background layer rather than the object layer. Despite this limitation regarding physical lighting effects, the structural decomposition remains surprisingly robust for out-of-domain data.

\subsection{Failure Cases}
\begin{figure*}[t]
    \centering
    \includegraphics[width=0.95\linewidth]{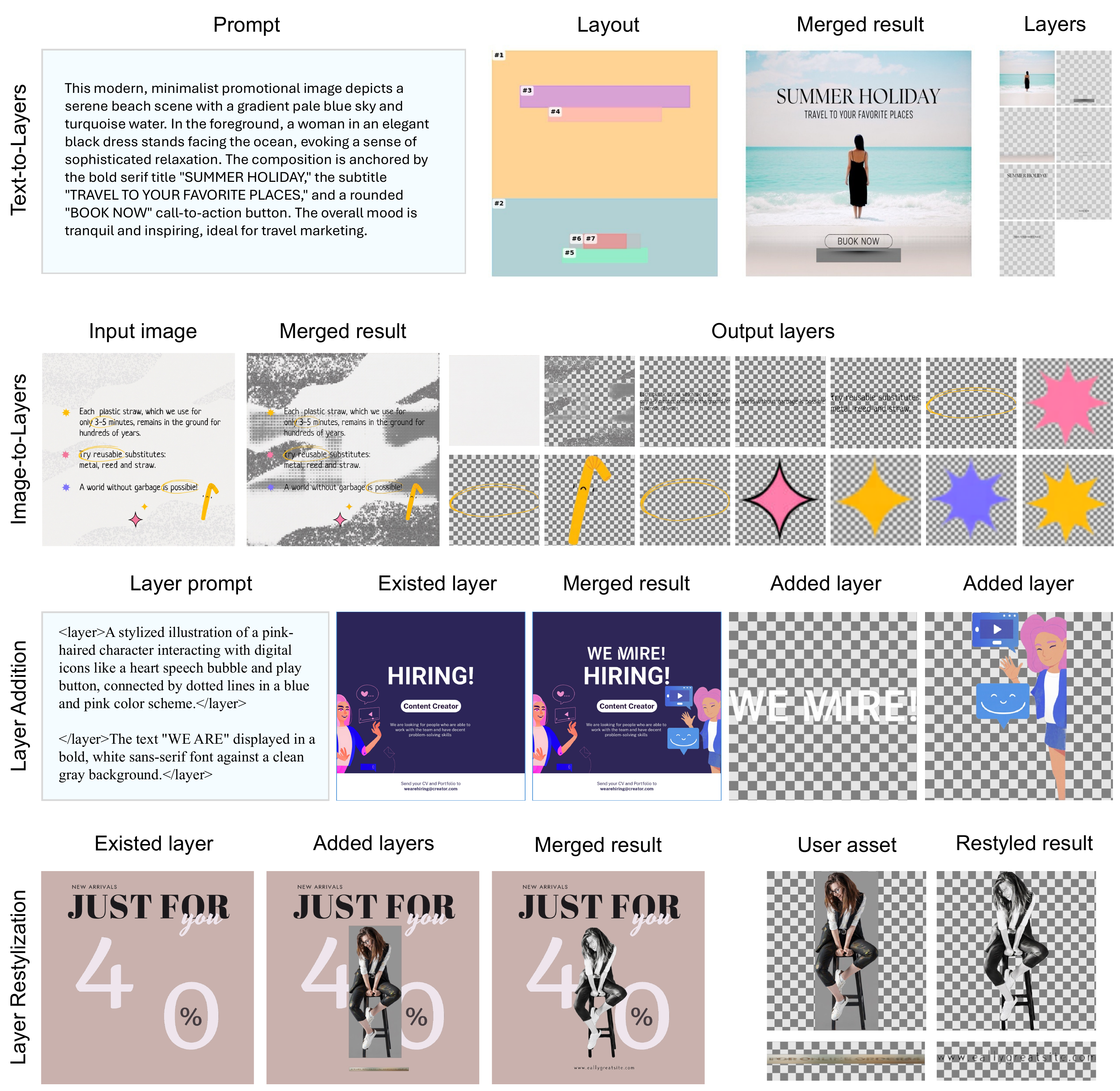}
    \caption{\footnotesize{\textbf{Failure cases and limitations.} We present representative failure cases across our tasks. A common issue (top right) is the "gray background" artifact, where transparent areas are decoded as gray due to the ambiguity of 3-channel VAE encoding. Other limitations include (bottom left) malformed glyphs when generating very small text, and (bottom right) occasional failures in identity preservation and instruction following during layer-to-layer editing.}}
    \label{fig:failure_cases}
\end{figure*}

Finally, we analyze representative failure cases in Fig.~\ref{fig:failure_cases} to provide a balanced view of our method's current limitations. We observe a common issue across all four tasks: some transparent backgrounds are decoded into gray instead of remaining transparent. This ambiguity arises because our VAE encoder currently uses a 3-channel input, which compresses transparent layers into a gray representation that the decoder sometimes misinterprets. Future work could address this by adopting a 4-channel encoder or alternative encoding schemes. Additionally, we identify task-specific limitations: 1) for text generation, the model sometimes struggles with rendering very small glyphs accurately; and 2) for layer-to-layer tasks, we occasionally observe failures in identity preservation (IP) and instruction following, particularly when complex style transfer or precise object insertion is required. These cases outline critical directions for future research in multi-layer generative modeling.

\clearpage

\end{document}